\documentclass[11pt]{article}

\usepackage{PRIMEarxiv}
\usepackage{tabularx}
\usepackage[utf8]{inputenc} %
\usepackage[T1]{fontenc}    %
\usepackage{url}            %
\usepackage{amsfonts}       %
\usepackage{nicefrac}       %
\usepackage{microtype}      %
\usepackage{lipsum}
\usepackage{fancyhdr}       %
\usepackage{graphicx}%
\usepackage[percent]{overpic}
\graphicspath{{media/}}     %
\usepackage{etoolbox}       %added by haoran
\usepackage{makecell}  % 导言部分引入makecell包
\usepackage{soul}
\usepackage{url}
\usepackage[hidelinks]{hyperref}
\usepackage[table]{xcolor}
\usepackage[utf8]{inputenc}
\usepackage{caption}
\usepackage{amsmath}
\usepackage{amssymb}
\usepackage{pifont}
\usepackage{wasysym}
\usepackage{booktabs}
\usepackage{algorithm}
\usepackage{algorithmic}
\usepackage[switch]{lineno}
\usepackage{listings}
\usepackage{xparse}
\usepackage{enumitem}
\usepackage{wrapfig}
\usepackage{adjustbox}
\usepackage{xspace}
\usepackage{listings}
\usepackage{subcaption}
\usepackage[most]{tcolorbox}

\usepackage{multicol}
\usepackage{multirow}
\usepackage{bbding}
\usepackage{circledsteps}

\newtcolorbox{promptbox}{
    colback=blue!5,          % 浅蓝色背景
    colframe=blue!70!black,  % 边框为深蓝色
    coltitle=white,          % 标题文字颜色为白色
    title=Case study - CoT Prompts example, % 标题
    fonttitle=\bfseries,     % 标题字体加粗
    colbacktitle=blue!70!black, % 标题栏为深蓝色
    boxrule=0.75pt,          % 边框厚度
    arc=3pt,                 % 圆角
    left=6pt,                % 左侧内边距
    right=6pt,               % 右侧内边距
    top=5pt,                 % 上侧内边距
    bottom=5pt,              % 下侧内边距
    boxsep=2pt,              % 内容和边框之间的距离
    fontupper=\sffamily\small  % 内容字体为无衬线小号字体
}

\usepackage{newtxtext, newtxmath}
\newcommand{\cmark}{\textcolor{green}{\scalebox{1.2}{\ding{51}}}} %打勾符号
\newcommand{\xmark}{\textcolor{red}{\ding{55}}} %打叉符号

\usepackage{threeparttable}%Table加reference脚注需要

\usepackage{fancyhdr}
\usetikzlibrary{mindmap,shadows}
\usepackage[numbers]{natbib}
\usepackage{datetime}
\usepackage{tocloft}
\usepackage{lipsum}

\usepackage{tikz}
\usepackage{pgfplots}
\usepackage{xcolor}
\usepackage{forest}
\usepackage{tablefootnote}
\usepackage{url}
\usepackage{subfig}

\usepackage{amsfonts,amssymb}
\usepackage{color, soul}
\usepackage{mathrsfs}
\usepackage{cleveref}
\usepackage{float}
\usepackage{wrapfig}

\usepackage{amsthm}
\usepackage{bm}
\usepackage{bbm}
\usepackage{algorithm}
\usepackage{algorithmic}
\usepackage{colortbl}
\usepackage{enumitem}
\usepackage{color}
\usepackage{balance}
\usepackage{stfloats}
\usepackage{makecell}

\definecolor{deepred}{HTML}{CC3333}
\usepackage{amsfonts,amssymb}
\usepackage{bbm}
\usepackage{listings}
\definecolor{myblue}{HTML}{3399CC}
\definecolor{myred}{HTML}{993333}
\definecolor{boxblue}{HTML}{3366CC}
\definecolor{lightpink}{RGB}{255, 230, 230} % #FFCCCC
\definecolor{darkred}{RGB}{185, 38, 74}      % #CC0033
\usepackage{mdframed}
\usepackage{tcolorbox}
\usepackage{enumitem}
\usepackage{fontawesome}

\pgfplotsset{compat=1.18}

\makeatletter
\providecommand\sf@counterlist{}
\makeatother

\title{Political-LLM: Large Language Models in Political Science}

\author{
{\bfseries Lincan Li$^{1}$}\quad
{\bfseries Jiaqi Li$^{2}$}\quad
{\bfseries Catherine Chen$^{3}$\footnotemark[1]}\quad
{\bfseries Fred Gui$^{3}$\footnotemark[1]}\quad
{\bfseries Hongjia Yang$^{4}$}\quad
{\bfseries Chenxiao Yu$^{2}$}\vspace{-1pt}\\
{\bfseries Zhengguang Wang$^{4}$}\quad
{\bfseries Jianing Cai$^{5}$}\quad
{\bfseries Junlong Aaron Zhou$^{6}$}\quad
{\bfseries Bolin Shen$^{1}$}\quad
{\bfseries Alex Qian$^{2}$}\vspace{-1pt}\\
{\bfseries Weixin Chen$^{2}$}\
{\bfseries Zhongkai Xue$^{7}$}\enspace
{\bfseries Lichao Sun$^{8}$}\
{\bfseries Lifang He$^{8}$}\
{\bfseries Hanjie Chen$^{9}$}\
{\bfseries Kaize Ding$^{10}$}\vspace{-1pt}\\
{\bfseries Zijian Du$^{11}$}\quad
{\bfseries Fangzhou Mu$^{12}$}\quad
{\bfseries Jiaxin Pei$^{13}$}\quad
{\bfseries Jieyu Zhao$^{2}$}\quad
{\bfseries Swabha Swayamdipta$^{2}$}\vspace{-1pt}\\
{\bfseries Willie Neiswanger$^{2}$}\quad
{\bfseries Hua Wei$^{14}$}\quad
{\bfseries Xiyang Hu$^{14}$}\quad
{\bfseries Shixiang Zhu$^{15}$}\quad
{\bfseries Tianlong Chen$^{16}$}\vspace{-1pt}\\
{\bfseries Yingzhou Lu$^{13}$}\
{\bfseries Yang Shi$^{17}$}\
{\bfseries Lianhui Qin$^{18}$}\
{\bfseries Tianfan Fu$^{19}$}\
{\bfseries Zhengzhong Tu$^{20}$}\vspace{-1pt}\\
{\bfseries Yuzhe Yang$^{21}$}\
{\bfseries Jaemin Yoo$^{22}$}\
{\bfseries Jiaheng Zhang$^{23}$}\
{\bfseries Ryan Rossi$^{24}$}\
{\bfseries Liang Zhan$^{25}$}\vspace{-1pt}\\
{\bfseries Liang Zhao$^{26}$}\
{\bfseries Emilio Ferrara$^{2}$}\
{\bfseries Yan Liu$^{2}$}\
{\bfseries Furong Huang$^{27}$}\
{\bfseries Xiangliang Zhang$^{28}$}\vspace{-1pt}\\
{\bfseries Lawrence Rothenberg$^{29}$}\quad
{\bfseries Shuiwang Ji$^{20}$}\quad
{\bfseries Philip S. Yu$^{30}$}\quad
{\bfseries Yue Zhao$^{2}$\footnotemark[1]}\quad
{\bfseries Yushun Dong$^{1}$\footnotemark[1]} \vspace{-2pt}\\
\\ 
{\bfseries $^{1}$Florida State University}\
{\bfseries $^{2}$University of Southern California}\
{\bfseries $^{3}$Louisiana State University} \vspace{-1pt}\\
{\bfseries $^{4}$University of Virginia}\
{\bfseries $^{5}$University of Pennsylvania}\
{\bfseries $^{6}$New York University}\
{\bfseries $^{7}$Oxford University}\vspace{-1pt}\\
{\bfseries $^{8}$Lehigh University}\
{\bfseries $^{9}$Rice University}\ 
{\bfseries $^{10}$Northwestern University}\
{\bfseries $^{11}$NVIDIA}\vspace{-1pt}\\
{\bfseries $^{12}$University of Wisconsin-Madison}\
{\bfseries $^{13}$Stanford University}\
{\bfseries $^{14}$Arizona State University}\vspace{-1pt}\\
{\bfseries $^{15}$Carnegie Mellon University}\
{\bfseries $^{16}$University of North Carolina at Chapel Hill}\
{\bfseries $^{17}$Utah State University}\vspace{-1pt}\\
{\bfseries $^{18}$University of California San Diego}\
{\bfseries $^{19}$Rensselaer Polytechnic Institute}\
{\bfseries $^{20}$Texas A\&M University}\vspace{-1pt}\\
{\bfseries $^{21}$University of California, Los Angeles}\
{\bfseries $^{22}$Korea Advanced Institute of Science \& Technology}\vspace{-1pt}\\
{\bfseries $^{23}$National University of Singapore}\
{\bfseries $^{24}$Adobe Research}\
{\bfseries $^{25}$University of Pittsburgh}\vspace{-1pt}\\
{\bfseries $^{26}$Emory University}\
{\bfseries $^{27}$University of Maryland}\
{\bfseries $^{28}$University of Notre Dame}\vspace{-1pt}\\
{\bfseries $^{29}$University of Rochester}\
{\bfseries $^{30}$University of Illinons at Chicago}\ \vspace{-2pt}\\
}

\begin{document}
\maketitle
\renewcommand{\thefootnote}{\fnsymbol{footnote}}
\footnotetext[1]{Corresponding authors: Yushun Dong (\href{mailto:yd24f@fsu.edu}{yd24f@fsu.edu}) is with the Department of Computer Science, Florida State University; Yue Zhao (\href{mailto:yzhao010@usc.edu}{yzhao010@usc.edu}) is with the Department of Computer Science, University of Southern California; Fred Gui (\href{mailto:pgui@lsu.edu}{pgui@lsu.edu}) is with the Department of Political Science, Louisiana State University; Catherine Chen (\href{mailto:catherinechen@lsu.edu}{catherinechen@lsu.edu}) is with the Manship School of Mass Communication and the Department of Political Science, Louisiana State University.}
%\footnotetext[2]{The originally developed online website resource is available at:\textcolor{boxblue}{\href{http://political-llm.org/}{http://political-llm.org/}}}

%\footnotetext[1]{Major contribution.}
%\footnotetext[2]{Yue Huang and Lichao Sun are co-corresponding authors: \href{mailto:yhuang37@nd.edu}{yhuang37@nd.edu}, \href{mailto:lis221@lehigh.edu}{lis221@lehigh.edu}}
%\footnotetext[3]{Visiting Students at LAIR Lab, Lehigh University.}
%\footnotetext[4]{Latest Update: Sep., 2024.}

\vspace{-2em}

\begin{abstract}
In recent years, large language models (LLMs) have been widely adopted in political science tasks such as election prediction, sentiment analysis, policy impact assessment, and misinformation detection. Meanwhile, the need to systematically understand how LLMs can further revolutionize the field also becomes urgent. In this work, we—a multidisciplinary team of researchers spanning computer science and political science—present the \textit{first} principled framework termed \textcolor{boxblue}{\href{http://political-llm.org/} Political-LLM} to advance the comprehensive understanding of integrating LLMs into computational political science. Specifically, we first introduce a fundamental taxonomy classifying the existing explorations into two perspectives: political science and computational methodologies. In particular, from the political science perspective, we highlight the role of LLMs in automating predictive and generative tasks, simulating behavior dynamics, and improving causal inference through tools like counterfactual generation; from a computational perspective, we introduce advancements in data preparation, fine-tuning, and evaluation methods for LLMs that are tailored to political contexts. We identify key challenges and future directions, emphasizing the development of domain-specific datasets, addressing issues of bias and fairness, incorporating human expertise, and redefining evaluation criteria to align with the unique requirements of computational political science. Political-LLM seeks to serve as a guidebook for researchers to foster an informed, ethical, and impactful use of Artificial Intelligence in political science. Our online resource is available at: \textcolor{boxblue}{\href{http://political-llm.org/}{http://political-llm.org/}}.
\end{abstract}

\newpage
\tableofcontents

\begin{tcolorbox}[colback=lightpink, colframe=darkred, title=\textbf{Important Notice}]
\begin{itemize}[leftmargin=*]
% \item \textbf{Online Resource:} The originally developed website resource is available at: \textcolor{boxblue}{\href{http://political-llm.org/}{http://political-llm.org/}}.
\item \textbf{Content Warning:} This paper may contain offensive content generated by LLMs. 
% \item \textbf{Disclaimer:} This study explores LLMs' capacity in election forecasting. Predictions do not reflect the authors' views and should not be interpreted as definitive forecasts.
\item \textbf{Disclaimer:} This study examines the capabilities of Large Language Models (LLMs) in the context of election forecasting. The predictions contained herein are for research purposes only, do not represent the views or opinions of the authors, and shall not be construed as definitive or conclusive forecasts.
% \item \textbf{Copyright Notice:} All figures and tables presented in this survey paper are original. Any reproduction or use requires proper attribution to this paper.
% \item \textbf{Copyright Notice:} All figures and tables presented in this survey paper are original and not AI-generated. Any redistribution should include proper reference to this work.
\item \textbf{Copyright Notice:} All figures in this paper are original creations (not generated with AI tools). Redistribution or reproduction is permitted only with proper attribution to this work.

\end{itemize}
\end{tcolorbox}

\newpage

% (2)
% applications, influences, and contributions to both society and scientific community

\begin{wrapfigure}[17]{r}{0.63\textwidth} 
\centering
\vspace{-1mm}
\includegraphics[width=0.62\textwidth]{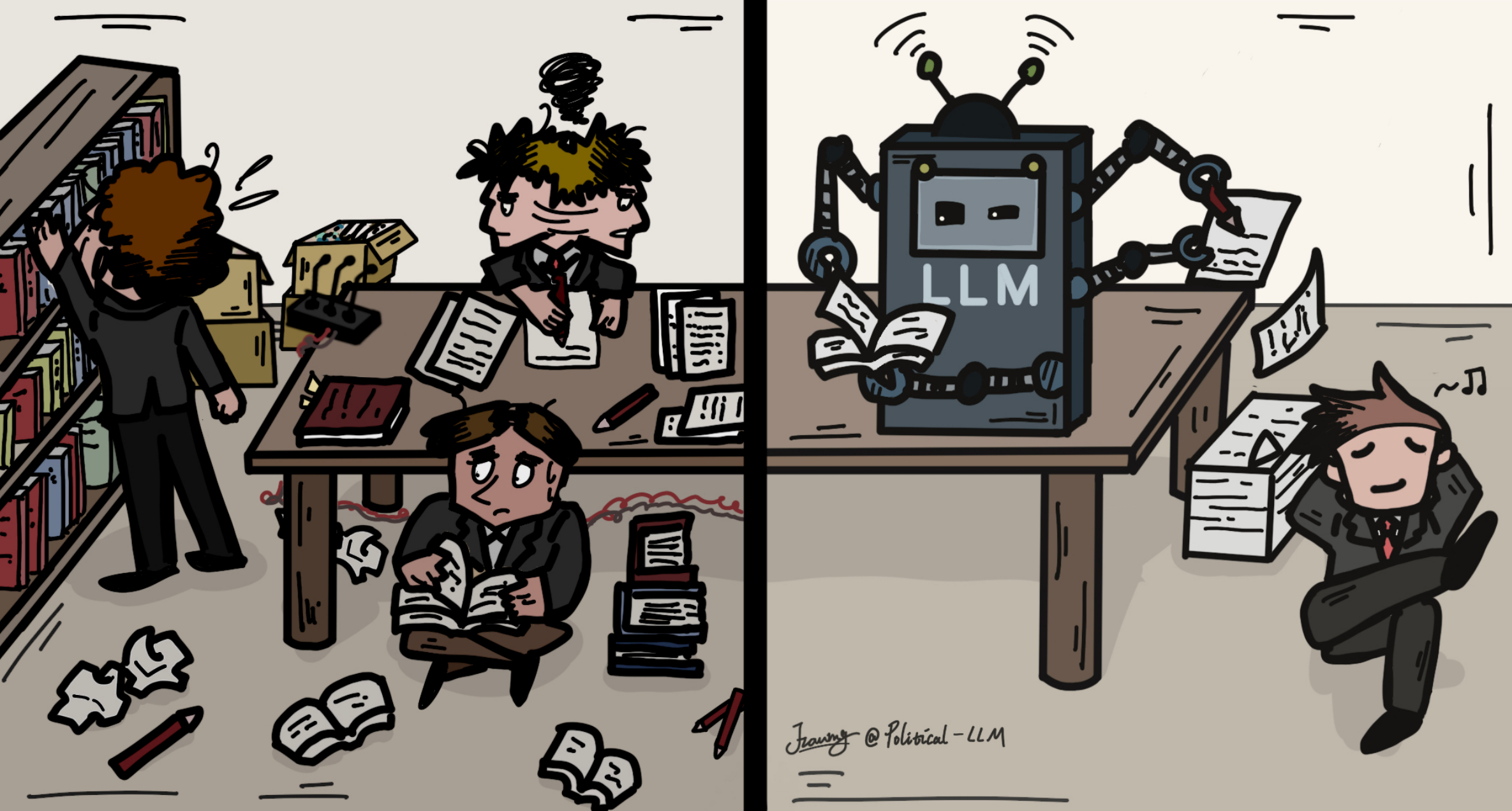}
\caption{LLMs are revolutionizing political science through advanced language analysis and interdisciplinary integration capabilities.}
\label{fig:cartoon}
\end{wrapfigure}

\section{Introduction}
Recent years have witnessed the extraordinary capabilities of Large Language Models (LLMs) and their contributions to a plethora of fields, such as healthcare~\cite{wornow2023shaky,wang2023large,xu2024retrieval,yue2024ct}, finance~\cite{huang2023finbert,wu2023bloomberggpt,xie2024pixiu}, scientific discoveries~\cite{zhang2023artificial,Zhang2024Survey,liu2024drugagent}, transportation~\cite{da2024open,da2024prompt,zhang2024urban,li2024opencity}, and education~\cite{kasneci2023chatgpt,10.1145/3613372.3614197,henkel2024can}, to name a few. The success of LLMs is mainly attributed to the pre-training over web-scale text corpora~\cite{minaee2024large}, which has equipped them with remarkable language intelligence to analyze complicated linguistic patterns~\cite{zhao2023survey,llmpolitical}. Such outstanding capabilities have been found to align with the pursuit of language analysis in a series of sub-fields in social science~\cite{argyle2023out,ziems2024can,demszky2023using}, where political science has stood out with abundant explorations and substantial advancements~\cite{rotaru2024artificial,rodman2024political,gover2023political}. Political science is broadly defined as the study of political systems, behavior, institutions, and policy-making processes, aiming to understand how power and resources are distributed within societies~\cite{moe2005power,gao2022post}. It relies on diverse forms of political data, including legislative documents, political speeches, public opinion surveys, and news reports, which serve as the foundation for political analysis~\cite{wilkerson2017large,chen2022election2020}.

Traditionally, political science relied heavily on qualitative methods~\cite{gerring2017qualitative}, such as content analysis and case studies, alongside quantitative approaches like statistical modeling and surveys, to examine trends and patterns in political behavior. These methods, while foundational, often faced challenges in scaling, handling multilingual and unstructured data, and deriving insights from vast corpora of text. The emergence of LLMs has helped overcome these hurdles by enabling automated, large-scale analysis of political data, providing researchers with unprecedented tools to process and interpret political texts more effectively. In particular, LLMs have been critical in analyzing extensive corpora of political texts~\cite{tornberg2023chatgpt,heseltine2024large}, encompassing a wide range of sources such as political speeches~\cite{liu2023summary,xu2024secap}, legislative documents~\cite{yue2023disc,gesnouin2024llamandement}, social media content~\cite{tornberg2023simulating,najafi2024turkishbertweet}, and news articles~\cite{zhang2024benchmarking,fang2024bias}. Through these practices, LLMs have enabled stakeholders such as researchers and policy makers to gain an in-depth understanding of various facets such as political behavior~\cite{rozado2024political}, public opinion~\cite{breum2024persuasive}, policy formulation~\cite{rivera2024escalation}, and latest election dynamics~\cite{gujral2024can}.
For instance, LLMs have revolutionized sentiment analysis by providing nuanced interpretations of public reactions to political events, policies, and figures~\cite{zhang2023sentiment}. This enhanced capability has been crucial in understanding voter sentiment~\cite{khan2023social} and forecasting election outcomes with greater accuracy~\cite{zhang2024electionsim}.
As another example, researchers have employed LLMs to predict legislative voting patterns~\cite{potter2024hidden}, identify emerging political trends~\cite{santurkar2023whose}, and assess the impact of policy changes on public opinion~\cite{bremer2023public}. The automation of these analytical processes has not only accelerated the pace of research but also increased its precision, allowing for more robust and comprehensive analyses~\cite{wang2024intelligent}. We use a cartoon in Figure~\ref{fig:cartoon} to illustrate the impressive impact LLMs have brought to the area of political science.

% (3) there are gaps

Despite the considerable advancements facilitated by LLMs, significant gaps remain hindering their full potential from being realized in this emerging field~\cite{ziems2024can,halterman2024codebook,mou2024unifying}. Here we identify three most significant gaps below.
The first gap, which arises from an interdisciplinary perspective, results from the absence of a systematic understanding of \textit{how the potential of LLMs interacts with political science}~\cite{ziems2024can}. 
For example, without a well-defined framework for adapting LLMs to analyze political data, researchers face challenges when attempting to utilize these models for sophisticated tasks, such as analyzing legislative patterns~\cite{baker2024simulating} or mapping ideological shifts over time~\cite{chen2024susceptible}. As such, researchers and practitioners working in political science can encounter difficulty when they resort to ready-to-use LLMs.
%
% Without this integration, researchers can miss opportunities to leverage LLMs for more sophisticated analyses, such as tracing the evolution of political discourse over time or predicting policy impacts based on public sentiment.
%
The second gap, which is rooted in computer science, comes from the lack of \textit{fundamental insights about exploiting appropriate techniques to improve LLMs for political science}~\cite{halterman2024codebook}.
As an example, LLMs can exhibit characteristics that are undesirable in political science, such as societal bias~\cite{he2023inducing}, hallucinations~\cite{yao2023llm}, privacy leakage~\cite{yao2024survey}, and high computational costs~\cite{marino2024integrating}. To facilitate the utility of LLMs in political science, appropriate techniques should be leveraged to handle these issues. For instance, knowledge editing techniques~\cite{de2021editing,wang2023knowledge} can be used to mitigate the bias exhibited by LLMs, and machine unlearning~\cite{liu2024rethinking,liu2024towards} can be leveraged to remove privacy-related information from LLMs, thereby building more appropriate LLMs for political science.
The final gap, which comes from political science, lies in \textit{the general deficiency of domain-specific knowledge integrated with LLMs}~\cite{mou2024unifying}. With the lack of such knowledge of political science, obtaining outputs that are properly informed by a nuanced context from general LLMs becomes difficult.
For instance, political texts often contain complex references to historical events, ideological nuances, and policy implications that general LLMs may fail to interpret accurately without specialized training data~\cite{liu2024llm,motoki2024more}.
% %
% Finally, the last gap, which arises from an interdisciplinary perspective, is resulted from the absence of a systematic understanding of integrating the currently available LLMs into political science applications and research works. 
% %
% For example, the effective use of LLMs requires knowledge of various computational techniques, such as natural language processing, machine learning algorithms, and data mining methods, which need to be systematically integrated with traditional political science methodologies. Without this integration, researchers may miss opportunities to leverage LLMs for more sophisticated analyses, such as tracing the evolution of political discourse over time or predicting policy impacts based on public sentiment.
%
% Therefore, despite the emerging attention and explorations, harnessing the power of LLMs in political science applications and research remains xx that calls for research collaborations from both fields.

Research interest in adapting LLMs to political science applications has been growing rapidly in recent years~\cite{potter2024hidden,linegar2023large,halterman2024codebook}. A search in major academic databases shows a more than 300\% increase in publications related to "LLMs and political science" between year 2020 and 2024, highlighting the field's continuous expansion and great potential. This surge is driven by LLMs' ability to process complicated political texts~\cite{liu2024poliprompt}, extract ideological patterns~\cite{kato2024u}, and simulate decision-making processes~\cite{yang2024llm,liu2024dellma} at an unprecedented scale. Back in 2020, Chatsiou et al.~\cite{chatsiou2020deep} discussed the potential influence and applications of LLMs in political science. Nevertheless, the naivety and simplicity of LLMs at that time greatly limited the insights and depth of the discussion. More recently, researchers have investigated specific tasks in political science domain, leveraging LLMs as the main approach.~\cite{liu2024llm,kato2024u,wu2023large} focused on debates analysis and ideological mapping.~\cite{chalkidis2024investigating,rotaru2024artificial,moghimifar2024modelling} centered on political election and voting, including election outcome prediction, election dynamics, and voting behavior modeling.~\cite{sanders2023demonstrations,hackenburg2023comparing} investigated the role of LLMs in shaping opinions and conducting polls.~\cite{lazar2024can,gudino2024large} explored the contribution of LLMs in enhancing democracy and society values. In contrast to the emergence of works on specific task-level applications, there are only a few survey studies related to LLMs in political science~\cite{linegar2023large,ziems2024can}. A comprehensive and systematic survey of recent advances aimed at fostering an in-depth understanding of this topic is urgently needed to assist researchers from multiple fields and to illuminate opportunities for cross-disciplinary ideas. 
% The detailed discussion and comparison with existing surveys can be found in \textcolor{blue}{\hyperlink{difference-w-existing-survey}{This Paragraph}}.

In this survey, we aim to provide a comprehensive examination of leveraging LLMs' power to harness the field of political science, addressing key challenges and identifying future research directions. Specifically, we begin by presenting a novel taxonomy to systematically classify existing works in this interdisciplinary domain. This taxonomy categorizes methods and applications, enabling researchers and practitioners to navigate the field more effectively and bridging the first gap identified above regarding the lack of systematic understanding. Next, we explore current advancements from both political and technical perspectives, highlighting techniques for adapting LLMs to political science applications and addressing domain-specific challenges. This section addresses the second gap by delving into solutions for mitigating undesirable characteristics of LLMs, such as societal biases, hallucinations, and privacy concerns, and discusses techniques like knowledge editing and machine unlearning. To bridge the third gap, we further discuss the integration of political science domain knowledge with LLMs, examining the distinct nature of political information and proposing countermeasures to adapt general LLMs for nuanced political tasks. We illustrate how specialized datasets and fine-tuning approaches can be utilized to enhance contextual accuracy and depth. Finally, we explore real-world applications of LLMs in political science, ranging from election prediction and policy analysis to misinformation detection, and conclude with an examination of the current challenges and promising future directions. The survey thus serves as a resource for advancing the understanding and application of LLMs in political science.

The main contributions of this survey paper are summarized as:
\vspace{-1em}
\begin{itemize}[leftmargin=*]
\setlength{\itemsep}{3pt}
\setlength{\parskip}{3pt}
    %\item \textbf{\emph{A Novel Principled Taxonomy.}} We propose a novel taxonomy for adapting LLMs in the realm of political science. This taxonomy includes xx: 1, 2, 3, 4, and 5. For each type of xx, we introduce their intuition, definitions, exemplary works, and common evaluation protocol.
    \item \textbf{\emph{A Novel Principled Taxonomy.}} We propose a novel taxonomy for adapting LLMs in political science, structured around two main categories: Classic Political Science Functionality and LLM-Driven Methodologies as shown in Figure~\ref{fig-taxonomy}. The first category addresses core political science tasks, including predictive and generative tasks, simulation, causal inference, and social impacts. The second category focuses on computational methods that customize LLMs to political contexts, including benchmark datasets,  data preparation strategy, model design under zero-shot/few-shot learning and fine-tuning scenarios, and inference techniques. The taxonomy provides a systematic framework to bridge existing knowledge gaps, guiding researchers in understanding and applying LLMs effectively within political science field.
    %\item \textbf{\emph{Comprehensive and Multi-Perspective Review.}} We provide a comprehensive and fundamental review of each type of works from both the perspectives of political science and the learning techniques. For each perspective, we xx.
    \item \textbf{\emph{Comprehensive and Multi-Perspective Review.}} We offer a comprehensive review of existing works from both political science and computer science perspectives, ensuring a balanced analysis that highlights the interdisciplinary nature of Political LLMs. From political perspective, we examine how these models capture and interpret complicated political concepts, historical nuances, and ideological shifts, with a focus on their practical implications. From a computer science perspective, we delve into the specific technologies used to adapt and enhance LLMs for political tasks, covering aspects like fine-tuning strategies, prompt engineering, model architecture adjustments, and inference schemes. For each perspective, we provide insights into both the theoretical and practical challenges involved.
    %\item \textbf{\emph{Challenges and Future Directions.}} We present the limitations of current research and point out pressing challenges. Open research questions are also discussed for further advances.
    \item \textbf{\emph{Challenges and Future Directions.}} We identify key limitations of current research and outline pressing challenges that remain unsolved, drawing attention to areas where further advances are needed. These challenges include mitigating societal biases in LLM outputs, enhancing contextual accuracy, managing privacy concerns, and addressing computational costs associated with deploying LLMs in political science. Additionally, we highlight open research questions, such as the need for domain-specific datasets, the development of new evaluation metrics suited to political science tasks, and innovative techniques to improve the interpretability of LLM outputs. These discussions are aimed at inspiring future research endeavors and fostering advancements that broaden the applicability and reliability of Political LLMs.
\end{itemize}

\hypertarget{difference-w-existing-survey}
{
\noindent \textbf{Difference with Existing Surveys.} Despite the abundant explorations in this interdisciplinary area, current survey works remain limited. Researchers have realized and discussed the potential revolutionary contribution of language models as early as in 2017~\cite{wilkerson2017large,terechshenko2020comparison}. However, these works mainly focus on traditional language models and are unable to provide insights about more recent LLMs.
After that, multiple survey works have realized the potential of LLMs for political science~\cite{chatsiou2020deep,rodman2024political}. However, their discussion lacks a systematic understanding of how LLMs can be adopted in various political science applications and research. 
More recently, LLM-based applications on specific political or social tasks have been reviewed in several survey works~\cite{lee2024applications,argyle2023out,llmpolitical,rozado2023political,ziems2024can,ornstein2022train,weidinger2021ethical}. Nevertheless, these works overwhelmingly focus on applications while the discussion from a technical perspective is ignored. Therefore, it remains unclear how LLMs can be improved to be better adapted.
Different from all the survey works above, we aim to present a systematic and comprehensive understanding of leveraging the power of LLMs in political science. Specifically, we equip this paper with a novel principled taxonomy to classify existing works, such that researchers and practitioners can have a broader picture of this interdisciplinary field. Meanwhile, we perform a discussion on each type of work from both political and technical perspectives, which reveals how LLMs can be improved to be better adapted. Table~\ref{tab:compare-existing-works} provides a detailed comparison between our survey and other related surveys in political or social science.
}

\begin{table*}[htbp]
\setlength{\tabcolsep}{13pt}
\renewcommand{\arraystretch}{0.8}
\centering
\caption{Comparison with Existing Surveys on Broad Political Science Field (Abbreviations: PoliSci = Political Science, CPS = Computational Political Science, CS = Computer Science).}
% \footnotesize
\label{tab:compare-existing-works}
\resizebox{0.99\textwidth}{!}{%
%\begin{tabular}{c|c|c|c|c|c|c|c|c}
\begin{tabular}{ccccccccc}
\toprule
                                                    & ~\cite{ziems2024can} & ~\cite{argyle2023out} & ~\cite{ornstein2022train} & ~\cite{rozado2023political} & ~\cite{weidinger2021ethical} & ~\cite{llmpolitical} & ~\cite{lee2024applications} & \textbf{Ours} \\ \midrule
\begin{tabular}[c]{@{}l@{}}\makecell{\vspace{-1mm}Proposed Taxonomy on \\ LLM for PoliSci}\end{tabular}                         & \xmark                   & \xmark                   & \xmark                                 & \xmark                    & \xmark                              & \xmark                            & \xmark              & \cmark          \\ \midrule
\begin{tabular}[c]{@{}l@{}}\makecell{\vspace{-1mm}Literature Review from \\ PoliSci Perspective}\end{tabular}           & \cmark                 & \xmark                 & \cmark                     & \xmark                    & \xmark                 & \cmark                    & \cmark                    & \cmark             \\ \midrule
\begin{tabular}[c]{@{}l@{}}\makecell{\vspace{-1mm}Literature Review from \\ CS Perspective}\end{tabular}               & \xmark              & \xmark            & \cmark                       & \xmark                    & \cmark                     & \cmark                     & \xmark                    & \cmark             \\ \midrule
\begin{tabular}[c]{@{}l@{}}\makecell{\vspace{-1mm}Structured Analysis of \\
CPS Methodologies}\end{tabular} & \xmark                 & \xmark                    & \xmark                     & \xmark             & \xmark                   & \xmark                  & \xmark                         & \cmark             \\ \midrule
\begin{tabular}[c]{@{}l@{}}\makecell{\vspace{-1mm}Include Experiments \\and Evaluations}\end{tabular}                                                                          & \cmark                     & \cmark                        & \cmark                                 & \cmark                    & \xmark                  & \xmark                               & \cmark                         & \cmark             \\ \midrule
\begin{tabular}[c]{@{}l@{}}\makecell{\vspace{-1mm}Application \\Examples}\end{tabular}                                                                                                     & \cmark                 & \cmark                  & \cmark                                 & \cmark              & \cmark                   & \cmark                               & \cmark                        & \cmark             \\ \midrule
\begin{tabular}[c]{@{}l@{}}\makecell{\vspace{-1mm}Comprehensive Summary \\ of Benchmarks}\end{tabular}                                     & \cmark                          & \xmark                             & \xmark                                 & \xmark                    & \xmark                              & \xmark                         & \xmark                  & \cmark            \\ \midrule
\begin{tabular}[c]{@{}l@{}}\makecell{\vspace{-1mm}Analyzing Limitations in \\ Existing Methodologies}\end{tabular}                    & \cmark         & \cmark             & \cmark                & \cmark               & \cmark         &\cmark             & \cmark             & \cmark          \\ \midrule
\begin{tabular}[c]{@{}l@{}}\makecell{\vspace{-1mm}Future Research \\Direction}\end{tabular}         & \cmark        & \cmark     & \cmark            & \cmark         & \cmark             & \cmark            & \xmark                   & \cmark            \\ \bottomrule
\end{tabular}
}
\begin{threeparttable}
\begin{tablenotes}
\footnotesize
\item Caleb Ziems et al., 2024~\cite{ziems2024can}; Lisa P. Argyle et al., 2023~\cite{argyle2023out}; Joseph T. Ornstein et al., 2022~\cite{ornstein2022train}; David Rozado, 2023~\cite{rozado2023political}; Laura Weidinger et al., 2021~\cite{weidinger2021ethical}; Mitchell Linegar et al., 2023~\cite{llmpolitical}; Kyuwon Lee et al., 2024~\cite{lee2024applications}.
\end{tablenotes}
\end{threeparttable}
\end{table*}

\noindent \textbf{Intended Audiences.}
%The intended audiences of this survey are (1) computer science researchers and practitioners who would like to systematically understand how LLMs have been leveraged in political science; and (2) political science researchers and practitioners who aims to exploit LLMs under domain-specific scenarios in an appropriate way.
The intended audience of this survey includes (1) computer science researchers and practitioners who seek a structured understanding of how LLMs are applied in political science, aiming to bridge interdisciplinary gaps; and (2) political science researchers and practitioners who seek to leverage LLMs in ways that are sensitive to the unique requirements of their field, such as nuanced interpretation and contextual accuracy~\cite{he2023inducing}. This survey also benefits interdisciplinary scholars who are interested in exploring the intersection of AI and social sciences, as well as policymakers and government agencies intending to employ LLM-based tools for public opinion analysis, election forecasting, and legislative research tasks.

\noindent \textbf{Structure of This Survey Paper.}
%\lipsum[1-1]
The remainder of this article is organized as follows. In Section~\ref{preliminary}, we provide the related preliminaries of LLM and political science. Following that, the proposed taxonomy of LLM for political science is detailed in Section~\ref{taxonomy}. Section~\ref{Traditional-Method} introduces the core functionalities of political science that can benefit from LLMs, including predictive tasks, generative tasks, simulation, causal inference, and ethical concerns \& social impacts. Section~\ref{Computer-Science} focuses on LLM-driven methodologies tailored for political science, covering benchmark datasets, data processing strategies, fine-tuning and inference on LLMs, and an empirical study on voting simulation. Subsequently, Section~\ref{sec:challenges} discusses potential future directions, outlining research opportunities to enhance the effectiveness and applicability of LLMs in the field of political science. Finally, we conclude the survey paper in Section~\ref{sec:conclusion}.

\section{Preliminaries} \label{preliminary}
\noindent \textbf{Computational Political Science (CPS).}
Computational Political Science (CPS) is an interdisciplinary field that integrates computational methods with political science to analyze political systems, behaviors, and outcomes~\cite{haq2020survey}. By leveraging tools such as data analytics, machine learning, and natural language processing (NLP), CPS enhances the understanding of complex political phenomena. The field has evolved from relying on traditional statistical models, such as regression-based analyses, to embracing AI-driven approaches that enable the processing of large-scale, unstructured political data. This shift has been particularly transformative in tasks like election forecasting, public opinion analysis, and policy evaluation, where modern techniques offer greater scalability and accuracy.

\noindent \textbf{Evolution of Language Models in Political Science.}
Early applications of AI in political science relied on rule-based systems and traditional machine learning methods~\cite{grimmer2021machine}, such as logistic regression~\cite{nicolau2007analysis} and support vector machines (SVMs)~\cite{d2014separating}, to perform basic political tasks. These methods were limited by their reliance on manually crafted features and structured data~\cite{grimmer2021machine}. The advent of pre-trained language models, including Word2Vec~\cite{mikolov2013distributed} and BERT~\cite{devlin2018bert}, marked a significant shift in natural language processing, enabling the analysis of large-scale, unstructured political text. By capturing contextual relationships and semantic nuances in data, these models greatly enhanced the ability to process complex political discourse, advancing domain applications like policy analysis, legislative interpretation, and public opinion mining.

\subsection{Large Language Models (LLMs)}

%\noindent \textbf{Foundational Architectures and the Evolution of LLMs.} 
The foundation of most LLMs lies in the Transformer architecture~\cite{vaswani2017attention}, which introduced the self-attention mechanism to effectively model long-range dependencies in text. This innovation marked a departure from earlier sequence models like RNNs and LSTMs, which struggled with vanishing gradients and limited context windows. Core components such as multi-head attention, feedforward layers, and positional encodings enabled Transformers to process sequences in parallel, significantly improving scalability and efficiency. Early LLMs, such as BERT~\cite{devlin2018bert}, leveraged the Transformer framework through masked language modeling, excelling in bidirectional context understanding. Autoregressive architectures like GPT~\cite{brown2020languagemodelsfewshotlearners} later extended these capabilities, focusing on sequential token prediction for fluent and coherent text generation. The advent of models like T5~\cite{raffel2020exploring} unified various NLP tasks under a single architecture by using sequence-to-sequence learning. Recent advancements~\cite{salemi2024evaluating,kirk2024improving,song2024low,zhu2024sampleattention} further evolved LLM architectures, emphasizing efficiency and task-specific adaptability. Additionally, innovations like multimodal architectures and scalable models such as LLaMA~\cite{touvron2023llamaopenefficientfoundation} and GPT-4~\cite{achiam2023gpt} demonstrate a shift toward systems capable of cross-domain understanding and dynamic interaction, underpinning the transformative potential of LLMs in computational tasks across fields.

%\noindent \textbf{LLM Pre-training Paradigms.} LLMs rely on two predominant pre-training paradigms: masked language modeling (MLM) and autoregressive modeling (AR). MLM, exemplified by BERT, predicts masked tokens within sentences, fostering bidirectional contextual understanding by learning from both left and right contexts simultaneously. This approach excels in tasks requiring nuanced comprehension, such as question answering and sentence classification. In contrast, AR models like GPT generate tokens sequentially, optimizing for fluency, coherence, and the ability to model long sequences. This unidirectional approach suits generative tasks, such as text completion and story generation. Recent advances combined both encoding (MLM) and decoding (AR) capabilities, enhancing model versatility. These hybrid models excel in tasks ranging from summarization to translation. Together, these paradigms equip LLMs with a robust foundational understanding of language, enabling them to generalize across diverse tasks while supporting both comprehension-focused and generation-focused applications.
\noindent \textbf{General Purpose LLMs.}
General-purpose LLMs like GPT~\cite{achiam2023gpt} and BERT~\cite{devlin2018bert} are developed through two primary training paradigms: autoregressive modeling (AR)~\cite{schuurmans2024autoregressive} and masked language modeling (MLM)~\cite{nozza2020mask}. AR models, exemplified by GPT, generate tokens sequentially, prioritizing fluency and coherence in text generation. MLM, as utilized in BERT, predicts masked tokens within sentences, fostering a nuanced contextual understanding of language. These pre-training paradigms equip LLMs with a robust foundational understanding of linguistic patterns, making them highly adaptable for fine-tuning on task-specific datasets. Leveraging these versatile models, researchers can efficiently address domain-specific challenges without undertaking resource-intensive pre-training.

%\noindent \textbf{LLM Fine-tuning Techniques.} Fine-tuning adapts pre-trained LLMs to specific tasks or domains, bridging the gap between general language understanding and specialized applications. Supervised fine-tuning uses labeled datasets to refine model outputs, ensuring alignment with task-specific objectives such as sentiment analysis, translation, or question answering. This approach enhances task accuracy by introducing domain-specific nuances into the model’s parameters. Instruction fine-tuning, a newer paradigm, trains models to follow user directives more effectively by using datasets of instructions and corresponding outputs. This method underpins models like InstructGPT, making them better at adhering to user intents and producing contextually appropriate responses. Beyond task-specific optimization, this technique improves the versatility of LLMs in handling a broader range of user inputs. Reinforcement Learning with Human Feedback (RLHF) builds on fine-tuning by incorporating iterative feedback loops. Human evaluators rank model responses, guiding the model toward outputs that align with human preferences. This approach not only refines the quality of outputs but also mitigates harmful or biased behaviors, ensuring ethical and user-friendly applications. 
\noindent \textbf{LLM Fine-tuning Techniques.}
Fine-tuning adapts pre-trained general-purpose LLMs to downstream specialized applications. Supervised fine-tuning refines model outputs using labeled datasets, aligning them with task-specific objectives~\cite{li2023label}. Instruction fine-tuning trains models to better follow user directives through datasets of instructions and outputs, enhancing adherence to user intents and versatility~\cite{zhang2023instruction}. Reinforcement Learning with Human Feedback (RLHF)~\cite{leerlaif} leverages human evaluators to rank responses, guiding LLMs to align with human preferences while reducing harmful or biased behaviors.

%\noindent \textbf{Zero-shot, Few-shot, and In-context Learning.} LLMs demonstrate remarkable capabilities in zero-shot, few-shot, and in-context learning, leveraging pre-trained knowledge to perform tasks with minimal or no additional training. Zero-shot learning allows models to generalize across tasks solely based on pre-existing knowledge, enabling them to respond to queries or generate outputs for entirely new domains without requiring task-specific data. Few-shot learning builds on this adaptability by utilizing a small number of labeled examples to guide the model. By including examples within the input prompt, the model fine-tunes its responses to align with specific task requirements. For instance, few-shot learning can be used to classify sentiment in legislative speeches or predict voter preferences using limited annotated datasets. In-context learning, a more dynamic approach, equips LLMs to adapt to novel tasks through task descriptions and examples embedded directly in the prompt. Unlike traditional training methods which require huge parameter updates, this approach relies on its ability to infer patterns from the provided context. This method is particularly effective for exploratory tasks like summarizing policy documents or generating counterarguments in debates, where pre-defined training datasets may not exist.

\noindent \textbf{Zero-shot, Few-shot, and In-context Learning.} LLMs demonstrate remarkable capabilities in zero-shot~\cite{kojima2022large}, few-shot~\cite{perez2021true}, and in-context learning~\cite{ram2023context}, leveraging pre-trained knowledge to perform tasks with minimal or no additional training. Zero-shot learning enables task generalization without task-specific training, while few-shot learning benefits from a minimal set of labeled examples. In-context learning, which is achieved through task descriptions and examples within prompts, empowers models to dynamically adapt to novel tasks without parameter updates.

\noindent \textbf{LLM Inference and Decoding Techniques.} Effective inference strategies are crucial for generating high-quality outputs from LLMs. Methods like greedy decoding~\cite{prabhu2024pedal} and beam search~\cite{xie2024self} prioritize sequence coherence, while nucleus sampling~\cite{grubisic2024priority} enhances diversity by sampling within the top-probability distribution. Advanced techniques like Retrieval-Augmented Generation (RAG)~\cite{salemi2024evaluating} integrate external knowledge bases, while prompt engineering~\cite{white2023prompt}, Chain-of-Thought (CoT)~\cite{yao2024tree}, and knowledge injection techniques~\cite{martino2023knowledge} improve task-specific performance, especially in complex scenarios.

%\noindent \textbf{LLM Scalability and Efficiency.} The scalability of LLMs hinges on advancements in distributed training, parameter-efficient techniques, and hardware optimizations. Distributed frameworks, such as those leveraging GPUs or TPUs, enable parallel processing across multiple nodes, facilitating the training of models with billions of parameters. Techniques like Low-Rank Adaptation (LoRA) and adapters focus on parameter-efficient fine-tuning by modifying only a small subset of model weights, significantly reducing memory and computational demands while preserving performance. Innovations in hardware acceleration, including the development of custom AI processors and energy-efficient designs, further enhance training scalability. These advancements allow researchers to tackle increasingly complex tasks and train larger models with reduced time and energy costs, promoting accessibility and environmental sustainability in large language model related research.

\noindent \textbf{LLM Scalability and Efficiency.} LLM scalability relies on distributed training frameworks~\cite{narayanan2021efficient}, efficient parameter adaptation~\cite{ding2023parameter}, fast inference and serving frameworks~\cite{wu2023fast}, and hardware optimizations~\cite{song2024powerinfer}. Techniques like LoRA~\cite{ren2024melora} and adapters~\cite{fu2023effectiveness} enable parameter-efficient fine-tuning, reducing computational requirements without compromising performance. Software frameworks such as vLLM~\cite{kwon2023efficient} and TensorRT-LLM~\cite{tensorrt-llm} facilitate fast LLM inference and serving through advanced batching and memory management. Hardware acceleration, including GPU and TPU advancements~\cite{song2024powerinfer,wu2023fast}, supports the training and inference of increasingly large models, driving efficiency in both computation and energy consumption.

\subsection{Core Computational Political Science Concepts}

\textbf{Political Data Sources and Text Generation.} Political data encompasses diverse sources such as political news, speeches, legislative records, party manifestos, social media contents, etc. Analyzing these data requires handling challenges like data scarcity, imbalance, and linguistic nuances, which hinder comprehensive analysis. One critical application in CPS is \textit{Political Text Generation}, where LLMs are employed to produce political content such as speeches, policy briefs, and debate scripts~\cite{zhang2023survey}. These generative models assist political figures and analysts by creating coherent, persuasive, and contextually relevant texts. LLMs can simulate political scenarios and craft narratives, shaping public opinion and enhancing political communication.

\textbf{Election Prediction and Voting Behavior.} Election prediction focuses on forecasting voter turnout, swing state dynamics, and overall electoral outcomes. LLMs analyze a combination of historical election data, public opinion surveys, and social media discourse to identify patterns influencing voter behavior~\cite{rotaru2024artificial,potter2024hiddenpersuadersllmspolitical}. These models provide insights into key demographic and psychological factors affecting voter preferences, aiding political campaigns and policymakers in tailoring strategies to engage the electorate effectively.

\textbf{Policy and Legislative Interpretation.} Policy and legislative interpretation involves analyzing complex legal texts, such as bills, statutes, administrative rules, and debates, to understand their implications and the ideologies they represent. LLMs excel at parsing and summarizing these documents, identifying key arguments, and predicting potential policy outcomes~\cite{cheong2024not}. This capability offers political scientists a deeper understanding of legislative processes and helps anticipate the effects of policy changes on societal and political structures.

\textbf{Misinformation/Fake News Detection.} Safeguarding political discourse requires addressing misinformation and fake news, which can significantly distort public opinion and decision-making. LLMs are adept at detecting false or biased information by analyzing the structure, intent, and credibility of news articles, social media posts, and political statements~\cite{wu2024fake}. By flagging harmful content, these models ensure the integrity of political information and contribute to maintaining a healthy democratic environment.

\textbf{Political Risk and Conflict Prediction.} Political risk and conflict prediction aim to forecast the likelihood of political instability, unrest, or international conflict~\cite{treisman2023great}. CPS-based methods can analyze geopolitical data, diplomatic communications, and historical trends to identify early signs of conflict and assess the risks involved in political decisions. These predictions are invaluable for policymakers and international organizations in making informed decisions and preparing for possible crises.

\textbf{Political Game Theory and Negotiation.} Political game theory and negotiation involve modeling strategic interactions between political entities, such as governments, parties, or international entities~\cite{meta2022human}. The latest advancements in LLMs hold promise in analyzing negotiation strategies, predicting the outcomes of political bargaining, and identifying optimal decision-making approaches. By simulating various political scenarios, LLMs are expected to have a better understanding of power dynamics, coalition building, and diplomatic negotiations in international politics.

LLMs act as pivotal tools bridging computational methodologies and political science applications. By integrating advanced language processing capabilities with political data analysis, LLMs enable breakthroughs in tasks such as election forecasting~\cite{rotaru2024artificial}, legislative text summarization~\cite{colombo2024leveraging}, and combating misinformation~\cite{wu2024fake}. These advancements demonstrate how LLMs transcend traditional limitations, providing scalable, adaptable, and effective solutions for political science research.

\section{Taxonomy on LLM for Political Science} \label{taxonomy}

% \begin{figure}[htbp]
% \centering
% \includegraphics[width=0.96\linewidth]{figures/Taxonomy.pdf}
% \caption{The proposed Taxonomy on LLM for Political Science.}
% \label{fig-taxonomy}
% \end{figure}

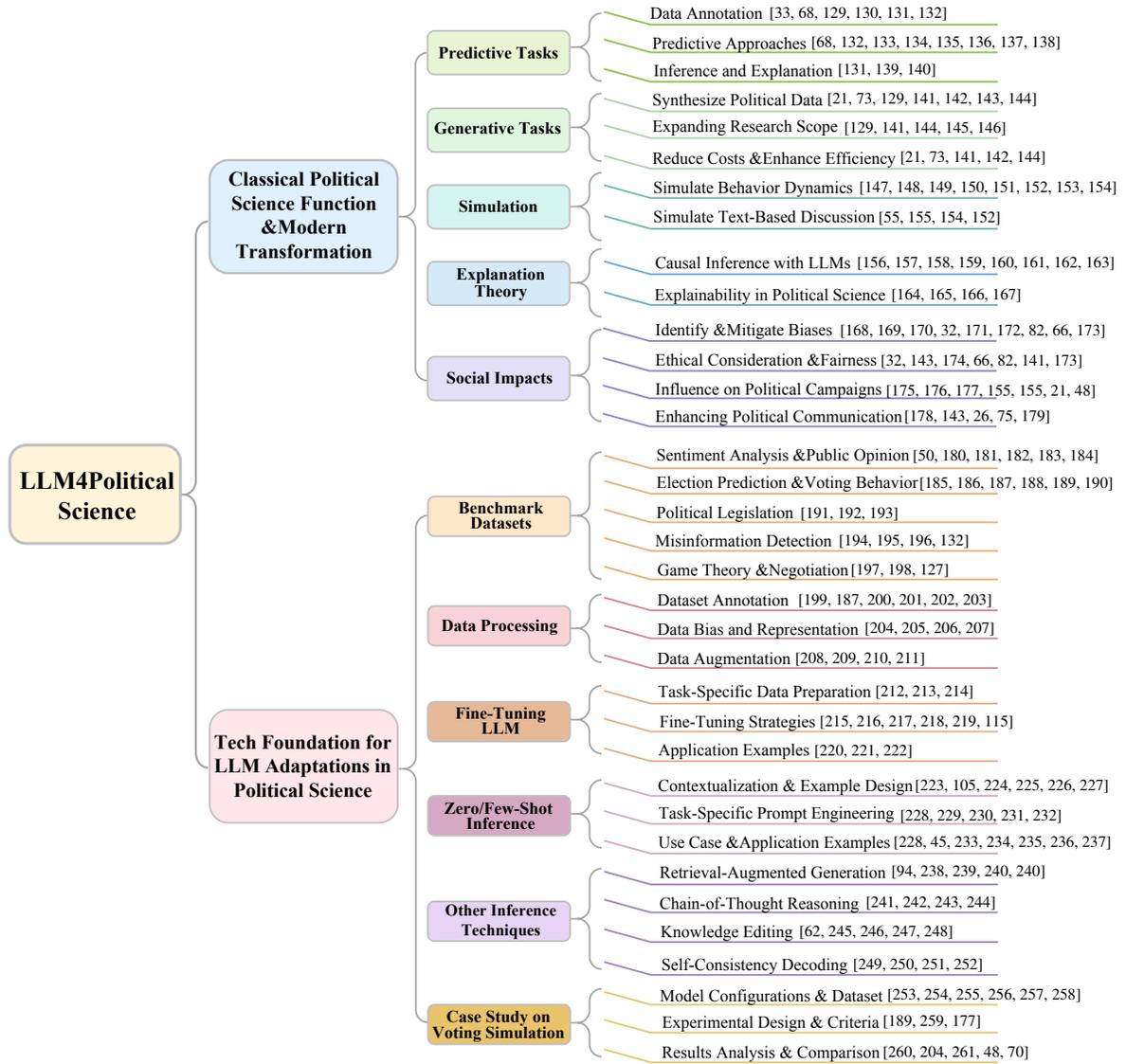
\begin{figure}[ht]
    \centering
    \resizebox{0.85\textwidth}{!}{
    \begin{overpic}{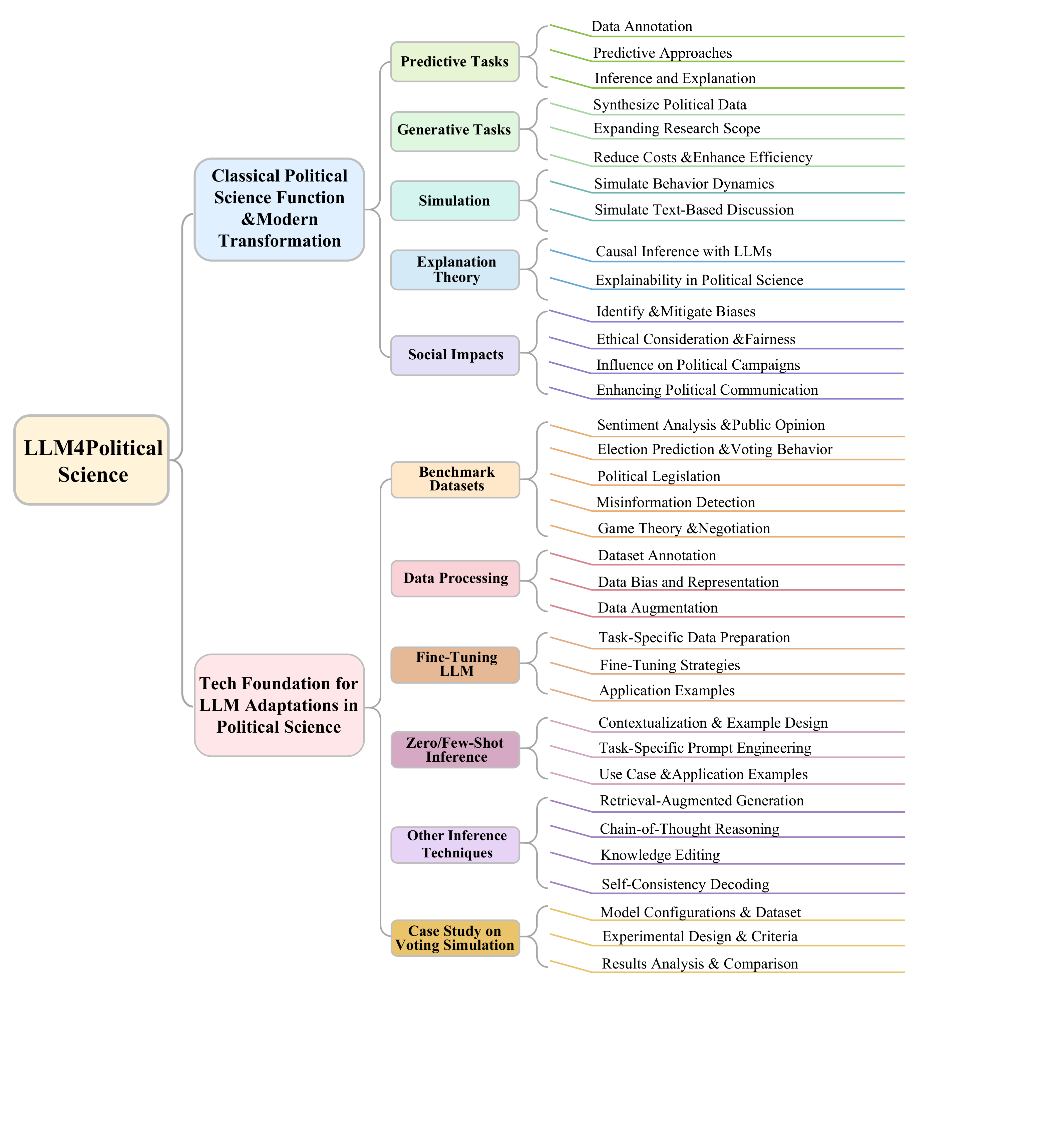} %Location of Your PDF Figurde file
        \put(71.6,98.5){\huge\cite{heseltine2024large, liu2024poliprompt,egami2024using, liu-etal-2022-politics,chalkidis_llama_2024, cao2024can}}  %insert citations on Specified coordiante
        \put(75.5,95.8){\huge\cite{liu2024poliprompt,cao2024can,gambini2024evaluating,wang_explainable_2024,wu_fake_2024,hu2024bad,whitehouse2022evaluation,kocielnik2023can}}
        \put(78,93.2){\huge \cite{chalkidis_llama_2024,lashitew2024corporate,fu2024deciphering}}
        \put(77.00,90.50){\huge\cite{argyle2023out,wu2023large,egami2024using,napolio2024executive,bisbee2024synthetic,alvarez2023generative,palmer2023large}}
        \put(78.6,87.8){\huge\cite{egami2024using,napolio2024executive,palmer2023large, alvarez2024evaluating, mellon2024ais}}
        \put(83.8,85){\huge\cite{argyle2023out,wu2023large,napolio2024executive,bisbee2024synthetic,palmer2023large}}
        \put(80,82.2){\huge\cite{park2023generative,gao2023large,wang2024survey,dai2024artificial,hua2023war,jin2024if,chuang2023simulating,guan2024richelieu}}
        \put(81.8,79.4){\huge\cite{baker2024simulating,moghimifar2024coalition,guan2024richelieu,jin2024if}}
        \put(79.8,75.2){\huge\cite{feder2022causal,ashwani_cause_2024, zevcevic2023causal, kiciman_causal_2024,li_prompting_2024, bhattacharjee_zero-shot_2024,gui_challenge_2023,wood-doughty_generating_2021}}
        \put(83,72.2){\huge\cite{zhao2024explainability,luo2024understanding,de2024use,dhawanend}}
        \put(78.2,68.8){\huge\cite{johnson2022ai,kim2023rise,lee2024large,tornberg2023chatgpt,stanczak2023quantifying,jiang2022communitylm,rozado2023political,motoki2024more,simmons2023moral}}
        \put(82,66.0){\huge\cite{tornberg2023chatgpt,alvarez2023generative,hackenburg2024microtarget,motoki2024more, rozado2023political, napolio2024executive, simmons2023moral}}
        \put(82.5,63.2){\huge\cite{bonikowski2022politics,foos2024use,yu2024will,moghimifar2024coalition,moghimifar2024coalition,argyle2023out,zhang2024electionsim}}
        \put(84.2,60.8){\huge\cite{argyle2023leveraging,alvarez2023generative,gover2023political,moghimifar2024modelling,ma2024chatgpt}}
        \put(85,57.1)
        {\huge\cite{santurkar2023whose,bastan-etal-2020-authors,chebolu2022survey,sharma2022fake,saha2022sentiment,waspodo2022indonesia}}
        \put(85.6,54.6){\huge\cite{DVN/ER9XTV_2022,DVN/IVIXLK_2022,DVN/ZFXEJU_2022,payne2010implicit,yu2024trumpwin2024predicting,DVN/42MVDX_2017}}
        \put(74.5,51.8){\huge\cite{kornilova2019billsum,shu2024lawllm,arregui2022new}}
        \put(78.1,49.2){\huge\cite{shu2020fakenewsnet,grover2022public,jin2017multimodal,cao2024can}}
        \put(79.2,46.5){\huge\cite{cunningham2013non,ari2023peace,meta2022human}}
        \put(74.3,43.7){\huge\cite{mochtak2023parlasent,DVN/ZFXEJU_2022,balloccu2024leak,rauniyar2023multi,tan2024large,ming2024autolabel}}
        \put(80.5,41){\huge\cite{qu2024performance,shahbazi2023representation,nakada2024synthetic,cloutier2023fine}}
        \put(74,38.2){\huge\cite{sahu2023promptmix,abaskohi2023lm,dos2024identifying,ding2024data}}
        \put(81.5,35.1){\huge\cite{zhang2024scaling,vm2024fine,jaradat2024multitask}}
        \put(76.3,32.3){\huge\cite{wu2024dlora,meloux2024novel,nabli2024acco,guan2024aptq,banerjee2023benchmarking,ding2023parameter}}
        \put(75.8,29.6){\huge\cite{petrova2020extracting,santosh2024lexsumm,guha2024legalbench}}
        \put(85.2,26.3){\huge\cite{wei2021finetuned,kojima2022large,kumar2023zero,di2024mapping,allaway2023zero,malladynamic}}
        \put(83.8,23.5){\huge\cite{kuila2024deciphering,hu2023synthesizing,burnham2024political,wahidur2024enhancing,yao2024more}}
        \put(83.2,21){\huge\cite{kuila2024deciphering,gujral2024can,ibrahim2024analyzing,kuntur2024under,pavlyshenko2024using,hu2024multi,chen2024fine}}
        \put(83,18.3){\huge\cite{salemi2024evaluating,wang2024evaluating,dong2024understand,arslan2024political,arslan2024political}}
        \put(80.6,15.4){\huge\cite{zhanggenerating,kareem2023fighting,dobrinoiu2024leveraging,tutunov2023can}}
        \put(74.4,12.7){\huge\cite{wang2023knowledge,zhang2024comprehensive,gupta2024stackfeed,zhang2024oneedit,peng2024event}}
        \put(79.4,9.6){\huge\cite{ahmed2023better,huang2023enhancing,cheng2024integrative,chen2023two}}
        \put(82.6,6.8){\huge\cite{islam2024gpt,rasheed2024taskcomplexity,chen2024magicdec,singh2024scidqa,ANES2016,kennedy2018evaluation}}
        \put(82.2,4.3){\huge\cite{yu2024trumpwin2024predicting,Argyle_Busby_Fulda_Gubler_Rytting_Wingate_2023,yu2024will}}
        \put(82.2,1.4){\huge\cite{qi2024representation,qu2024performance,majumdar2024generative,zhang2024electionsim,yang2024llm}}
    \end{overpic}
    }
    \caption{The proposed Taxonomy on LLM for Political Science.}
    %\Description{The proposed taxonomy of LLM for Political Science in this work.}
    \label{fig-taxonomy}
\end{figure}

Figure~\ref{fig-taxonomy} presents a comprehensive taxonomy for understanding the integration of LLMs in political science, organized into two main categories: classified political science functionalities \& tasks and LLM-driven computational approaches for political science. The first category highlights how traditional political science functions are enhanced through LLM capabilities, while the second category focuses on computational techniques for effectively implementing LLMs in political research.

The classified political science functional tasks span five primary categories, including Predictive Tasks, Generative Tasks, Simulation, Explainability \& Causal Inference, and Social Impacts. Predictive Tasks (section~\ref{4-1-predictive-tasks}) use LLMs to analyze and forecast trends in public opinion, electoral outcomes, and policy impacts. Generative Tasks (section~\ref{4-2-generative-task}) enable the synthesis of political data, such as summarizing legislative documents or generating debate transcripts~\cite{wagner2024power,von2023assessing}, expanding the scope and accessibility of political data analysis. Simulation (section~\ref{4-3-simulation}) leverages LLMs to model complex political behaviors and interactions~\cite{park2023generative,huang2024social}, reducing research costs and enhancing efficiency in studying dynamics like voting patterns and policy impact. Explainability \& Causal Inference (section~\ref{4-4-explainability}) applies LLMs to identify relationships and generate counterfactuals in political analysis~\cite{kiciman_causal_2024,li_prompting_2024}, providing insights into causality and potential biases. Finally, Ethical Concerns \& Social Impacts (section~\ref{values_societal_impacts}) analyzes the influence of LLM-driven applications for political campaigns and communication strategies, emphasizing the ethical considerations and public ramifications of political science research.

The LLM-driven computational approaches for political science consist of five components: Benchmark Datasets, Data Processing, Fine-Tuning LLMs for Political Science Tasks, LLM Inference under Zero/Few-Shot In-Context Learning Setting, Other LLM Inference Techniques, and Case Study on Voting Simulation. Benchmark Datasets (section~\ref{Benchmark}) provide foundational resources on political topics, while Data Processing (section~\ref{dataset-preparation}) addresses issues of bias, annotation, and augmentation to ensure data reliability and representativeness. Fine-tuning (section~\ref{fine-tuning}) explores methods to tailor LLMs for specific political science tasks, optimizing performance through targeted training strategies. Inference under Zero-Shot (section~\ref{zero-shot}) and Few-Shot Learning (section~\ref{few-shot}) highlight approaches for achieving task-specific insights with minimal labeled data, focusing on prompt engineering and example selection. Finally, Other Inference Techniques (section~\ref{other-inference}) such as retrieval-augmented generation~\cite{salemi2024evaluating}, chain-of-thought reasoning~\cite{ling2024deductive}, knowledge editing~\cite{chen2024can}, and self-consistency decoding~\cite{huang2023enhancing} enhance the adaptability of LLMs for nuanced political tasks. Together, these computational approaches construct a robust framework tailored for political science research.

\section{Classical Political Science Functions and Modern Transformations} \label{Traditional-Method}
%Original Title: Leveraging LLMs for Political Science Tasks: Traditional Meets Modern

LLMs have brought transformative changes to political science, reshaping traditional methodologies and unlocking new analytical opportunities. This section provides a structured overview of current research, categorizing it into five key areas. Four of these areas focus on the functional applications of LLMs in political science, while the fifth explores normative considerations, emphasizing societal and ethical implications. 

We divide the functional categories into predictive, generative, simulation, and explainable tasks. While computer science researchers often categorize LLM-based research into predictive and generative tasks \cite{demszky2023using, khurana2023natural, minaee2024large}, we propose two additional dimensions - simulation and causal inference, in order to address the unique complexity of LLM for Political Science.

While simulation is inherently generative, we distinguish it as a separate category due to its unique focus on replicating human-like attitudes, behaviors, and decision-making processes in specific political contexts. In this review, we list research that focuses on producing new content without emulating human cognitive processes as "generative tasks", and research mimicking how human actors or groups would react, taking into account motivations, biases, and contextual influences as "simulation". 

%Using LLMs to simulate human attitudes and behaviors is another promising application of LLM in political science. LLMs have been applied in simulating complex decision-making processes of political actors, governments, and voters \cite{chuang2023simulating, jin2024if,guan2024richelieu}. Such simulations allow political science researchers to test hypotheses in controlled, hypothetical environments, offering insights into how political decisions unfold and how various factors interact.

Additionally, political science is not merely concerned with making predictions but also with understanding the causes behind political phenomena. For instance, in addition to predicting the outcome of elections, political scientists are also interested in why certain outcomes occur. Therefore, using LLMs to support inference tasks (e.g., processing vast datasets, and identifying causal mechanisms) is promising in political science and a necessary addition to predictive and generative tasks.

\subsection{Automation of Predictive Tasks} \label{4-1-predictive-tasks}

\textbf{Definition.} Predictive tasks in Computational Political Science involve anticipating future events or trends based on existing data, and they are fundamental in political science for applications such as election forecasting, policy support prediction, and analyzing voter behavior. In political science, predictive tasks are crucial because they provide insights that can guide decision-making, inform policy, and help researchers understand complex social dynamics. Traditional predictive methods in political science often require extensive manual labor. For instance, certain predictive tasks may require researchers to manually collect survey responses, historical election data, or economic indicators, which can be time-consuming and prone to human error. In contrast, recent advancements in LLMs offer an alternative by automating predictive tasks. The automation of predictive tasks reduces manual effort and possible human error, while increasing speed, consistency, and scalability.

% \begin{wrapfigure}[24]{r}{0.47\textwidth}
% \centering
% \includegraphics[width=0.47\textwidth]{figures/sec4-1_Automated_predictive_task}
% \caption{The workflow of LLM-based automated predictive task, using the U.S. Presidential Election prediction as example.}
% \label{fig:predictive_task}
% \end{wrapfigure}

\textbf{Enhancing Prediction with LLM-based Data Annotation.} LLM-based automation significantly enhances predictive capabilities by providing consistent and scalable solutions for data-intensive tasks. This is especially helpful in data annotation. Annotating large datasets manually is time-consuming and prone to inconsistencies~\cite{heseltine2024large, liu2024poliprompt,egami2024using}. LLMs can rapidly process and annotate data in a consistent manner. Political science researchers have employed LLMs to annotate~\textit{Political Ideology}~\cite{heseltine2024large, liu-etal-2022-politics,chalkidis_llama_2024, cao2024can}, \textit{Fake News}~\cite{wang_explainable_2024,wu_fake_2024,hu2024bad,whitehouse2022evaluation}, \textit{Tone (sentiment)}~\cite{heseltine2024large, liu2024poliprompt, cao2024can,fu2024deciphering,lashitew2024corporate}, and content of various \textit{Political Texts}~\cite{heseltine2024large, liu2024poliprompt,kocielnik2023can,gambini2024evaluating,cao2024can}. Researchers also find that the quality of automated LLM annotation outperforms crowd workers and even some domain experts. Therefore, data annotation by LLM not only enhances efficiency but also reduces the potential for human bias and error in the data annotation process.

\textbf{Prediction Tasks in English-Speaking Contexts.} The effectiveness of LLMs in predictive tasks is demonstrated through their applications in both English and non-English contexts. In English-speaking settings, platforms like ChatGPT and Llama are frequently used for large-scale political text analysis. For instance, Lashitew and Mu~\cite{lashitew2024corporate} analyze comments and letters submitted by companies to the U.S. Securities and Exchange Commission regarding climate change disclosure regulations. Leveraging GPT-3, they efficiently process and analyze a large volume of text data, identifying patterns and sentiments within the corporate responses. Additionally,~\cite{fu2024deciphering} explore the application of GPT-4 in processing and analyzing public feedback collected online in New Zealand. They focus on responses to a proposed plan change in Hamilton City, New Zealand, assessing GPT-4's effectiveness in summarizing feedback, identifying topics, and analyzing sentiment. Results showed GPT-4 performed these tasks accurately.

\textbf{Predictive Tasks in Non-English Contexts.} LLMs have also shown robust performance in multilingual environments and in diverse regional applications.~\cite{heseltine2024large} evaluate the performance of GPT-4 in coding political texts across variables such as relevance, negativity, sentiment, and ideology across the United States, Chile, Germany, and Italy. The findings indicate that GPT-4's annotations closely align with those of human experts, suggesting that LLMs can effectively assist in political text analysis. Moreover, Chalkidis and Brandl~\cite{chalkidis_llama_2024} utilize Llama to evaluate speeches from European Parliament debates, with the EUandI questionnaire serving as a reference or benchmark to verify political leanings. The study demonstrated that Llama has considerable knowledge of national parties' positions and is capable of contextual reasoning as well as ChatGPT. Mellon et al.~\cite{mellon2024ais} take a step further to evaluate six different popular LLMs in categorizing open-text survey responses and detecting issue importance. Their task involved classifying the most important issue responses from the British Election Study Internet Panel into 50 distinct categories. The study concluded that LLMs, particularly Claude-1.3, can effectively code open-text survey responses, providing a scalable alternative to human coders.

\begin{wrapfigure}[25]{r}{0.47\textwidth}
\centering
\vspace{-2mm}
\includegraphics[width=0.47\textwidth]{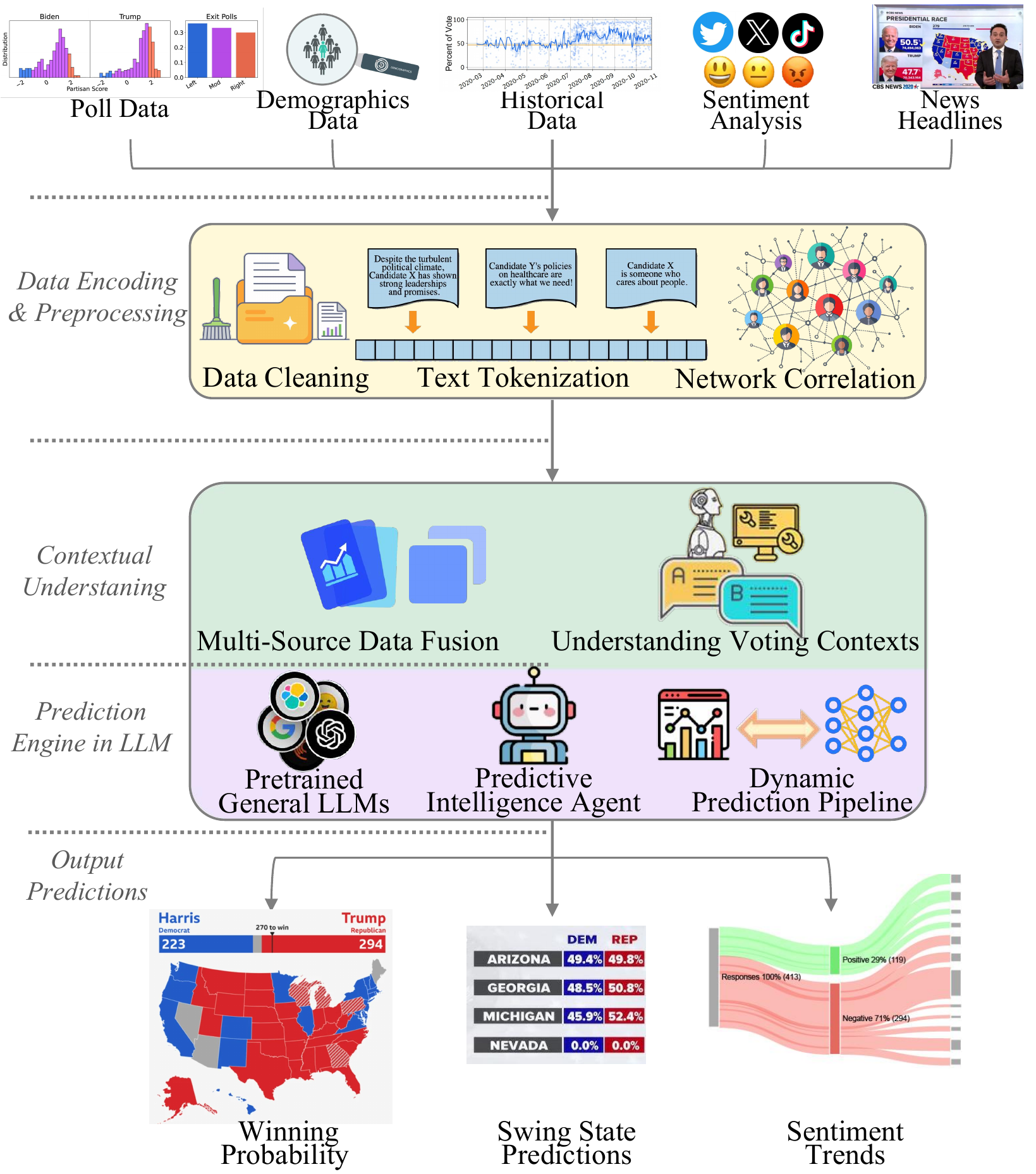}
\caption{The workflow of LLM-based automated predictive task, using the U.S. Presidential Election prediction as an example.}
\label{fig:predictive_task}
\end{wrapfigure}

\textbf{LLM-based Advancements.} To better illustrate the workflow of LLMs in predictive tasks, we provide a diagram showcasing the U.S. Presidential Election outcome prediction as an example. This example highlights how LLMs integrate diverse data sources, process them into structured representations, and generate actionable predictions. As shown in Figure~\ref{fig:predictive_task}, the workflow begins by integrating data sources like polls, demographics, social media sentiment, and news headlines. After preprocessing (cleaning, normalization, and vectorization), the LLM performs contextual understanding and generates outputs such as winning probabilities and swing state predictions, showcasing its ability to automate complex, data-driven tasks like U.S. election predictions.

In addition to predictive applications in various domains, recent research also highlights tailored frameworks and approaches developed specifically for political science. Such research often includes adjustments to LLMs to improve their applicability and accuracy in political contexts. For instance, PoliPrompt~\cite{liu2024poliprompt} is a three-stage framework leveraging LLMs for text classification in political science. This framework shows exceptional performance in classifying topics within multi-class news datasets, such as BBC news reports, labeling nuanced political science concepts, and analyzing the tones of campaign advertisements from the 2018 midterm election. This kind of tailored approach helps ensure that the model outputs are relevant and accurate within specific political frameworks. Similarly, by studying the classification on text alignment or opposition toward a particular issue, Cao and Drinkle~\cite{cao2024can} find that incorporating metadata (e.g., party affiliation) into political stance detection tasks can notably enhance model performance on ParlVote+ benchmark.

\textbf{Summary and Challenges.} The automation of predictive tasks by LLMs offers transformative potential in political science research~\cite{lazar2024can,linegar2023large}. From scaling data annotation processes to handling multilingual data and adapting to specific political frameworks, LLMs provide a powerful tool for researchers aiming to predict and analyze trends in political behavior and sentiment, as well as test political theories.
However, existing research still faces notable challenges. LLMs can sometimes lack contextual understanding in nuanced political discourse, particularly in multilingual or culturally specific settings, where subtle language differences may lead to misinterpretation~\cite{he2023inducing}. Additionally, the reliance on pre-existing data and the potential for inherent biases in training datasets can result in biased predictions, impacting the accuracy and neutrality of the automated outputs~\cite{bang2024measuring}. Furthermore, while LLMs have shown proficiency in annotation and classification, their performance may degrade when faced with highly complex or specialized political tasks that require deeper domain knowledge~\cite{wu2023large}. Addressing these limitations is essential for maximizing the utility and reliability of LLMs in political science applications.

\subsection{Automation of Generative Tasks} \label{4-2-generative-task}

\textbf{Definition.} Generative tasks in political science involve creating synthetic data, simulating scenarios, or augmenting incomplete datasets, offering new insights where traditional data sources are either unavailable or insufficient~\cite{argyle2023out, bisbee2024synthetic, wu2023large, napolio2024executive}. Unlike analytical tasks that focus on interpreting existing information, generative tasks expand the boundaries of what can be studied by creating representations of missing data or by projecting possible future scenarios~\cite{argyle2023out, bisbee2024synthetic, wu2023large}. Generative tasks are particularly valuable for political science applications where the complete dataset is often hard to obtain due to privacy concerns, logistical constraints, or high costs associated with traditional data collection methodologies~\cite{napolio2024executive, palmer2023large}.

The absence of complete data often underscores the complexity, scope, and depth of political science research questions~\cite{argyle2023out, bisbee2024synthetic, wu2023large, napolio2024executive}. For example, understanding the roles and performance of executive agencies, which exert significant influence over policy, presents substantial challenges due to the limited availability of data \cite{napolio2024executive}. Traditional CPS approaches, such as principal component analysis (PCA)-based methods, demand extensive input data, limiting their application in issue-specific analyses, such as polarizing topics like abortion or gun control. LLMs are capable of extracting valuable insights from incomplete datasets if provided with well-structured prompts, broadening the analytical capacity of studies~\cite{argyle2023out, bisbee2024synthetic, wu2023large}. This innovation enables the exploration of previously constrained research areas \cite{napolio2024executive, palmer2023large}. Existing research in this domain can be grouped into two major categories: \textit{Synthesizing Political Data} and \textit{Enhancing Research Scope}.

\begin{wrapfigure}[18]{r}{0.47\textwidth} 
\centering
\vspace{-4mm}
\includegraphics[width=0.47\textwidth]{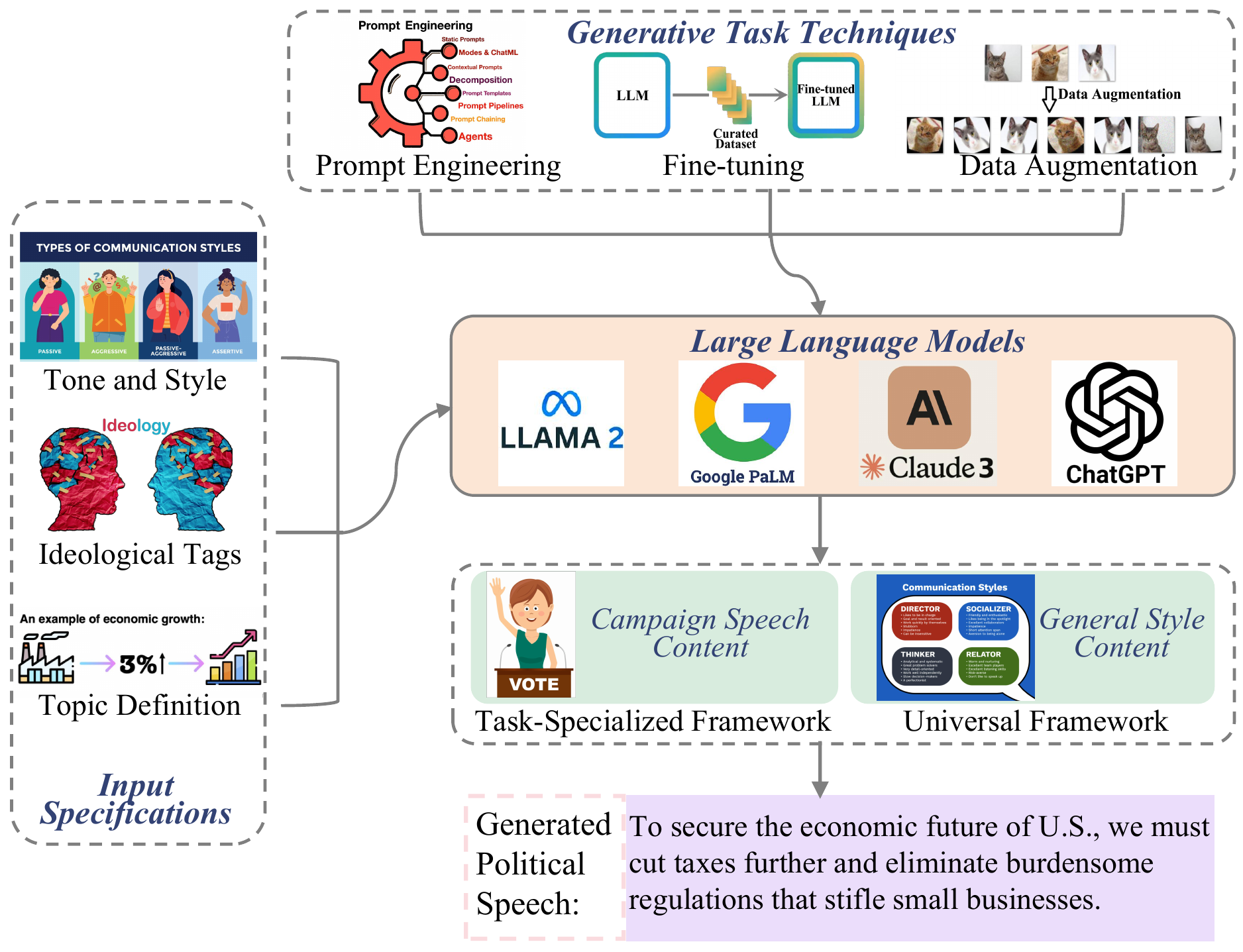}
\caption{Workflow for LLM-based generative tasks, illustrating the synthesis of political speeches with specific ideology, style, and focus of content.}
\label{fig:generative_task}
\end{wrapfigure}

\noindent \textbf{Synthesizing Political Data.} The ability to generate synthetic data is a powerful application that directly addresses the critical issue of data scarcity and facilitates the exploration of latent variables. Data collection is often a significant hurdle in political science due to the costs and time involved in conducting surveys, gathering reliable public opinion data, or accessing confidential voting records. Synthetic data generation by LLMs offers an efficient, cost-effective alternative that can serve as a proxy for real-world data, providing insights where traditional data sources are limited~\cite{wang2024twin}. For instance, Bisbee et al.~\cite{bisbee2024synthetic} demonstrate that LLM-generated synthetic data can effectively replicate survey responses, simulating various public opinion trends even in the absence of comprehensive survey datasets. They successfully explore public sentiment on immigration, healthcare, and  climate policy issues. This application is particularly useful for analyzing time-sensitive political questions, where delays in data collection could mean losing valuable insights into changing public opinion. Another noteworthy study comes from Argyle et al.~\cite{argyle2023out}, who show that LLMs can simulate human responses, mimicking the distribution of survey data across demographic groups and regions. In this case, LLMs help mitigate the data scarcity issue by generating synthetic samples that reflect genuine population characteristics, supporting research on political trends in underserved or underrepresented communities. We provide the workflow of LLM-based generative tasks in Figure~\ref{fig:generative_task}, using the synthesis of political speeches or manifestos as an example. Starting with inputs like topic definitions, ideological tags, and tone preferences, the model preprocesses and contextualizes data to generate coherent outputs. Techniques such as prompt engineering and fine-tuning guide the process. The outputs, including political speeches tailored to ideological perspectives, demonstrate how LLMs can address challenges of data scarcity and enable synthetic data generation for political research.

LLMs also play a critical role in estimating political ideologies in situations where conventional data sources, such as voting records, media publications, or public statements, are incomplete. Wu et al.~\cite{wu2023large} illustrate how LLMs can infer political ideologies by analyzing existing contextual information and filling in missing details, thereby offering a fuller, more nuanced picture of the ideological spectrum in specific political landscapes. Moreover, Alvarez et al.~\cite{alvarez2023generative} explore the potential of LLMs in simulating voter behavior and party strategies, thus extending traditional political modeling frameworks. By generating synthetic data that represents hypothetical voter responses to specific policies or campaign strategies, LLMs help researchers examine potential outcomes in elections or other political events. Such applications offer new avenues for understanding the impact of political campaigns and policy proposals, even when comprehensive polling data is unavailable.

LLMs further enhance research potential by enabling the generation of large and dynamic datasets that track the latest political trends over time.~\cite{palmer2023large} emphasize the utility of LLMs in constructing extensive synthetic datasets by generating responses or synthesizing textual data. This enables the analysis of long-term shifts in public opinion or political rhetoric across diverse populations.

\noindent \textbf{Enhancing Research Scope.} Beyond data synthesis, LLMs enable researchers to explore previously unattainable research areas by providing insights into complex or hard-to-measure variables. This capacity to expand the scope of political science research is especially valuable in analyzing intricate social dynamics, government policies, and ideological nuances where data gaps often hinder rigorous analysis. For example, Napolio's~\cite{napolio2024executive} work on the ideological positioning of executive agencies illustrates how LLMs can provide insights into policy stances and organizational biases even in the absence of direct, comprehensive data. The use of LLMs to fill data gaps allows for a deeper understanding of government operations and policy influences that would otherwise remain hidden. Similarly, Egami et al.~\cite{egami2024using} demonstrate that LLMs can work with imperfect or noisy data, producing robust analytical results even when complete datasets are unavailable. This flexibility reduces dependency on high-quality data and supports rigorous analysis in fields like public policy and election studies, where data completeness is often challenging to achieve.

LLMs are also adept at analyzing extensive political text corpora, which enables researchers to uncover subtle patterns in discourse that are difficult to capture through traditional manual analysis. Palmer and Spirling~\cite{palmer2023large} highlights the ability of LLMs to process large volumes of text, revealing shifts in political narratives and public sentiment over time. Similarly, the use of LLMs to analyze political Q\&A sessions in~\cite{alvarez2024evaluating} shows how these models can detect nuances in rhetoric and speaker intent, providing valuable insights into the subtleties of political communication. Furthermore, Mellon et al.~\cite{mellon2024ais} showcase the utility of LLMs in coding open-ended survey responses at scale. This application allows researchers to classify responses efficiently, identifying dominant issues and sentiments within a population. By automating the analysis of qualitative data, LLMs offer a powerful solution for understanding public concerns and policy impacts, contributing to a more comprehensive understanding of societal dynamics.

%The zero-shot capabilities of LLMs also support research areas where labeled data is unavailable, as demonstrated by Wu et al.~\cite{wu2023large}, who show that LLMs can estimate political ideologies without pre-labeled training data. This capacity enables political scientists to analyze new datasets without the need for extensive data preparation. The context-aware learning capabilities of LLMs, as observed by Bosley et al.~\cite{bosley2023we}, further enhance their utility in political science by facilitating the detection of hidden connections in complex political narratives, allowing for more in-depth and nuanced analyses.

\textbf{Summary and Challenges.} LLMs have reshaped the field of generative tasks in political science, enabling new possibilities in data synthesis and research scopes. These models provide political scientists with the tools needed to address data scarcity issues, create realistic proxies for hard-to-collect data, and simulate complex political phenomena. However, the challenges in ensuring the validity, neutrality, and reliability of synthetic data remain significant. Biases embedded in LLM-generated data can potentially skew results if not rigorously managed, and reliance on synthetic data requires careful validation to ensure accuracy. Moreover, while LLMs are proficient in generating insights, the interpretability of these models in highly nuanced political contexts remains a challenge. Addressing these limitations will be essential for leveraging the full potential of LLMs in generative political science research.

\subsection{Simulation of LLM Agents}\label{4-3-simulation}

\textbf{Definitions.} The concept of Simulation Agents in LLM for political science refers to the use of large language models to create interactive environments in which autonomous agents simulate behaviors, decisions, or dialogues. These tasks aim to explore dynamic systems, such as political behaviors, negotiations, or conflicts, by modeling interactions between agents. While both Generative Tasks and Simulation Agents leverage LLMs, their objectives and methodologies are distinct. Generative tasks focus on creating new data or textual content to address data scarcity, enabling researchers to fill gaps or produce synthetic datasets for foundational analysis. In contrast, simulation agents emphasize modeling interactions and dynamics within complex environments, offering insights into strategies, behaviors, and evolving systems. We present a comprehensive comparison between these tasks in Table~\ref{tab:task_comparison}.

\begin{table}[!htbp]
\centering
\footnotesize
\setlength{\tabcolsep}{5pt}
\renewcommand{\arraystretch}{0.9}
\caption{Comparison of Generative Tasks and Simulation Tasks in Political Science}
\vspace{0.5mm}
\label{tab:task_comparison}
\begin{tabular}{>{\centering\arraybackslash}p{3cm} p{6cm} p{6cm}}
\toprule
\textbf{Key Attribute}  &\multicolumn{1}{c}{\textbf{Generative Tasks}} & \multicolumn{1}{c}{\textbf{Simulation Tasks}}  \\ \midrule
{\textit{Objective}}        & Create new data or textual content to address data scarcity.           & Model interactions and dynamics within complex environments.          \\ \midrule
{\textit{Focus}}         & Producing synthetic data or content for foundational analysis.         & Exploring strategies, behaviors, or evolving systems through agents.  \\ \midrule
{\textit{Output}}          & Independent generated results, such as datasets or textual outputs.    & Analytical results on interactions, strategies, or behavior patterns. \\ \midrule
{\textit{Research Context}}   & Filling data gaps and enhancing data availability.                     & Studying dynamic processes and agent interactions.                    \\ \midrule
{\textit{Methodology}}        & Generative models producing outputs based on prompts.                  & Simulations of agents interacting within predefined environments.      \\ \midrule
{\textit{Application Examples}} & Synthetic survey data, opinion generation, or text classification.    & Negotiation models, conflict dynamics, or opinion shift simulations.  \\ \bottomrule
\end{tabular}
\end{table}

The use of LLMs to simulate human-like behavior in interactive environments represents a significant advancement in political science~\cite{park2023generative}. These simulations offer new ways to address complex societal questions, particularly those involving the behavior of political actors in intricate environments~\cite{de2014agent}. Traditional methods, such as Agent-Based Models (ABMs)~\cite{de2005computational}, rely on predefined parameters and restricted environments, often limiting their capacity to capture the complexity and realism of political dynamics. LLMs overcome these constraints by using natural language prompts to define behavior rules and environmental contexts~\cite{gao2023large}, allowing for adaptive, context-sensitive, and personalized agent behaviors~\cite{wang2024survey}. Current research in this area is focused primarily on two applications: (1) \textit{using agents to simulate behavior dynamics} and (2) \textit{using agents to simulate text-based discussion processes}~\cite{guan2024richelieu, moghimifar2024coalition, wang2024survey}.

\textbf{Simulate Behavior Dynamics.} Recent studies demonstrate the potential of LLMs to replicate complex social behaviors in political settings, addressing limitations in traditional ABM approaches. Dai et al.~\cite{dai2024artificial} simulate agents shifting from conflict to cooperation in resource-constrained environments through Hobbesian Social Contract Theory, exploring how political entities navigate scarcity and develop governance structures. Hua et al.~\cite{hua2023war} take a historical approach, modeling strategic decision-making during major global conflicts such as the World Wars, focusing on the interplay between diplomacy and military tactics in the evolution of warfare. Jin et al.~\cite{jin2024if} extend these simulations to a cosmic scale, where agents with distinct worldviews engage in cooperation and conflict, highlighting how ideological divergence influences inter-civilization dynamics. Other research builds on these approaches by introducing more nuanced political scenarios. Chuang et al.~\cite{chuang2023simulating} simulate opinion dynamics within political networks, where agents adjust their beliefs based on interactions with other agents, providing a closer examination of political polarization and consensus-building processes. Similarly, Guan et al.~\cite{guan2024richelieu} use LLM-based agents to model AI diplomacy, where agents negotiate and evolve their strategies in complex international relations, mirroring real-world diplomatic negotiations. These studies collectively showcase how LLM-driven simulations of behavior dynamics can provide valuable insights into governance, conflict resolution, and social interaction, offering novel ways to study political and diplomatic behavior in various contexts~\cite{dai2024artificial,guan2024richelieu,yao2024comal,jin2024if,chuang2023simulating,hua2023war}.

\noindent \textbf{Simulate Text-based Discussion.} Shifting from physical to text-based simulations, recent studies have explored political interactions through dialogue, using LLM agents to simulate complex discussions and negotiations. Baker et al.~\cite{baker2024simulating} model U.S. Senate policy debates, where LLM agents simulate legislative decision-making and bipartisanship, providing insights into how political actors navigate ideological divides and negotiate policy outcomes. Moghimifar et al.~\cite{moghimifar2024coalition} take a different approach by simulating multi-party coalition negotiations using LLM-driven dialogue. Their work highlights the intricacies of building and maintaining political alliances through textual interaction, emphasizing how agents negotiate, compromise, and form agreements in multi-party systems. Guan et al.~\cite{guan2024richelieu} extend this approach to international diplomacy, focusing on how LLM agents evolve strategies in alliance-building and negotiation on the global stage. Their research underscores the dynamic nature of diplomatic discourse, where agents adapt to shifting geopolitical contexts and evolving relationships between states. Additionally, Jin et al.~\cite{jin2024if} explore the use of LLMs in simulating text-based discussions between civilizations with divergent worldviews, pushing the boundaries of how text-based interactions can simulate inter-group communication and conflict resolution on a cosmic scale. These studies collectively illustrate the capacity of LLM simulations to model political decision-making through textual interactions, offering a contrast to action-oriented simulations like those seen in warfare and conflict resolution~\cite{baker2024simulating, chuang2023simulating, moghimifar2024coalition, guan2024richelieu, jin2024if}.

\noindent \textbf{Summary and Challenges.} LLM-driven simulations provide a novel framework for exploring the complexity of political behavior and interactions by enabling adaptive, context-aware modeling that was previously unattainable with traditional methods. These simulations bridge gaps in understanding how dynamic processes, such as opinion shifts, negotiation strategies, and conflict resolution, evolve under different political scenarios. Despite these advancements, significant challenges persist. Ensuring the neutrality of simulations remains difficult due to biases inherent in LLM training datasets, which can skew outcomes and interpretations. Moreover, ethical concerns arise when simulations replicate sensitive behaviors or policy decisions, potentially influencing real-world political discourse. Last but not least, the computational costs of running large-scale simulations can limit accessibility for many researchers. Addressing these issues will require robust validation techniques, interdisciplinary collaboration, and ongoing innovation to ensure that LLM simulations remain reliable and ethically sound tools for political science research.

\subsection{LLM Explainability and Causal Inference} \label{4-4-explainability}

\textbf{Definition of Explainability.} Explainability in the context of LLMs refers to the ability to provide interpretable and understandable outputs that clarify how and why specific predictions or decisions are made. In politically sensitive applications, explainability ensures that stakeholders can trace model outputs to underlying reasoning processes, fostering trust and transparency. Interpretability is critical for validating insights derived from LLM analyses and ensuring fairness in decision-making for political science.

\textbf{Definition of Causal Inference.} Causal Inference is the process of identifying and understanding cause-and-effect relationships between variables. It goes beyond correlation by attempting to answer questions like "What caused this outcome?" or "What would happen if a specific intervention were applied?" In political science, causal inference is central to assessing the impact of policies, understanding voter behavior, and analyzing societal dynamics. LLMs offer new opportunities for causal inference tasks, enabling researchers to detect patterns, identify potential causal relationships, and generate counter-factuals.

One of the ultimate goals of science is to \textit{explain} phenomena and uncover cause-and-effect relationships. In political science, causal inference plays a crucial role in understanding the impact of policies, campaigns, and social dynamics~\cite{feder2022causal,ashwani_cause_2024, zevcevic2023causal, kiciman_causal_2024}. While causal inference has been a focus in social and medical sciences~\cite{feder2022causal}, it has received comparatively less attention in computer science~\cite{feder2022causal}. LLMs, with their remarkable capabilities in language generation and pattern recognition, provide new tools for enhancing causal inference. However, they also face significant limitations in moving beyond correlation to meaningful causal reasoning~\cite{bagheri_c2p_2024}. These challenges hinder their ability to provide deeper explanations of the phenomena they analyze, an essential requirement for advancing scientific understanding. Despite these limitations, recent research highlights the potential of LLMs to support causal inference-related tasks, providing tools for researchers to explore cause-and-effect relationships in innovative ways.

\noindent \textbf{Explainability of LLMs in Political Science.} The explainability of LLMs, referring to the ability to generate interpretable insights, directly impacts their utility in causal inference~\cite{zhao2024explainability}. Researchers can leverage explainability tools, such as attention mechanisms~\cite{luo2024understanding} and prompt engineering~\cite{de2024use}, to identify relevant variables and interactions within data. For instance, post-hoc analysis methods~\cite{dhawanend} enable researchers to interpret why an LLM has generated specific outputs, facilitating the identification of potential causal pathways in text-based datasets. This capability enhances the transparency and reliability of LLM-driven causal analysis, especially in politically sensitive contexts.

\noindent \textbf{Applications of LLM Causal Inference.} Recent advancements demonstrate the potential of LLMs in identifying and modeling causal relationships. For example, LLMs have been utilized to detect causal patterns within large datasets, uncovering complex dependencies that traditional methods might overlook~\cite{kiciman_causal_2024, jiralerspong_efficient_2024}. By combining LLMs with domain expertise, researchers can identify key variables and interactions more effectively, leading to robust causal models. Another key application is the generation of counterfactual scenarios, which explore hypothetical outcomes under alternative conditions. LLMs can generate counterfactuals to assess the impact of different political policies or interventions, providing researchers with tools to test "what-if" scenarios~\cite{ashwani_cause_2024, li_prompting_2024, bhattacharjee_zero-shot_2024}. Furthermore, LLMs have been employed to identify necessary and sufficient causes in controlled experimental settings, offering insights into the factors driving specific outcomes~\cite{kiciman_causal_2024}. LLMs can also generate simulated datasets to evaluate causal inference methods. For instance, Gui and Toubi~\cite{gui_challenge_2023} used LLMs to conduct between-subject experiments to assess treatment effects, while Wood et al.~\cite{wood-doughty_generating_2021} evaluated causal estimation methods like propensity score matching and inverse propensity weighting using GPT-generated text data. These studies highlight the versatility of LLMs in creating experimental environments for testing and validating causal inference techniques.

\noindent \textbf{Strengths and Limitations of LLMs in Causal Inference.} One advantage of LLMs is their immunity to the carryover effect~\cite{batzdorfer2024conspiracy}, which is a confounding factor in sequential human experiments, since each interaction can be reset to a neutral state. This feature allows researchers to isolate treatment effects without prior interactions influencing subsequent results~\cite{gui_challenge_2023}. Additionally, LLMs possess human-like abilities to generate causal graphs or extract background causal context from natural language~\cite{vashishtha2023causal}, expanding their utility in tasks traditionally requiring human expertise~\cite{kiciman_causal_2024}. However, the limitations of LLMs in causal inference remain significant. Critics argue that LLMs, trained on data where causal knowledge is embedded, often act as "causal parrots," merely reciting learned patterns without true causal understanding~\cite{zevcevic2023causal}. This lack of genuine reasoning raises concerns about over-reliance on LLM-generated causal insights. Moreover, unpredictable failure modes—such as inconsistencies when questions are framed differently—undermine the reliability of LLMs for causal inference tasks~\cite{kiciman_causal_2024}. Ethical concerns, including potential misuse of causal findings in politically sensitive contexts~\cite{ashwani2024cause}, further complicate their deployment.

\noindent \textbf{Summary and Challenges.} LLMs hold significant promise for advancing causal inference in political science by enabling researchers to identify causal relationships, generate counterfactual scenarios, and simulate experimental data. Their unique capabilities, such as immunity to carryover effects and the ability to model causal contexts, make them valuable tools for exploring cause-and-effect relationships. However, limitations such as biases, inconsistent outputs, and ethical challenges must be addressed to ensure their effective application. As research continues, LLMs are poised to play an increasingly critical role in enhancing explainability and causal inference in the social sciences.

\subsection{Ethical Concerns in LLM Development and Deployment}
\label{values_societal_impacts}
\noindent \textbf{General Concerns About Embedded Values in LLMs.}
Large language models are increasingly influencing societal and political discourse, raising fundamental questions about the values and biases they embed. The design and deployment of LLMs often involve implicit decisions about whose perspectives and moral frameworks are represented, potentially shaping public perception and decision-making in ways that are not always transparent. Johnson and Iziev~\cite{johnson2022ai} highlight the ethical dilemmas surrounding trust in AI-generated content, emphasizing the difficulty in ensuring that LLMs align with societal norms while avoiding the reinforcement of harmful biases. Similarly, Kim and Lee~\cite{kim2023rise} examine the implications of LLM-driven conversational agents in political campaigns, noting the potential for these tools to inadvertently promote specific ideologies under the guise of neutrality. Lee et al.~\cite{lee2024large} further explore how LLMs reflect and propagate structural societal biases, particularly those affecting subordinate social groups. The study reveals that LLMs tend to portray these groups as more homogeneous, aligning with longstanding human cognitive biases, and underscores the importance of addressing such systemic issues in model training and evaluation. As LLMs continue to integrate into decision-making systems and public-facing applications, understanding their embedded values becomes imperative. This broad analysis sets the stage for more focused discussions on specific biases and potential mitigation strategies in subsequent sections.

\noindent \textbf{Specific Manifestations of Biases and Preferences in LLM Outputs.}
The outputs of LLMs often reflect biases and preferences that manifest in specific, measurable ways, influencing how these models are perceived and utilized across different contexts. These manifestations not only reveal the underlying training data biases but also highlight the importance of careful model deployment. For instance, Tornberg~\cite{tornberg2023chatgpt} provides a comprehensive analysis of ChatGPT's language use, showing how the model tends to favor Western-centric cultural norms and professional jargon. This skew has implications for accessibility and inclusivity, as it may alienate users from non-Western backgrounds or those with varying levels of language proficiency. In addition, Stanczak et al.~\cite{stanczak2023quantifying} introduce a framework for quantifying biases in LLM outputs, with a focus on gender and occupational stereotypes. The study demonstrates that despite improvements in reducing overtly biased outputs, subtle biases persist, particularly in contexts where societal norms conflict with the training data distribution. Jiang et al.~\cite{jiang2022communitylm} also investigate how LLMs trained on community-specific data exhibit distinct preferences that align closely with the values and norms of those communities. While this approach can increase relevance for specific audiences, it raises concerns about the potential for reinforcing echo chambers and ideological polarization when these models are used in broader contexts. The findings collectively illustrate the challenges of mitigating biases in LLM outputs, calling for more robust evaluation mechanisms and the inclusion of diverse training data to minimize the risk of harmful stereotypes or cultural insensitivity.

\noindent \textbf{Practical Strategies for Mitigating Biases in LLMs.} 
Efforts to address the biases embedded in LLMs have led to the development of various practical strategies. These approaches aim to minimize the harm caused by biased outputs while maintaining the utility of the models in diverse contexts. Recent studies provide valuable insights into how such strategies can be implemented effectively. Rozado~\cite{rozado2023political} emphasizes the importance of balancing ideological representations within LLMs to mitigate political biases. The study outlines a method of systematically curating training datasets to ensure parity in the representation of diverse viewpoints. This proactive approach not only reduces overt political biases but also fosters fairness in politically sensitive applications, such as journalism and policymaking. Building on this, Motoki et al.~\cite{motoki2024more} highlight the role of iterative fine-tuning using diverse feedback sources. By incorporating user feedback from underrepresented communities, LLMs can better align with a broader range of cultural norms and values. The findings in~\cite{motoki2024more} suggest that this dynamic feedback loop significantly enhances model responsiveness to marginalized perspectives, making it a crucial step in real-world deployments. Simmons~\cite{simmons2023moral} takes a complementary approach by advocating for embedding explicit moral reasoning frameworks into LLM training pipelines. This strategy involves integrating ethical guidelines and decision-making frameworks into the model’s architecture. Simmons argues that such measures not only mitigate biases but also equip models with the capacity to navigate morally ambiguous scenarios, thereby improving trustworthiness in high-stakes applications. These efforts demonstrate that mitigating biases in LLMs is both technically achievable and ethically essential.

\noindent \textbf{Broader Societal Implications of LLM Biases.}  
The biases embedded in LLMs extend beyond technical and academic concerns, influencing societal structures and interactions in profound ways. As LLMs become increasingly integrated into decision-making processes, communication platforms, and personalized services, understanding their broader societal impacts is critical. Scholar like Tornberg~\cite{tornberg2023chatgpt} highlights how biases in LLMs can perpetuate existing social inequalities by reinforcing dominant narratives. The study examines ChatGPT’s performance in generating culturally sensitive responses, revealing disparities in the model’s treatment of various sociocultural groups. Tornberg argues that such imbalances risk entrenching systemic inequities, especially when LLMs are used in education, public discourse, and policymaking. Alvarez et al.~\cite{alvarez2023generative} complement this analysis by exploring the role of generative AI in amplifying misinformation and political polarization. The study discusses how LLMs, if left unchecked, can contribute to the spread of ideologically skewed content, potentially exacerbating societal divisions. Alvarez emphasizes that biases in LLM outputs are not isolated technical flaws but are deeply intertwined with broader societal challenges, such as media manipulation and the erosion of public trust. Hackenburg and Margetts~\cite{hackenburg2024microtarget} extend these concerns to the realm of targeted advertising and political microtargeting. This study illustrates how biased LLMs can be leveraged to craft persuasive narratives tailored to specific demographics, raising ethical questions about manipulation and autonomy. Hackenburg warns that the misuse of biased language models in these contexts may deepen socioeconomic disparities and influence political outcomes in undemocratic ways. These studies highlight the importance of designing LLMs that are fair, transparent, and inclusive, particularly as they are increasingly applied in sensitive domains like political analysis and social sciences.

\noindent \textbf{Summary and Challenges.}
The intersection of LLMs, societal values, and biases presents a complex but essential area of study. While advancements in LLMs enable transformative applications, their inherent biases pose significant ethical challenges. Addressing these challenges requires: 
\begin{itemize}[leftmargin=*]
\item {\textit{Awareness}}: Achieving a deeper understanding of how biases manifest in LLM outputs.  
\item {\textit{Accountability}}: Aligning LLMs with diverse societal needs under common ethical standards and guidelines.
% Ensuring that developers and researchers adhere to ethical standards and guidelines to align LLMs with diverse societal needs.  
\item {\textit{Transparency}}: Building methods for identifying, monitoring, and mitigating biases in real-world applications.  
\end{itemize}
Future research must prioritize creating robust methodologies for bias mitigation, with a focus on enhancing fairness, inclusivity, and accountability in LLM development and deployment~\cite{motoki2024more, rozado2023political, napolio2024executive, simmons2023moral}.

\subsection {Societal Impacts} \label{4-5-social-impact}

\noindent \textbf{Definitions and Context.} The societal impacts of political-LLM sphere extend beyond technical concerns to encompass profound ethical, communicative, and informational implications. From influencing election outcomes to enhancing political communication, LLMs hold the potential to transform the societal landscape in both positive and negative ways. This section explores the multifaceted effects of LLMs on political campaigns, public communication, and civic engagement, while addressing potential risks and ethical challenges.

\noindent \textbf{Transforming Political Campaigns.} LLMs have revolutionized the way political campaigns are conducted by enabling hyper-personalized messaging and voter targeting~\cite{bonikowski2022politics,hackenburg2024microtarget,moghimifar2024coalition,foos2024use,yu2024will}.~\cite{bonikowski2022politics} is an early work which highlights the potential of LLMs in measuring populism, nationalism, and authoritarianism through automated analysis of U.S. presidential debates. Hackenburg~\cite{hackenburg2024microtarget} demonstrates how LLMs can analyze large datasets to generate messages tailored to individual voter profiles, influencing voter perceptions and potentially altering election outcomes. Beyond voter engagement, LLMs play a strategic role in shaping campaign narratives that resonate with diverse audiences. Moghimifar et al.\cite{moghimifar2024coalition} show that LLM-based agents can model political coalition negotiations, providing insights into political alliances and enabling more dynamic campaign strategies. Foos~\cite{foos2024use} discusses how generative AI tools, including LLMs, are transforming election campaigns by facilitating AI-to-voter conversations and enabling scalable, multilingual interactions under diverse democracies. Lately, Yu et al.~\cite{yu2024will} propose a novel multi-step reasoning framework using LLMs for U.S. election predictions, incorporating time-sensitive factors like candidates’ policies and demographic trends to enhance accuracy. Together, these works showcase the multifaceted capabilities of LLMs in modernizing political campaigns and amplifying their impact across various dimensions.

\noindent \textbf{Enhancing Political Communication}
In an era of increasingly complex political discourses, LLMs offer tools to bridge the gap between policymakers and the public~\cite{argyle2023leveraging,alvarez2023generative,gover2023political,moghimifar2024modelling,ma2024chatgpt}. By simplifying intricate political and legislative content, LLMs make critical information more accessible to citizens, fostering greater political understanding and participation. Argyle et al.~\cite{argyle2023leveraging} discuss how LLMs can distill party manifestos into understandable summaries, addressing barriers that often hinder public engagement. Similarly, Alvarez et al.~\cite{alvarez2023generative} highlight the potential of generative AI to enhance transparency and comprehension in elections, allowing voters to make more informed decisions. These advancements suggest that LLMs could play a pivotal role in democratizing information and improving the accessibility of political communication.

\noindent \textbf{Democratizing Information Access.}
LLMs hold the promise of empowering individuals by breaking down complex topics into easily understandable language, thereby democratizing access to information. This capability can foster a more informed citizenry and enable greater accountability among political actors. By providing equitable access to political knowledge, LLMs ensure that more people, regardless of educational background, can participate in democratic processes. For instance, LLMs can assist in translating political jargon or simplifying policy discussions, helping individuals navigate traditionally opaque political systems. This democratization of information will lead to a more inclusive political landscape.

\noindent \textbf{Ethical Risks.} While LLMs offer substantial benefits, their societal deployment also raises critical ethical concerns. One major issue is the potential misuse of LLMs to disseminate misinformation or biased content, which could manipulate public opinion or destabilize democratic processes. Bai et al. \cite{bai2023artificial} discuss the persuasive power of LLM-generated text in influencing political opinions, underscoring the need for safeguards to mitigate risks. Furthermore, the ability of LLMs to generate realistic but misleading content poses challenges in distinguishing fact from fiction, creating vulnerabilities for misinformation campaigns. Addressing these ethical challenges  require robust governance frameworks and continuous monitoring.

\noindent \textbf{Summary and Challenges.} The societal impacts of LLMs are vast and multifaceted, offering opportunities to enhance political communication while raising ethical and democratic concerns. To fully leverage the potential of LLMs while mitigating risks, future research and governance efforts must focus on:  
\begin{itemize}[leftmargin=*]
\item {\textit{Responsible Deployment}}: Establishing guidelines for the ethical use of LLMs in politically sensitive contexts.  
\item {\textit{Transparency}}: Developing tools to track and explain LLM-generated content to avoid misuse.  
\item {\textit{Public Awareness}}: Educating users about the benefits and potential risks of LLMs to promote informed and responsible decision-making.  
\item {\textit{Misinformation Prevention}}: Implementing safeguards to detect and counteract biased or false narratives.  
\end{itemize}
By addressing these challenges, LLMs can contribute to a more equitable and transparent political environment, ensuring their societal impacts remain positive.

\section{Technical Foundations for LLM Applications in Political Science} \label{Computer-Science}
%Technical Foundations for LLM Adaptation in Political Science
%Original Title: Computational Approaches for Advancing LLMs in Political Science

\subsection{Benchmark Datasets}
\label{Benchmark}

\begin{table*}[htbp]
\footnotesize
\centering
\caption{Existing benchmark datasets in LLM for Political Sciences.}
\label{tab:benchmarks}
\begin{tabularx}{\textwidth}{p{4cm} p{4cm} X}
\toprule
\textbf{Benchmark Datasets} & \textbf{Application Domain} & \textbf{Evaluation Criteria} \\
\midrule
\centering OpinionQA Dataset\cite{santurkar2023whose} & \centering Sentiment Analysis \& Public Opinion & Ability to answer 1,489 questions \\
\addlinespace
\centering PerSenT\cite{bastan-etal-2020-authors} & \centering Sentiment Analysis \& Public Opinion & Performance on 38,000 annotated paragraphs \\
\addlinespace
\centering GermEval-2017\cite{chebolu2022survey} & \centering Sentiment Analysis \& Public Opinion & Accuracy on 26,000 annotated documents \\
\addlinespace
\centering Twitter\cite{sharma2022fake} & \centering Sentiment Analysis \& Public Opinion & Analysis of 5,802 annotated tweets \\
\addlinespace
\centering Bengali News Comments\cite{saha2022sentiment} & \centering Sentiment Analysis \& Public Opinion & Performance on 13,802 Bengali news texts \\
\addlinespace
\centering Indonesia News\cite{waspodo2022indonesia} & \centering Sentiment Analysis \& Public Opinion & Sentiment analysis on 18,810 news headlines \\
\midrule
\centering U.S. Senate Statewide 1976-2020 \cite{DVN/PEJ5QU_2017} & \centering Election Prediction \& Voting Behavior & Analysis of 3,629 data points\\
\addlinespace
\centering U.S. House 1976-2022 \cite{DVN/IG0UN2_2017} & \centering Election Prediction \& Voting Behavior & Analysis of 32,452 data points \\
\addlinespace
\centering U.S. Senate Returns 2020\cite{DVN/ER9XTV_2022} & \centering Election Prediction \& Voting Behavior & Prediction accuracy on 759,381 data points \\
\addlinespace
\centering U.S. House Returns 2018\cite{DVN/IVIXLK_2022} & \centering Election Prediction \& Voting Behavior & Analysis of 836,425 data points \\
\addlinespace
\centering State Precinct-Level Returns 2018\cite{DVN/ZFXEJU_2022} & \centering Election Prediction \& Voting Behavior & Analysis of 10,527,463 data points \\
\addlinespace
\centering 2008 ANES Time Series Study\cite{payne2010implicit} & \centering Election Prediction \& Voting Behavior & Analysis of 2,322 pre-election and 2,102 post-election surveys \\
\addlinespace
\centering 2016 ANES Time Series Study\cite{yu2024trumpwin2024predicting} & \centering Election Prediction \& Voting Behavior & Analysis of 2,322 pre-election and 2,102 post-election surveys \\
\addlinespace
\centering U.S. President 1976–2020\cite{DVN/42MVDX_2017} & \centering Election Prediction \& Voting Behavior & Analysis of 4,288 data points \\
\midrule
\centering BillSum\cite{kornilova2019billsum} & \centering Legislation \& Administrative Rules & Summarization of 33,422 U.S. Congressional bills \\
\addlinespace
\centering CaseLaw\cite{shu2024lawllm} & \centering Legislation \& Administrative Rules & Analysis of 6,930,777 state and federal cases \\
\addlinespace
\centering DEU III\cite{arregui2022new} & \centering Legislation \& Administrative Rules & Performance on 141 legislative proposals and 363 controversial issues \\
\addlinespace
\centering Federal Register Final Rule Data 2000-2014\cite{DVN/ZH7J2G_2018} & \centering Legislation \& Administrative Rules & Titles and Summaries of 61,216 U.S. Federal Regulations \\
\midrule
\centering PolitiFact\cite{shu2020fakenewsnet} & \centering Misinformation Detection & Detection across six integrated datasets \\
\addlinespace
\centering GossipCop\cite{grover2022public} & \centering Misinformation Detection & Detection across ten integrated datasets \\
\addlinespace
\centering Weibo\cite{jin2017multimodal} & \centering Misinformation Detection & Classification of 4,488 fake news and 4,640 real news items \\
\addlinespace
\centering SciNews\cite{cao2024can} & \centering Misinformation Detection & Detection in 2,400 scientific news stories \\
\midrule
\centering UCDP\cite{cunningham2013non} & \centering Game Theory \& Negotiation & Analysis of armed conflicts and peace agreements \\
\addlinespace
\centering PNCC\cite{ari2023peace} & \centering Game Theory \& Negotiation & Data on peace agreements and conflict resolution \\
\addlinespace
\centering WebDiplomacy\cite{meta2022human} & \centering Game Theory \& Negotiation & Analyze 12,901,662 messages exchanged between players \\
\bottomrule
\end{tabularx}
\end{table*}

To meet the specific demands of political science applications, various benchmark datasets grounded in real-world data have been developed to evaluate LLMs on tasks such as sentiment analysis, election prediction, legislative summarization, misinformation detection, and conflict resolution. Each dataset is designed with domain-specific criteria to assess the alignment of LLM outputs with real-world political and social contexts, ensuring their relevance and applicability to practical scenarios. A comprehensive list of these datasets, along with their respective tasks and charactristics, is presented in Table.~\ref{tab:benchmarks}  to facilitate reference and comparison.

\noindent \textbf{Sentiment Analysis \& Public Opinion Dataset.} Various datasets have been developed to accurately assess LLMs in sentiment analysis and public opinion. For instance, OpinionQA \cite{santurkar2023whose} is designed as a test environment where LLMs answer questions about public opinion, capturing subtle sentiments across 1,489 well-crafted queries. This dataset is valuable because it benchmarks how closely LLMs can align with actual human opinion patterns—a key factor for extracting sentiment accurately in social sciences. Similarly, PerSenT \cite{bastan-etal-2020-authors} focuses on tracking sentiments toward specific entities mentioned in news articles. It tests how well LLMs can detect and follow opinions expressed by particular individuals, allowing for sentiment to be aggregated over multiple mentions of popular entities to support comprehensive public opinion analysis. In addition, GermEval-2017 \cite{chebolu2022survey} provides a corpus of social media comments about Deutsche Bahn, the railway service in Germany, tailored for aspect-based sentiment analysis. This would help organizations and service providers derive actionable insights from feedback by homing in on specific aspects such as noise levels or punctuality. Datasets like Twitter \cite{sharma2022fake}, Bengali News Comments \cite{saha2022sentiment}, and Indonesia News \cite{waspodo2022indonesia} extend the sentiment analysis to widely used social and news media platforms in multiple languages. These multilingual datasets are very important for cross-linguistic and cultural sentiment studies, which find specially relevant applications in global social media and market research.

\noindent \textbf{Election Prediction \& Voting
Behavior Dataset.} The U.S. Senate Statewide 1976-2020 \cite{DVN/PEJ5QU_2017} dataset contains state-level election returns, while the U.S. House 1976-2022 \cite{DVN/IG0UN2_2017} dataset provides district-level returns, offering resources for analyzing nearly five decades of electoral trends. Other than that, The U.S. Senate Returns 2020 \cite{DVN/ER9XTV_2022} and U.S. House Returns 2018 \cite{DVN/IVIXLK_2022} datasets offer detailed precinct-level voting data, allowing LLMs to analyze U.S. voting patterns and voter behavior with the highest granularity, which supports election prediction and voting behavior studies. The State Precinct-Level Returns 2018 dataset \cite{DVN/ZFXEJU_2022}, with its extensive 10 million data points, provides a substantial resource for LLMs to train on and analyze voting behaviors comprehensively. The 2008 American National Election Study (ANES) \cite{payne2010implicit} offers insights into voter preferences and political attitudes through surveys conducted before and after the election, capturing difference in voter sentiment, which LLMs can model to reflect public opinion changes. The U.S. President 1976–2020 dataset \cite{DVN/42MVDX_2017} provides historical data essential for LLMs to examine long-term political trends and election outcomes across multiple decades. These datasets serve as invaluable training sources for LLMs to support political campaigns, media analysis, and social science research into electoral behaviors and trends.

\noindent \textbf{Legislation \& Administrative
Rules Dataset.} For summarizing and analyzing legislation and administrative rules, key datasets include BillSum \cite{kornilova2019billsum}, CaseLaw \cite{shu2024lawllm} and Federal Register \cite{DVN/ZH7J2G_2018}. BillSum aims at offering support to summarize US Congressional bills; it empowers LLMs to process mid-length legislative text and to produce brief summaries, which would considerably reduce the efforts of experts from the legal community and policy analysis. The CaseLaw dataset provides an extensive collection of state and federal cases, serving as a foundation for LLMs to analyze legal precedents and support judicial decision-making. The DEU III dataset \cite{arregui2022new} spans three decades of EU legislative decision-making, enabling the evaluation of LLMs in analyzing policy positions and negotiation dynamics among EU member states and institutions. Beyond legislation, the U.S. Federal Register dataset \cite{DVN/ZH7J2G_2018} includes titles and summaries of all final federal rules from 2000 to 2014, focusing on administrative decisions. This dataset provides a valuable resource for LLMs to analyze regulatory trends and the decision-making processes of federal agencies.

\noindent \textbf{Misinformation Detection Dataset.} To address the negative effects of fake news and misleading information, several open-sourced datasets have been constructed~\cite{grover2022public,jin2017multimodal}. PoliFact~\cite{shu2020fakenewsnet} supports the use of large language models to distinguish between false and genuine news by focusing on publisher behavior, user interactions, and network structures. Similarly, SciNews~\cite{cao2024can} concentrates on misinformation in scientific reporting, providing a resource that helps preserve the integrity of science communication and limit the spread of misleading health and science information.

\noindent \textbf{Game Theory \& Negotiation Dataset.} In the domain of conflict resolution and game theory research, there are datasets that guide the study of strategic interactions and peace negotiations. For example, the Non-State Actors in Armed Conflict (NSA) dataset \cite{cunningham2013non} includes information on state-rebel group dyads, enabling more detailed examinations of conflicts with actor-specific data. In addition, the Peace Negotiations in Civil Conflicts (PNCC) dataset \cite{ari2023peace} documents formal negotiation phases during civil conflicts. Moreover, the WebDiplomacy dataset \cite{meta2022human} consists of message exchanges between players in a simulated diplomatic negotiation setting, enabling a clearer understanding of communication patterns and strategic decision-making in conflict scenarios.

These benchmark datasets, taken together, provide a solid mainstay for a truly large number of LLMs applications in political science, from voter sentiment analysis to the exploration of legislative choices, tracking misinformation, and modeling conflict negotiations.

\subsection{Dataset Preparation Strategies}
\label{dataset-preparation}

Dataset preparation is a critical step in adapting LLMs for downstream political science applications~\cite{yu2024makes}. Given that the adaptation of LLMs in computational political science (CPS) is still in its infancy, the publicly available benchmark datasets remain scarce. The preparation of CPS datasets requires careful consideration of both domain-specific and generalizable strategies~\cite{lin2024designing,wagner2024power}. Drawing insights from adjacent research fields like general sentiment analysis, fake news detection, and LLM-based dialogue generation, political datasets can be adapted to align with tasks such as election prediction, policy analysis, and political discourse generation.

\noindent \textbf{Broad Source of Dataset Collection.} One primary approach of dataset preparation involves collecting text data from publicly available political sources, such as speeches, legislative records, news articles, and social media platforms. For instance, in OpinionQA~\cite{santurkar2023whose} and PerSenT~\cite{bastan-etal-2020-authors}, the data is sourced from political discussions and news media, which is then annotated for tasks like opinion alignment and sentiment detection. To ensure the data is relevant and representative, these dataset collections usually focus on specific political events, ideologies, or actors, which are essential for training LLMs to understand political discourse.

\begin{figure}[htbp]
\centering
\includegraphics[width=0.98\linewidth]{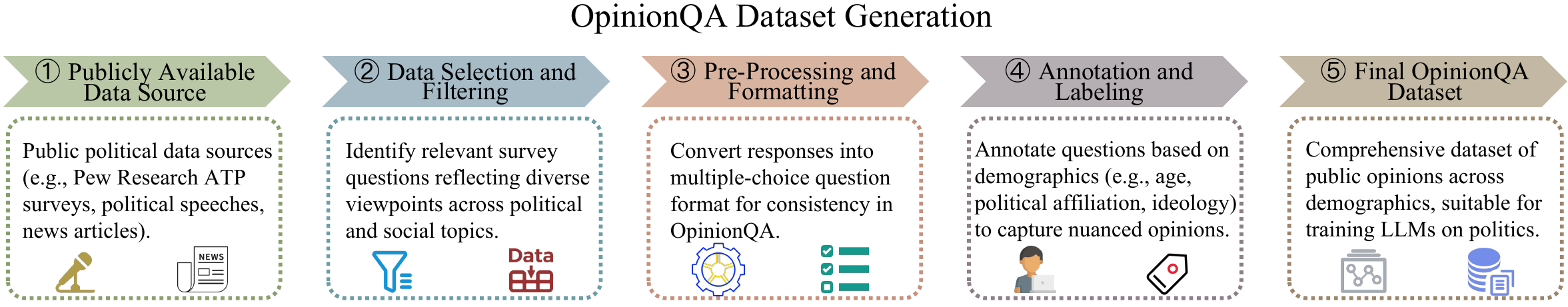}
\caption{Illustration of the OpinionQA dataset preparation on publicly available data source.}
%\Description{Illustration of Dataset Prep on Publicly Available Data}
\label{fig-OpinionQA}
\end{figure}

We elaborate the developing process of OpinionQA dataset in Fig.~\ref{fig-OpinionQA}. To start with, researchers utilized publicly available data from various political and social surveys as the source data. They particularly leverage Pew Research's American Trends Panel (ATP) surveys, which span a wide array of topics, including science, politics, and social issues. The dataset compilation process involves selecting pertinent survey questions that reflect diverse viewpoints across key issues and topics in the United States. These survey responses are preprocessed to create a multiple-choice question format, which serves as a reliable structure for language models to interpret. Through the methodology, each question in OpinionQA is annotated based on survey results, representing public opinion across various demographics such as age, political affiliation, income, and ideology. This approach ensures that the dataset encapsulates the complexity and nuance of real-world opinions, which are essential for training language models to simulate and interpret politically charged discourse accurately.

\begin{wrapfigure}[25]{r}{0.47\textwidth} 
\centering
\vspace{-0.1in}
\includegraphics[width=0.46\textwidth]{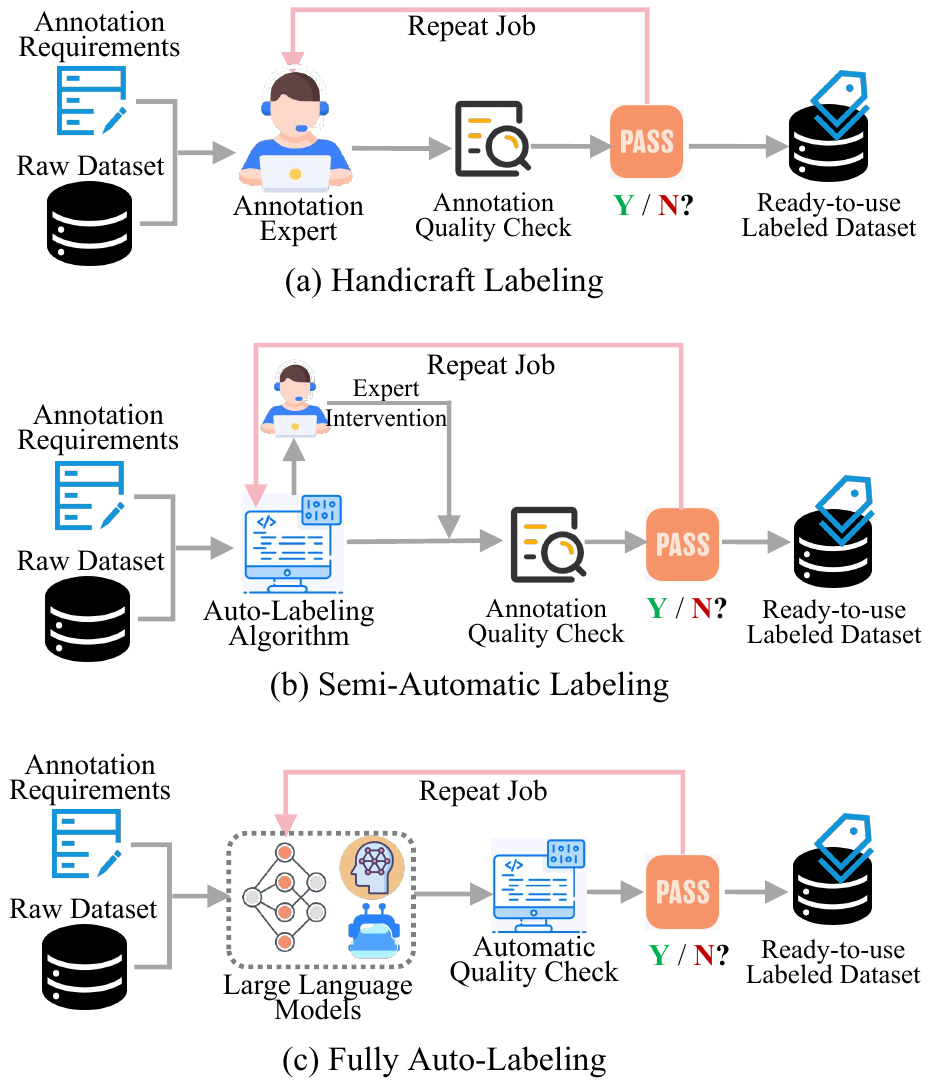}
\caption{Dataset annotation approaches, including traditional manual approach, semi-automated approach, and LLM-based fully automated approach.}
\label{fig-Annotation}
\end{wrapfigure}

\noindent \textbf{Annotation Strategies.} Annotation is another essential aspect of dataset preparation. Datasets intended for political sentiment analysis or misinformation detection require detailed labeling, often involving either expert or crowd-sourced annotations~\cite{mochtak2023parlasent}. For instance, the State Precinct-Level Returns 2018 dataset~\cite{DVN/ZFXEJU_2022} includes a substantial amount of real, unannotated data. Training LLMs with such data may involve adding annotations to capture sentiment toward political entities or identify media biases. Annotation schemes should be crafted to reflect nuanced political ideologies and opinions, ensuring that the dataset reflects the diversity and complexity of political discourse~\cite{balloccu2024leak,rauniyar2023multi}.

As shown in Figure~\ref{fig-Annotation}, annotation can be conducted through different approaches. These methods range from fully manual labeling~\cite{tan2024large}, where annotation experts review and label the data by hand, to semi-automated processes that use algorithms to assist with labeling~\cite{huang2024selective}, with experts intervening as needed. In fully automated labeling, LLMs or other automated systems can handle the labeling work entirely, followed by a quality check~\cite{ming2024autolabel}. Each method has its trade-offs among accuracy, scalability, and manual effort required.
% \begin{figure}[htbp]
% \centering
% \includegraphics[width=0.60\linewidth]{figures/Sec5-2-Annotation-Methods.pdf}
% \caption{Dataset annotation approaches, including traditional manual approach, semi-automated approach, and LLM-based fully automated approach.}
% \label{fig-Annotation}
% \end{figure}

\noindent \textbf{Dataset Bias and Representation.} Addressing bias and representation is particularly crucial in political science datasets. Datasets must account for the diversity of political systems, ideologies, and demographics~\cite{qu2024performance,shahbazi2023representation}. Researchers must ensure the collected political datasets are balanced across different viewpoints and the data does not over-represent certain political ideologies. Techniques such as oversampling underrepresented groups or creating synthetic data using LLMs can be employed to achieve this balance~\cite{nakada2024synthetic,cloutier2023fine}.

\noindent \textbf{Data Preprocessing \& Normalization.} Given the complexity of political language, appropriate preprocessing and normalization are indispensable~\cite{chai2023comparison}. Preprocessing steps such as entity recognition, text cleaning, and the extraction of key political terms help standardize the input and improve the model's ability to learn from diverse political contexts~\cite{ehrmann2023named}. These techniques ensure that LLMs can process the input text effectively.

\noindent \textbf{Data Augmentation.} Augmentation strategies like paraphrasing or generating synthetic data with LLMs help to expand the dataset size in cases where political data is limited~\cite{sahu2023promptmix,abaskohi2023lm}. Data augmentation helps diversify the training set, allowing the model to generalize better to new and unseen political scenarios~\cite{dos2024identifying,ding2024data}.

To further illustrate how these strategies applied to practical scenarios, we now introduce three examples of dataset preparation tailored for specific LLM-based political science tasks. Each example demonstrates how researchers effectively leverage LLMs to address key challenges in political data curation and annotation:

\noindent \textbf{(1) Developing a Dataset for LLM-Based Political Debiasing.} For the political debiasing task, constructing a dataset involves curating a balanced collection of political texts that represent diverse political ideologies and viewpoints. For instance, to debias LLM outputs, we can gather news articles, social media posts, and political speeches from various political parties, regions, and ideologies. The dataset will need to be annotated with the political bias present in each text. This can be done using a combination of manual annotation by political experts and automated tools to identify biased language, sentiment, and framing. The goal is to provide a dataset that allows the model to recognize and mitigate its inherent biases by learning from a balanced set of inputs across the political spectrum.

\noindent \textbf{(2) Automated Annotation Using LLMs: Example in Legislative Interpretation.} LLM-based legislative interpretation is a promising application in political science. Using a dataset like BillSum~\cite{kornilova2019billsum}, which includes U.S. legislative documents, LLMs can be employed to automatically annotate sections of the legislation with relevant policy categories, key provisions, and political implications. LLMs can also be fine-tuned on a smaller, manually annotated set of legislative texts in order to classify different legal concepts and policy issues. This automated annotation streamline will accelerate the process of categorizing large volumes of legislative content, helping political analysts and lawmakers quickly interpret and summarize complex bills.

\noindent \textbf{(3) Generating Synthetic Political Datasets Using LLMs.} The limitations in acquiring large and diverse political datasets due to privacy, restrictions, and sensitivities make generating synthetic datasets with LLMs a promising solution. Considering election prediction as an example, LLMs are able to generate hypothetical voter opinion surveys based on historical election data and known demographic trends. By training LLMs on existing public opinion survey datasets, researchers can generate synthetic datasets that simulate different electoral conditions, voter behaviors, and political trends. This approach will greatly enhance the availability of diverse political data for training and testing election prediction models.

%\subsection{LLM-based Model Design for Political Science}
%\cite{simmons-2023-moral}
%\cite{zhou-tan-2023-entity}
%\cite{thapa-etal-2023-assessing}
%\cite{lee-etal-2024-exploring-inherent}
%\cite{ramezani-xu-2023-knowledge}

\subsection{Fine-Tuning LLMs for Political Science}\label{fine-tuning}

This subsection explores the fine-tuning of LLMs for political science applications, using \textit{Automatic Summarization of Legislation} as the application task. Specifically, we employ BillSum dataset~\cite{petrova2020extracting}, which includes U.S. Congressional and California state bills, to fine-tune an LLM to generate concise, accurate summaries of legislative documents. This task highlights the challenges of summarizing complex, structured texts like bills, which differ significantly from conventional summarization domains.

\noindent \textbf{Domain-Specific Dataset--BillSum.} Curated datasets play a critical role in fine-tuning large language models~\cite{zhang2024scaling}, as they provide the domain-specific knowledge and tailored examples necessary to adapt the pre-trained LLM's general capabilities to specialized applications~\cite{vm2024fine,jaradat2024multitask}. BillSum~\cite{petrova2020extracting} is designed for the legislative summarization task, containing over 22,000 U.S. Congressional bills and summaries as well as additional California state bills as an out-of-domain test set. This dataset is particularly challenging due to the hierarchical and technical language structure of legislative texts~\cite{santosh2024lexsumm}. BillSum’s structure enables us to focus on mid-length legislation and provides ground-truth summaries created by experts, making it ideal for training an LLM to handle specialized summarization tasks in the legislative domain~\cite{guha2024legalbench}.

\noindent \textbf{Fine-Tuning Process for LLMs.} Fine-tuning involves adapting a pre-trained LLM (e.g., Llama2) to the specific requirements of legislative summarization. The procedure includes the following steps:

\begin{enumerate}[label=\arabic*., leftmargin=*]
\item \textit{Data Preprocessing:} Legislative documents in BillSum must be preprocessed to create well-aligned input-output pairs suitable for text-to-text learning formats. For example, each bill text is paired with its corresponding summary, formatted to maintain structural consistency. Preprocessing may also involve tokenization adjustments and removal of excessive legal jargon that is irrelevant to summary generation.

\item \textit{Fine-Tuning Setup:} To optimize fine-tuning, parameter-efficient techniques such as Low-Rank Adaptation (LoRA)~\cite{wu2024dlora} or prefix-tuning~\cite{meloux2024novel} are applied. These techniques adjust a minimal subset of model parameters, reducing computational costs and enabling the model to retain same level of accuracy while adapting to domain-specific context. Hyperparameters are carefully tuned based on summarization metrics like ROUGE and BLEU scores, which evaluate the accuracy and completeness of generated summaries~\cite{banerjee2023benchmarking}.

\item \textit{Training Process:} Training is conducted on GPU clusters to maximize performance, with techniques such as gradient accumulation~\cite{nabli2024acco} and mixed-precision~\cite{guan2024aptq} training to manage memory efficiently. The model iteratively adjusts its parameters by minimizing the loss function relative to the ground-truth summaries provided in the dataset. This involves several epochs with regular validation to prevent overfitting, ensuring the model generalizes well to unseen legislative texts.
\end{enumerate}

\noindent \textbf{Prompt Engineering for Fine-Tuning.} Prompt engineering plays a crucial role in guiding the LLM to produce accurate and concise summaries that capture essential aspects of legislative texts. In fine-tuning, prompts are designed to clarify the task, structure the response, and focus the model on important content~\cite{ding2023parameter}. Below are examples of prompt structures that can be used in training with BillSum:

\begin{tcolorbox}[colback=blue!5!white, colframe=blue!75!green, title=\textbf{Prompt Engineering Examples for U.S. Legislative Summarization}]
\centering
\faLightbulbO: Prompt \hfill
\raggedright
\rule{15cm}{1pt}
\linespread{1.25}
\selectfont

\faLightbulbO \ \ding{182}: Read the following U.S. Congressional bill text and provide a summary that highlights the main objectives, intended outcomes, and any significant amendments.

\faLightbulbO \ \ding{183}: Summarize this bill in no more than 5 sentences, focusing on its primary goals and any actions it authorizes or mandates. Avoid technical jargon.

\faLightbulbO \ \ding{184}: Provide a legislative summary of the bill below, identifying the key provisions and any departments or agencies involved. The summary should be clear and accessible to the general public.

\faLightbulbO \ \ding{185}: Using simple language, describe the main points of this bill, including what it aims to change, whom it affects, and any funds it allocates.
\end{tcolorbox}

These prompts serve to guide the LLM towards generating summaries that are not only accurate but also accessible to a broader audience. By providing explicit instructions, these prompts help the model focus on the legislative document's core components, such as goals, affected parties, and any key changes to existing laws.

\noindent \textbf{Expected LLM Outputs.} After fine-tuning, the LLM is expected to perform effectively across several legislative summarization tasks:

\begin{itemize}[leftmargin=*]
\item \textit{Summarization of Legislative Bills:} Fine-tuned on BillSum, the model should generate coherent, concise summaries that capture the main objectives and impacts of U.S. Congressional bills. For example, the model might summarize a bill on environmental regulation by highlighting proposed restrictions, target pollutants, and enforcement mechanisms.
\item \textit{Generalization to Other Legislative Texts:} Given BillSum’s inclusion of California state bills as an out-of-domain test set, the fine-tuned model should also demonstrate the ability to generalize to state-level legislation, even when the language or structure differs from federal bills.
\item \textit{Summary Structure and Clarity:} The fine-tuned model should produce summaries that are structured to facilitate understanding, ideally avoiding overly technical language or verbose descriptions. This includes providing summaries that are straightforward and tailored for readers without specialized legal knowledge.
\end{itemize}

Fine-tuning an LLM on the BillSum dataset enables the model to handle the complex task of legislative summarization. The process combines domain-specific fine-tuning with practical prompt engineering to ensure that the model generates accurate, concise, and accessible legislative summaries, thus enhancing transparency and efficiency in legal information dissemination.

\subsection{Inference with LLMs: Zero-Shot In-Context Learning}
\label{zero-shot}

This subsection demonstrates how to practice Zero-Shot Learning (ZSL) in political science. We use \textit{Sentiment Analysis during the U.S. Presidential Election} as the application case. Zero-shot learning enables a pre-trained LLM to perform sentiment analysis on political statements without additional task-specific training.

\noindent \textbf{Overview of Zero-shot Learning (ZSL).} ZSL enables pretrained LLMs to perform political science tasks without additional task-specific data or model updates~\cite{wei2021finetuned,kojima2022large}. This approach is particularly valuable in computational political science (CPS) due to the frequent scarcity of annotated data~\cite{kumar2023zero}. By leveraging extensive pretraining on general-purpose data, ZSL allows LLMs to infer political language patterns through context, making it possible to analyze sentiment~\cite{kumar2023zero}, classify ideology~\cite{di2024mapping}, or interpret complex policy discourse~\cite{allaway2023zero} directly from raw prompts.

\noindent \textbf{Task-Specific Prompt Engineering.} Effective prompt engineering is essential to guide the LLM accurately in zero-shot mode~\cite{kuila2024deciphering}. In the case of U.S. Presidential Election sentiment analysis, prompts must be designed to capture the nuances of political statements and infer the sentiment accurately~\cite{hu2023synthesizing,burnham2024political}. Here are practical examples of sentiment classification prompts, using statements from publicly available news articles or social media posts regarding the 2024 election.

\begin{tcolorbox}[colback=blue!5!white, colframe=blue!75!green, title=\textbf{Public Opinion Poll Sentiment Classification during U.S. Presidential Election}]
\centering
\faLightbulbO: Prompt \hfill \faRedditAlien: LLM response \hfill \faQuestionCircleO: Explanation
\raggedright
\rule{15cm}{1pt}
\linespread{1.25}
\selectfont

\faLightbulbO: Analyze the sentiment of the following statement about the presidential election as Positive, Negative, or Neutral. "Despite the turbulent political climate, Candidate X has shown strong leadership qualities and promises substantial reforms that could benefit the economy."

\faRedditAlien: Positive.

\faQuestionCircleO: The statement uses phrases such as :"strong leadership" and "substantial reforms," indicating a positive sentiment towards Candidate X's potential economic impact.

% --------------------------------------------------------------------------------------------------------------------
% \vskip 1ex
% \centering
% \rule{15cm}{1pt}
% \vskip 1ex

% \faLightbulbO: Prompt; \hfill \faRedditAlien: LLM response; \hfill \faQuestionCircleO: Explanation
% \raggedright

\end{tcolorbox}

Another example tailored for analyzing public opinion on social media:

\begin{tcolorbox}[colback=blue!5!white, colframe=blue!75!green, title=\textbf{Social Media Sentiment Analysis for U.S. Presidential Election}]
\centering
\faLightbulbO: Prompt \hfill \faRedditAlien: LLM response \hfill \faQuestionCircleO: Explanation
\raggedright
\rule{15cm}{1pt}
\linespread{1.25}
\selectfont

\faLightbulbO: Determine whether the sentiment expressed in this tweet about Candidate Y in the upcoming election is Positive, Negative, or Neutral. "Candidate Y's policies on healthcare are exactly what we need! Finally, someone who cares about the people."

\faRedditAlien: Positive.

\faQuestionCircleO: The tweet expresses approval through phrases like "exactly what we need" and "cares about the people," signaling positive support for Candidate Y's healthcare policies.

\end{tcolorbox}

These prompts are structured to guide the LLM in accurately detecting sentiment by highlighting specific phrases and context, improving the reliability of sentiment classification in a zero-shot setting.

\noindent \textbf{Embedding Context in Prompts.} Including contextual information is crucial for ZSL tasks, especially in political sentiment analysis, where statements often depend on the socio-political context~\cite{wahidur2024enhancing}. For example, prompts can specify details such as the speaker's political affiliation, the event's date, or relevant policy areas. This helps the model interpret statements with greater accuracy. In politically charged scenarios, prompts might include contextual cues, like "This statement was made during a recent debate on immigration policy," enabling the model to better understand the sentiment nuances within the political context.

\noindent \textbf{Applications in Political Science.} ZSL has broad applications across political science tasks. In addition to sentiment analysis, ZSL enables stance classification, allowing LLMs to determine whether a statement supports or opposes a particular issue without task-specific data. This approach also extends to policy categorization and ideological scaling, where LLMs can classify political statements or documents into ideological categories, such as conservative, liberal, or centrist, without extensive labeled data. By eliminating the need for annotated datasets, ZSL enables LLMs to quickly adapt to exploratory political science tasks, making it a flexible, cost-effective option for rapid analysis.

\noindent \textbf{Use Case: Sentiment Analysis in the U.S. Presidential Election.} As a demonstration, we apply zero-shot sentiment analysis to analyze public opinion during the 2024 U.S. Presidential Election~\cite{kuila2024deciphering,gujral2024can}. By leveraging Llama2, we perform sentiment classification on statements from news articles, debate transcripts, and social media posts, enabling us to gauge public sentiment toward candidates and their policies without requiring fine-tuning on election-specific data.

\textit{Example Prompt for Sentiment Analysis.} The following prompt is designed to assess sentiment on election-related statements, guiding the LLM to infer the sentiment based on contextual knowledge:

\vspace{-1mm}
\begin{tcolorbox}[colback=blue!5!white, colframe=blue!75!green, title= \textbf{Political Sentiment Analysis in 2024 U.S. Election}]
\centering
\faLightbulbO: Prompt \hfill \faRedditAlien: LLM response \hfill \faQuestionCircleO: Explanation
\raggedright
\rule{15cm}{1pt}
\linespread{1.25}
\selectfont

\faLightbulbO: Evaluate the sentiment in the following statement about the presidential election (Positive, Negative, or Neutral). "The recent tax proposal from Candidate X is likely to hurt middle-class families while favoring large corporations."

\faRedditAlien: Negative.

\faQuestionCircleO: The sentiment is negative due to phrases like "hurt middle-class families" and "favoring large corporations," which indicate disapproval of Candidate X’s tax proposal.
\end{tcolorbox}
\vspace{-1mm}

This prompt helps the LLM capture sentiment indicators in election-related contexts, such as impacts on specific social groups or policy critiques~\cite{wicke2024red}. By structuring prompts carefully, ZSL can facilitate nuanced sentiment analysis on real-world political data.

In summary, ZSL enables effective and efficient sentiment analysis of political statements during key events (e.g., presidential elections). With the use of prompt engineering to provide clear task instructions and context, ZSL can support political science research by rapidly assessing public sentiment, ideology, and stance~\cite{ibrahim2024analyzing} without needing additional training data, making it an invaluable tool in data-scarce political environments.

% \subsection{In-Context Learning: Few-Shot Inference for LLMs}
\subsection{Inference with LLMs: Few-Shot In-Context Learning}
\label{few-shot}

%Few-Shot Learning Inference
\noindent \textbf{Overview of Few-shot Learning.} Few-shot learning allows LLMs to perform specialized tasks with minimal labeled data by embedding a small number of example prompts within the input. This approach bridges the gap between zero-shot and full fine-tuning by enhancing the model’s contextual understanding through example-driven cues.~\cite{burnham2024political,kuila2024deciphering} Few-shot learning is particularly beneficial in political science applications where obtaining annotated data can be challenging and costly~\cite{malladynamic}, providing a flexible, cost-effective solution.

\noindent \textbf{Designing Effective Few-shot Examples.} The effectiveness of few-shot learning heavily relies on the selection of representative examples~\cite{hu2023synthesizing,burnham2024political}. For the task of fake news detection in the 2024 U.S. Presidential Election, it is critical to include examples that capture various degrees of misinformation subtlety and diversity in topics~\cite{kuntur2024under}. These examples should reflect the language and tactics often used in fake news, such as sensationalism, partial truths, and emotionally charged language~\cite{pavlyshenko2024using}. Below is a sample prompt designed to guide the model in distinguishing real news from fake news.

\begin{tcolorbox}[colback=blue!5!white, colframe=blue!75!green, title=\textbf{Determine if Each News Headline is Real or Fake}]
\centering
\faDatabase: Corpus database \hfill \faTags: Label \hfill \faLightbulbO: Prompt \hfill \faRedditAlien: LLM response
\raggedright
\rule{15cm}{1pt}
\linespread{1.25}
\selectfont

\faDatabase: "Presidential candidate X pledges new economic reforms to boost national job growth and reduce unemployment."

\faTags: Real.

\faDatabase: "Urgent: Voting machines in <Swing State> are flipping votes from Candidate X to Candidate Y, claims anonymous election officer."

\faTags: Fake.

% \faDatabase: "Thousands attend rally in support of presidential candidate Y's healthcare plan."

% \faTags: Real.

\faLightbulbO: "Scientists confirm presidential candidate X is involved in alien cover-up."

\faRedditAlien: ...............
\end{tcolorbox}

This prompt helps the model understand distinctions between credible and misleading information by using a mix of realistic and exaggerated statements. Examples like "foreign interference" or "alien cover-up" are designed to train the model to recognize patterns typical of fake news narratives.

\noindent \textbf{Embedding Context in Prompts.} Contextual cues are essential in few-shot learning for politically sensitive tasks. For instance, when detecting fake news during the 2024 U.S. Presidential Election, prompts can be enhanced by adding details such as the publication date or the source of the statement~\cite{whitehouse2022evaluation}. This contextual embedding helps the model assess the plausibility of a statement more accurately. Specifying that a statement originated from a reliable media versus an anonymous social media post can guide the model's judgment~\cite{molina2021fake}. 

\begin{tcolorbox}[colback=blue!5!white, colframe=blue!75!green, title=\textbf{Example with contextual Cues for Fake News Detection during the U.S. Presidential Election}]
\centering
\faDatabase: Corpus database \hfill \faTags: Label \hfill \faLightbulbO: Prompt \hfill \faRedditAlien: LLM response
\raggedright
\rule{15cm}{1pt}
\linespread{1.25}
\selectfont

\faDatabase: "Presidential candidate X states that billions of taxpayer dollars are wasted annually on benefits for illegal immigrants. X pledges to end these benefits and redirect the funds to support hardworking American taxpayers."

\faTags: Real.

% \faDatabase: "Anonymous social media account in TikTok claims that election will be delayed due to secret government plot."

% \faTags: Fake.

\faDatabase: "CNN, The New York Times, and The Washington Post report historic voter turnout in early voting during the 2024 U.S. Presidential Election."

\faTags: Real.

\faLightbulbO: "Unconfirmed sources suggest presidential candidate plans to abolish social security if elected."

\faRedditAlien: ................
\end{tcolorbox}

In this prompt, contextual cues like "CNN, The New York Times, and The Washington Post" and "Anonymous social media account in TikTok" help the model make more informed predictions. Adding such cues aids the model in understanding the credibility of the information, which is crucial for politically charged topics.

\noindent \textbf{Balancing Prompt Length and Example Diversity.} In few-shot learning, the balance between prompt length and example diversity is critical~\cite{yao2024more}. While additional examples can improve task performance, including too many can reduce prompt clarity and increase processing time~\cite{chen2024fine}. For fake news detection, three to four carefully chosen examples are typically sufficient to cover different levels of misinformation and factual reporting styles. Examples should be concise yet comprehensive, representing a variety of fake news tactics, such as sensationalism, misinformation about election logistics, or fabricated scandals.

\noindent \textbf{Use Cases in Political Science.} Few-shot learning has proven effective across various political science tasks. In the context of Fake News Detection for the 2024 U.S. Presidential Election, few-shot learning helps LLMs identify misleading narratives without extensive training data~\cite{kuila2024deciphering}, making it particularly valuable for real-time misinformation monitoring~\cite{hu2024multi}. Other applications include:

\begin{itemize}[leftmargin=*]
\item {\textit{Public Opinion Analysis}}: By using few-shot examples of sentiment-laden statements, LLMs can analyze public opinion towards candidates or policies, which enables to capture nuanced shifts in voter sentiment.
    
\item {\textit{Policy Stance Classification}}: With carefully crafted few-shot examples, LLMs can classify political statements as supportive or oppositional towards specific policies, helping understand public responses.
    
\item {\textit{Legislative Influence Prediction}}: Few-shot learning can help predict the likelihood of legislative support or opposition by providing examples of previous legislative behavior and context, assisting analysts in forecasting policy outcomes.
\end{itemize}

Few-shot learning's adaptability makes it an invaluable tool in political science, allowing for nuanced understanding and classification in tasks with limited data. By incorporating diverse, context-rich examples, few-shot learning enables LLMs to navigate complex political discourse with improved accuracy and interpretability, as exemplified by its application in fake news detection during a high-stakes election cycle.

\subsection{Other Techniques Enhancing LLM Inference}~\label{other-inference}
%Other LLM Inference Methods
%\noindent \textbf{Retrieval-Augmented Generation (RAG).} RAG combines the power of knowledge retrieval and text generation to enhance the accuracy and relevance of LLM outputs. In the context of political science, RAG can be used to dynamically retrieve relevant information from external datasets or knowledge bases, such as policy documents, voting records, or historical data, and integrate this information into generated answers. For instance, when analyzing political discourse or predicting election outcomes, RAG can retrieve real-time data on polling, demographic trends, or legislative proceedings, ensuring that the generated output is contextually up-to-date and precise.

\noindent \textbf{Retrieval-Augmented Generation (RAG).} RAG is a powerful technique that combines knowledge retrieval with text generation to enhance the relevance and accuracy of responses produced by large language models~\cite{salemi2024evaluating}. Unlike standard language models that rely solely on pre-trained knowledge, RAG systems dynamically retrieve external data from a knowledge source (such as a database or corpus) to enrich the model's responses~\cite{wang2024evaluating}. This method mitigates the limitations of static knowledge in LLMs, ensuring that generated content is both up-to-date and contextually aligned with real-world events~\cite{dong2024understand}. In the context of political science, RAG is particularly valuable for tasks that require accurate, real-time information, such as policy analysis, electoral forecasts, and sentiment tracking~\cite{arslan2024political,arslan2024political}. By integrating recent polling data, legislative updates, or public opinion trends, RAG enables LLMs to respond to complex political questions with greater precision. This capability is essential for informed analysis and decision-making.

\begin{figure}[htbp]
\centering
\includegraphics[width=0.85\linewidth]{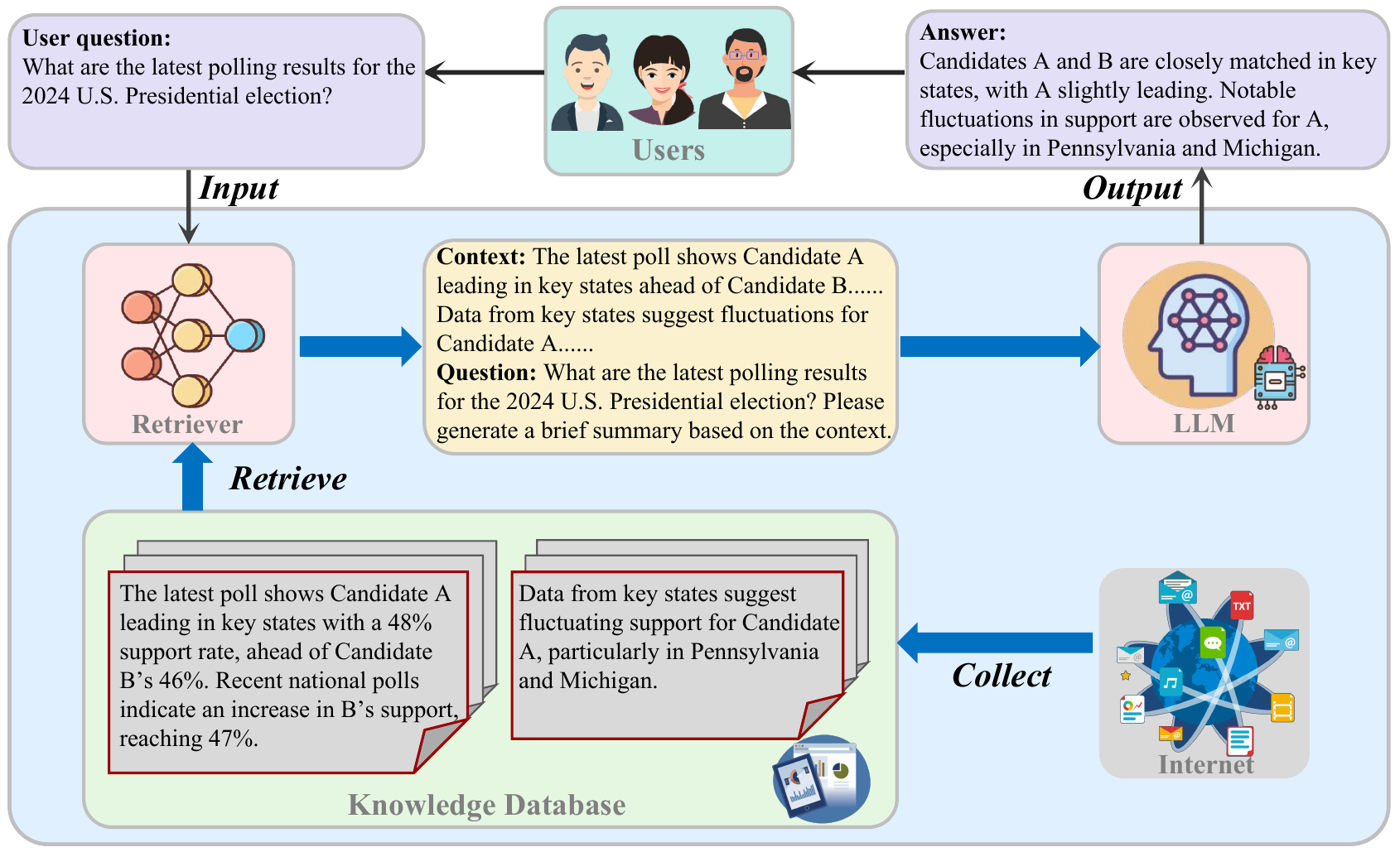}
\caption{Illustration of retrieval-augmented generation techniques on U.S. Presidential Election analysis.}
%\Description{Illustration of RAG}
\label{fig-LLM-Inference-RAG}
\end{figure}

In the 2024 U.S. presidential election polling scenario, RAG is employed to provide real-time and contextually accurate responses about the latest public opinion data. As shown in Figure~\ref{fig-LLM-Inference-RAG}, when users pose a question "What are the latest polling results for the 2024 U.S. Presidential election?", the RAG system initiates a two-step process. First, the retriever component accesses an up-to-date knowledge database containing polling data, including support rates for each candidate in key states and national trends from reputable sources like Gallup and the Pew Research Center. This retrieved data is then incorporated into the context, which includes specific figures and observations about fluctuating support levels, particularly in critical states such as Pennsylvania and Michigan. Next, the LLM uses this enriched context to generate a coherent response, producing an answer that not only addresses the user’s question but also reflects the most current information available. RAG is capable of integrating retrieval with generation, enabling LLMs to provide responses that are both timely and grounded in factual data, which is especially valuable in politically dynamic fields where data changes rapidly.

\noindent \textbf{Chain-of-Thought Reasoning}
\label{COT}
Chain-of-Thought (CoT) reasoning is a technique that guides LLMs through a step-by-step logical reasoning process, allowing the model to handle complex, multi-step analysis~\cite{zhanggenerating}. It simulates a structured human-like thought process, reducing the risk of oversimplification or bias in sensitive analysis. CoT reasoning is particularly valuable for political science tasks that require nuanced and multi-dimensional insights~\cite{kareem2023fighting,dobrinoiu2024leveraging}. By breaking down intricate questions into a sequence of simpler reasoning steps, CoT helps LLMs interpret complex political issues with greater accuracy and transparency~\cite{tutunov2023can}.

Considering a practical application of CoT reasoning in analyzing public opinion trends on a contentious policy, such as immigration reform. In this example, the objective is to predict the level of public support or opposition to a new immigration policy by examining multiple influencing factors in a sequential and structured manner. The reasoning process is proceed as follows:
\begin{enumerate}[leftmargin=*]
\item \textit{Historical Context Analysis}: The model first retrieves information on past immigration policies and public responses to them. This includes identifying previous legislative actions on immigration and analyzing public opinion data from those periods. By understanding historical trends, the model establishes a baseline for comparison with the current policy.

\item \textit{Demographic Impact Assessment}: Next, the model assesses demographic shifts that might influence public opinion, such as changes in immigrant populations, regional population density, and employment statistics in sectors affected by immigration. This step allows the model to evaluate how different demographic groups might respond to the policy.

\item \textit{Social Sentiment Analysis}: The model then analyzes recent public sentiment data from social media, news articles, and survey results. By focusing on keywords and sentiment trends related to immigration reform, the model identifies the prevailing attitudes and emotional responses of the public. This step is crucial for capturing real-time opinions.

\item \textit{Economic Implications}: The model evaluates potential economic effects of the policy, including its impact on labor markets, public spending, and economic growth. It considers both supportive and opposing viewpoints on the economic consequences of immigration reform. This analysis provides additional context for understanding why different groups may support or oppose the policy.

\item \textit{Synthesis and Prediction}: Finally, the model synthesizes information from each step, combining historical, demographic, social, and economic insights to produce a comprehensive assessment of public opinion on the immigration reform policy. By sequentially reasoning through each aspect, the model generates a nuanced prediction of public support levels.
\end{enumerate}

Through this step-by-step process, CoT reasoning enables the LLM to build a well-rounded understanding of public opinion trends. Each reasoning step enriches outputs, allowing the LLM to incorporate multiple facets of the issue and providing an interpretable chain of logic. This structured approach is ideal for complex political science tasks where multiple factors interact dynamically.

\noindent \textbf{Knowledge Editing.} It is a technique that allows for dynamic modification of the internal knowledge within an LLM without requiring full retraining~\cite{wang2023knowledge}. Knowledge Editing is highly feasible for political scenarios, where information, contexts, and factual data frequently changes. Instead of retraining the LLM, Knowledge Editing enables targeted updates to specific knowledge nodes or parameters within the model~\cite{zhang2024comprehensive,gupta2024stackfeed}. This characteristic allows the LLM to provide accurate responses to questions related to recent political events, policy changes, or updated facts, thus ensuring the model outputs remain relevant and reliable.

Knowledge Editing can be applied to update an LLM's understanding of specific policies or adjust its responses concerning historical events~\cite{zhang2024oneedit,peng2024event}. For instance, while researchers are analyzing an LLM's perspective on climate policies, the government introduces a new climate reduction target. Using Knowledge Editing, researchers can embed this new target directly into the LLM without a full-scale retraining. The process involves identifying the specific model parameters or representations tied to "climate policy" or "emission targets" and then modifying these nodes to incorporate the latest data. This targeted update ensures that the LLM refers to the updated policy goals when generating content about climate initiatives.

%\textcolor{red}{[TO DO]: Add an Illustration of Knowledge Editing under the topic of government climate protection goal}

\noindent \textbf{Self-Consistency Decoding.} This approach is designed to improve the reliability and robustness of responses generated by LLMs~\cite{ahmed2023better}. The basic idea behind Self-Consistency Decoding is to prompt the model multiple times with slightly varied initial conditions, allowing it to generate multiple candidate responses~\cite{huang2023enhancing}. These responses are then evaluated, and the most frequent or consistent response is selected as the final output~\cite{cheng2024integrative}. This method is particularly useful in reducing the effects of randomness in model generation, which can often lead to inconsistent or contradictory answers, especially in complex domains like political science. In political science applications, Self-Consistency Decoding can help stabilize the outputs of LLMs, ensuring that responses are not only accurate but also align consistently with established political theories or interpretations.

In political science tasks, Self-Consistency Decoding can be applied to analyze and interpret contentious or nuanced topics, such as public opinion on controversial policies~\cite{chen2023two}. For example, the LLM is prompted to analyze sentiments around immigration reform across different demographic groups. By asking the LLM to generate multiple analysis of the same policy, each with slight variations in prompt wording or contextual details, researchers can obtain a set of responses that capture different perspectives or interpretations. Self-Consistency Decoding is then used to identify the dominant perspective or interpretation. This approach not only enhances the robustness of LLM-generated insights but also helps researchers ensure that the final output reflects a consensus interpretation, minimizing the risk of bias or instability in model-generated analyses.

\subsection{Case Study: Political Bias and Feature Generation in LLM-Driven Voting Simulations}
\label{case_study}

As mentioned in Section~\ref{values_societal_impacts}, previous studies have shown that LLMs may exhibit potential political bias. To effectively quantify and evaluate this bias, we conducted a case study based on the benchmark dataset referenced in Section~\ref{Benchmark}. This study aimed to address two key aspects: (1) the biases displayed by different LLMs during voting simulation, and (2) the quality of LLM-generated political features compared to the original dataset, assessing their effectiveness for feature generation tasks in political science.

\subsubsection{LLM Model Configurations, Computational Resources, and Dataset Selection}

We evaluated four LLMs with varying parameter sizes: two commercial models, GPT-4o~\cite{islam2024gpt} and GPT-4o-mini~\cite{rasheed2024taskcomplexity}, and two open-source models, Llama 3.1 8-B~\cite{chen2024magicdec} and Llama 3.1-70B~\cite{singh2024scidqa}.

The hardware configurations were tailored to meet the computational requirements of each model. For GPT-4o and GPT-4o-mini, experiments were conducted on a GPU server equipped with an AMD EPYC Milan 7763 processor, 1 TB of DDR4 memory, 15 TB SSD storage, and 6 NVIDIA RTX A6000 GPUs. For Llama 3.1 models, a node with 8 NVIDIA A100 GPUs (each with 40 GB of memory), dual AMD Milan CPUs, 2 TB of RAM, and 1.5 TB of local storage was utilized.

The dataset used in this study was the 2016 Time Series Study from the American National Election Studies (ANES)~\cite{ANES2016,kennedy2018evaluation}. This dataset was selected for its comprehensive demographic, political ideology, and religious data, which provided a robust basis for analyzing the potential biases and generative capabilities of LLMs in a political science context.

\subsubsection{Experimental Design and Evaluation Methods}

\noindent \textbf{Experimental Design.}
To investigate voting simulation bias and feature generation quality, we adapted methodologies proposed in previous studies~\cite{yu2024trumpwin2024predicting,Argyle_Busby_Fulda_Gubler_Rytting_Wingate_2023}. In our experimental design, each selected LLM is provided with detailed persona information, including demographic characteristics, political ideology, and religious affiliation, as well as contextual information on candidates and policies relevant to the election year. Each LLM is then employed to simulate election voting behavior for each persona, allowing us to observe any biases that emerged in the simulated vote distributions~\cite{yu2024will}.

We designed two experimental pipeline setups for each LLM: a baseline group (denoted as [model name]-base) and a generation group (denoted as [model name]-gen). In the base group, LLMs used the original, unaltered ANES dataset inputs to simulate voting behaviors. In the generation group, we applied a multi-step Chain of Thought (CoT) approach, as described in Section~\ref{COT}. Here, LLMs were first prompted to generate political ideology features based on demographic inputs. These generated features were then combined with other persona details and used as inputs for the final voting simulation. This two-pipeline design allows us to evaluate the capability of these LLMs in generating relevant features within a political science context. Additionally, it enables us to analyze how these generated features might influence the bias in LLM voting simulations. The following popular general-purpose LLMs are selected as the benchmark models for our experiments: (I) gpt4o-mini-base, (II) gpt4o-base, (III) llama3.1-8B-gen, (IV) gpt4o-mini-gen, (V)gpt4o-gen, (VI) llama3.1-70B-base, (VII) gpt4o-NP, (VIII) llama3.1-8B-base, (IX) llama3.1-70B-gen.

The specific CoT design is illustrated with the following example prompts:

\begin{promptbox}
\textcolor{boxblue}{\textbf{Step 1: Ideology Assessment.}}  
You are a persona with the following demographic characteristics: [demographics]. The current year is [year]. Here are the policy agendas of the two parties:

[Two parties' policy agenda].

When it comes to politics, would you describe yourself as:

\begin{center}
\begin{tabularx}{\linewidth}{X X}
    No answer \& Very liberal\\
    Somewhat liberal \& Closer to liberal\\
    Moderate \& Closer to conservative\\ 
    Somewhat conservative \& Very conservative\\
\end{tabularx}
\end{center}

\textcolor{boxblue}{\textbf{Step 2: Voting Simulation.}}  
You are a persona with the following demographic characteristics: [demographics]. Your political ideology is described as [conservative-liberal spectrum]. The current year is [year]. Here are the policy agendas of the two parties: [Two parties' policy agenda]. Additionally, here are the presidential candidates' biographical and professional backgrounds:

[Presidential candidates' biographical and professional backgrounds].

Based on this information, please answer the following question:

\begin{enumerate}[leftmargin=*]
    \item As of today, will you vote for the Democratic Party (Hilary Clinton), the Republican Party (Donald Trump), or do you have no preference?
    \begin{itemize}[leftmargin=*]
        \item Democratic
        \item Republican
        \item No Preference
    \end{itemize}
\end{enumerate}
\end{promptbox}

\noindent \textbf{Evaluation Criteria.}
We design two different evaluation criteria to evaluate the bias in voting simulation and the quality of feature generation. For voting simulation, we calculate the ratio:  
\( \mathcal{R} = \frac{\text{Republican Votes}}{\text{Republican Votes} + \text{Democratic Votes}} \)
and compared the LLM-generated simulation results with actual outcomes from the 2016 American National Election Studies (ANES) dataset~\cite{ANES2016}. 

To evaluate feature generation, we constructed specific metrics for each LLM by comparing the generated political ideology features with the original ANES 2016 features. In each matrix, the x-axis represents the original political ideology as recorded in the ANES 2016 dataset, while the y-axis represents the political ideology features generated by different LLMs based on contextual information. The numbers 1 to 7 correspond to "Very Liberal," "Somewhat Liberal," "Closer to Liberal," "Moderate," "Closer to Conservative," "Somewhat Conservative," and "Very Conservative," respectively. These metrics provide a direct and quantitative comparison of model-generated outputs against ground truth, enabling us to assess the alignment and reliability of LLM predictions in the context of political science.

\subsubsection{Results Analysis and Performance Comparison}

The analysis focuses on two key aspects: (1) the biases exhibited by LLMs in voting simulation~\cite{qi2024representation,qu2024performance}, and (2) LLMs' performance in feature generation~\cite{majumdar2024generative,zhang2024electionsim,yang2024llm}. To provide a robust basis for evaluation, the ratio of party affiliation (e.g., Republicans to Democrats) is predicted based on samples from the 2016 ANES dataset. Each observation in the ratio corresponds to an individual voter with specific demographic and ideological labels derived from ANES. This setup ensures that the simulated voting distributions align with the demographic and political tendencies reflected in the ANES sample, providing a reliable benchmark.

\begin{figure}[htbp]
\centering
\includegraphics[width=0.85\linewidth]{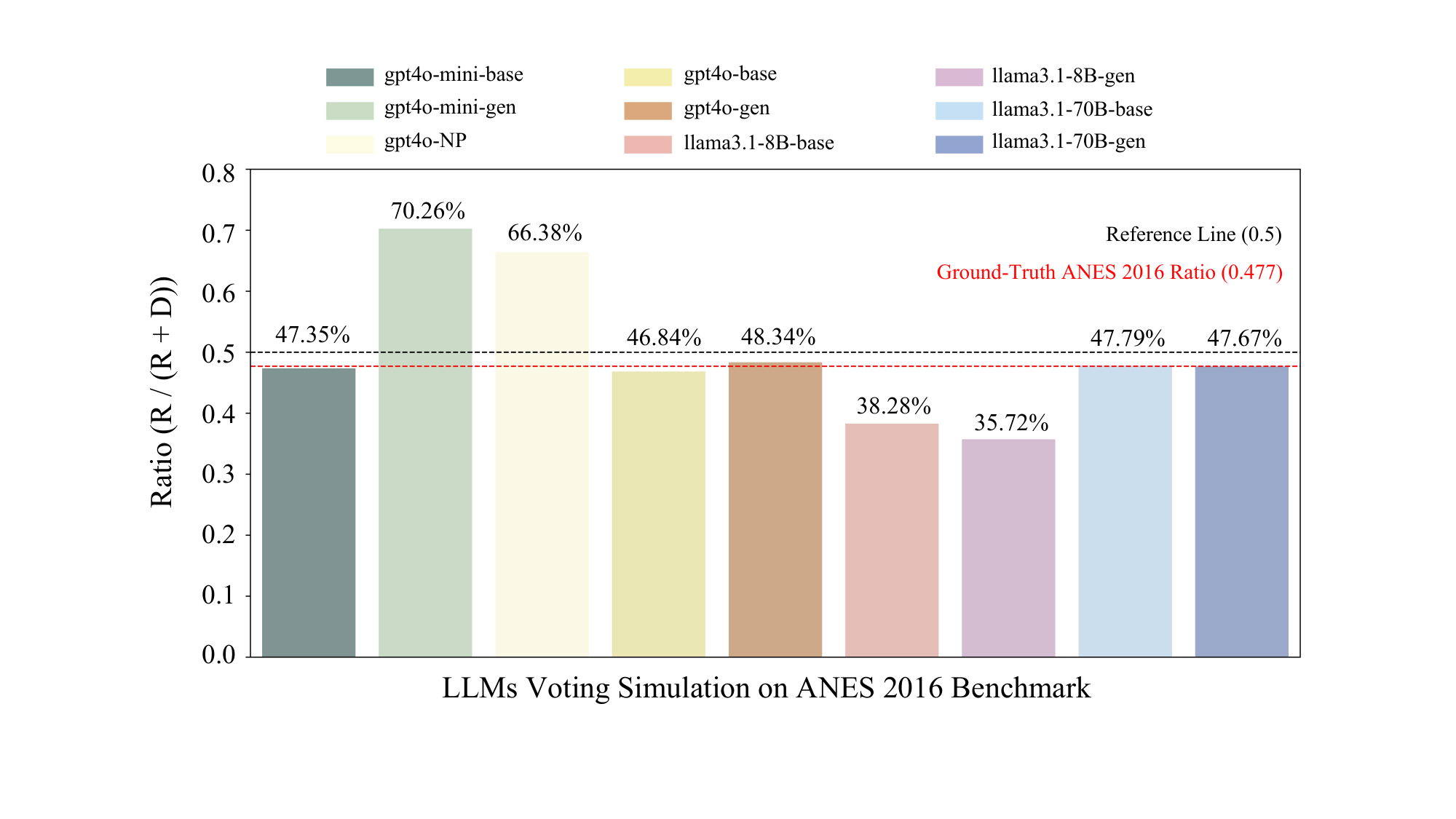}
\caption{The "base" labels represent the simulation results of different models using the complete ANES dataset features, while the "gen" labels represent results obtained by generating political ideology through the Chain of Thought approach and subsequently using the generated features for simulation. Additionally, "gpt4o-NP" denotes the simulation results of GPT-4o on the ANES dataset with political features removed. The red dashed line indicates the actual ratio derived from the 2016 ANES dataset, while the black dashed line represents the 50\% half-and-half reference point.}
\label{fig1_result}
\end{figure}

\noindent \textbf{Illustration of Voting Simulation Results.}
Figure~\ref{fig1_result} presents the voting simulation results across four models and eight pipeline variations on the 2016 benchmark dataset. Previous studies have shown that LLMs inherently possess political biases and that political features can mitigate these biases to some extent. To further illustrate this mitigation, we added an additional control group for GPT-4o, in which the model performed voting simulation on the ANES dataset without any political features. This yielded a total of nine results for comparison. Findings indicate that larger models, such as GPT-4o and Llama 3.1-70B, produced predictions closely aligned with the ground truth baseline of 47.7\% across both the baseline and generation pipelines. However, when political features were removed, GPT-4o displayed a significant skew towards the winning party of the 2016 election, highlighting the role of political features in bias correction. In contrast, smaller models displayed varied performance. GPT-4o-mini achieved similar accuracy to the larger models when using the original data but showed a pronounced skew towards the winning party in the 2016 election. Consistent with findings from prior studies \cite{potter2024hiddenpersuadersllmspolitical}, Llama 3.1-8B exhibited a tendency to avoid responses favoring Republican positions while being more permissive of responses supporting Democratic positions.

\begin{figure}[htbp]
\centering
\includegraphics[width=0.95\linewidth]{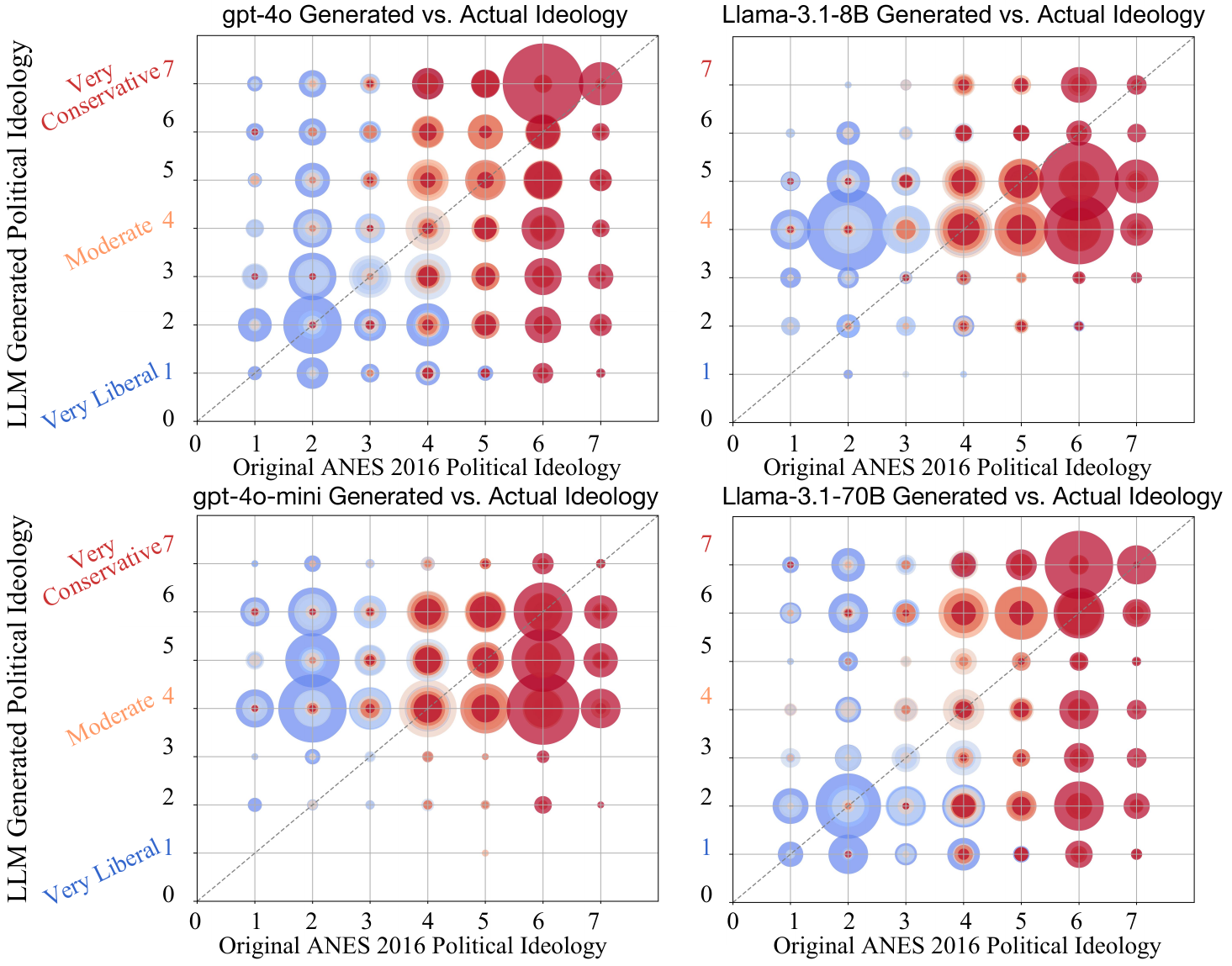}
\vspace{1.5ex}
\includegraphics[width=0.90\linewidth]{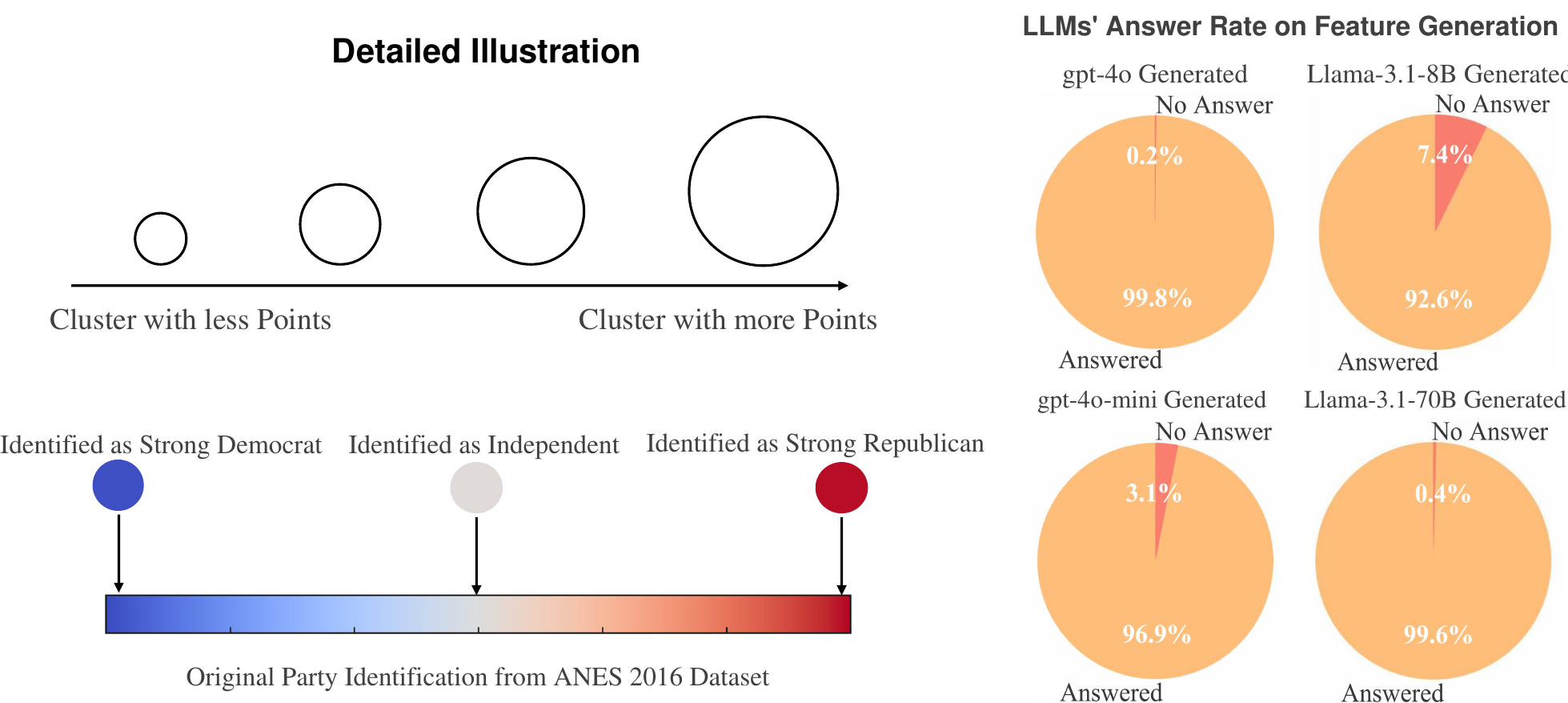}
\caption{The figure contains four 7x7 feature comparison matrices, corresponding illustrative diagrams, and four pie charts to evaluate the quality of feature generation by different LLMs in the field of political science.}
\label{fig2_results}
\end{figure}

\noindent \textbf{Comparison on Feature Generation Quality.}
In empirical surveys, missing or corrupted features are a persistent challenge, which makes the ability of LLMs to generate robust and accurate features particularly significant. For feature generation evaluation, Figure~\ref{fig2_results} highlights clear differences in generation capability across nine general-purpose LLMs. Specifically, GPT-4o and Llama 3.1-70B demonstrated higher generation quality, with their generated political ideology distributions closely matching those of the original ANES features. In Figure~\ref{fig2_results}, the size of each circle represents the relative quantity of data points, and we can observe that the majority of generation clusters for the larger models are concentrated along the diagonal line, indicating better alignment with the true feature distributions. In contrast, smaller models, such as GPT-4o-mini and Llama 3.1-8B, exhibited limited generation capabilities. Regardless of persona-specific features, these models consistently generated political ideologies aligned with the 2016 winning party, suggesting a limitation in their ability to accurately reflect diverse political perspectives. Besides, the pie charts summarize the proportion of responses in the feature generation process that were not labeled as "No Answer." GPT-4o and Llama 3.1-70B demonstrated high response rates of 99.8\% and 99.6\%, respectively. In contrast, smaller parameter models had relatively lower response rates, with GPT-4o-mini at 96.9\% and Llama 3.1-8B at 92.6\%.

Additionally, Figure~\ref{fig2_results} reveals some interesting insights regarding the ANES dataset itself. The color of the circles indicates the party identification of respondents, with red representing Republicans and blue representing Democrats. Notably, within the matrices, it can be observed that some respondents self-identified as strongly liberal but were registered as strong Republicans (represented by red circles located on the left side of the graph, where \(x < 4\)). Such inconsistent clusters highlight underlying inconsistencies in the dataset, which may influence model outcomes and suggest certain limitations in data collection practices. In Section~\ref{sec:challenges}, we will further explore these potential challenges and discuss directions for future work.

\section{Future Directions \& Challenges} \label{sec:challenges}

\subsection{Pipelines of Integrating Political Science with LLMs}

The integration of LLMs into political science research pipelines offers transformative opportunities, but it also presents several challenges. LLMs excel at automating tasks like policy analysis, election forecasting, and legislative summarization, yet adapting these general-purpose models to the nuanced demands of political science remains difficult. 

A significant challenge lies in LLMs' contextual understanding of domain-specific constructions, such as ideology, stance, and policy framing~\cite{he2023inducing}. To address this issue, future works must prioritize domain-adaptive pre-training or fine-tuning strategies that enhance LLMs' ability to process political texts effectively. In addition, hybrid workflows that combine LLM outputs with human expertise should be developed to ensure both reliability and interpretability~\cite{wu2023large}.

\noindent \textbf{Modularized Pipelines.} These pipelines are designed to decompose complex research tasks into manageable sub-components, each optimized for specific objectives, offering a promising solution to address the challenges. For instance, in election forecasting, a pipeline can include separate modules for data preprocessing (e.g., cleaning polling data and demographic information), contextual understanding (e.g., analyzing regional voting patterns), and predictive modeling (e.g., projecting voter turnout or swing state dynamics). Similarly, for legislative analysis, distinct modules can be employed to handle sub-tasks such as summarizing legal texts, extracting key policy themes, and evaluating potential societal impacts.

\noindent \textbf{Pipeline integrated with Retrieval-Augmented Generation.} Another promising approach is the integration of RAG into political science pipelines. RAG combines LLMs with external knowledge retrieval systems, enabling the dynamic incorporation of up-to-date and domain-specific information from political datasets, government databases, or news sources~\cite{khaliq2024ragar,salemi2024evaluating}. This method ensures that the model outputs remain contextually relevant, even in rapidly evolving political environments where timeliness is critical. For instance, in policy analysis, a RAG-based pipeline could retrieve recent legislative updates or party manifestos to supplement the LLM's generative capabilities, improving both accuracy and depth of insights.

\subsection{Data Scarcity and the Construction of Domain-Specific Datasets} 

% Data scarcity is a persistent challenge in CPS, particularly when working with LLMs that require extensive, high-quality labeled data. Unlike computer vision and general NLP fields, political science lacks large-scale datasets tailored for nuanced tasks such as election behavior modeling or policy framing analysis. Section~\ref{4-2-generative-task} highlighted data scarcity as a central challenge in LLM-based generative tasks. Future research should focus on developing robust, domain-specific datasets that capture the complexity of political discourse, as well as exploring advanced techniques for synthetic data generation. Retrieval-Augmented Generation (RAG)~\cite{alvarez2023generative} offers a promising strategy for dynamically retrieving external knowledge and generating synthetic examples aligned with real-world political contexts.

% Additionally, partnerships between political scientists, data scientists, and public institutions are essential for accessing valuable datasets, such as legislative records and public opinion surveys. Synthetic data generation could also address rare-event modeling, simulating underrepresented phenomena to enhance research on political minorities or emerging policy issues. Establishing standardized protocols for validating synthetic data will be crucial to maintaining accuracy and neutrality in LLM outputs.

Data scarcity remains a significant challenge in computational political science (CPS), especially when utilizing LLMs that rely on extensive, high-quality labeled data for effective performance. Unlike computer vision or general NLP, political science lacks large-scale, domain-specific datasets tailored for nuanced tasks such as election behavior modeling, legislative sentiment analysis, or policy framing studies. This limitation hinders the ability of LLMs to fully capture the complexity of political language and context, leading to inaccuracies or biases in model outputs. Furthermore, the dynamic nature of political discourse and the emergence of new events exacerbate the scarcity issue, making it difficult to maintain up-to-date datasets for research. 

\noindent \textbf{Developing High-Quality, Domain-Specific Datasets.}
One effective strategy to address data scarcity is the development of high-quality, domain-specific datasets curated explicitly for political science tasks. For instance, datasets could be constructed from political speeches, legislative records, or annotated campaign materials, enabling researchers to train LLMs on specialized tasks like policy analysis or ideological classification. A notable example is the "BillSum" dataset, which provides U.S. Congressional bills paired with human-generated summaries to support legislative text summarization~\cite{alvarez2023generative}. Such datasets help ensure that LLMs are exposed to diverse political contexts, enhancing their ability to process and analyze complex political texts accurately.

\noindent \textbf{Synthetic Data Generation using LLM-based Methods.}
A promising solution to data scarcity is the generation of synthetic datasets, which can simulate real-world political phenomena and expand the diversity of available data. By leveraging LLMs, researchers can create datasets that replicate complex political behaviors, such as voter turnout patterns, policy framing, or legislative negotiations. For example, synthetic election scenarios can be designed to model rare events, such as shifts in voting behavior during political crises or the impact of new campaign strategies. These datasets not only fill gaps in underrepresented areas but also enable researchers to test hypotheses that are otherwise difficult to explore due to the lack of empirical data. Ensuring the accuracy and neutrality of these synthetic datasets through rigorous validation protocols is essential to maintaining their reliability and applicability in political science research.

\noindent \textbf{Standardizing Validation Protocols for Synthetic Data.}
To ensure the reliability of synthetic datasets, standardized validation protocols must be established. These protocols should assess synthetic data for accuracy, neutrality, and representativeness. For example, in studies analyzing minority voting behavior, synthetic datasets should undergo rigorous checks to confirm that they accurately reflect the real-world dynamics of marginalized groups rather than amplifying stereotypes. Methods such as cross-validation with human-coded datasets or statistical comparisons to ground-truth data can ensure that synthetic examples align with empirical observations and minimize risks of bias or distortion in LLM outputs.

\noindent \textbf{Collaborative Partnerships for Data Access and Curation.} Collaboration among political scientists, data scientists, and public institutions is crucial for enhancing access to valuable data resources. Furthermore, partnerships with government agencies and local/international organizations can provide access to legislative records, public opinion surveys, and election data that are often difficult to obtain through individual efforts. For instance, a partnership with a national statistics agency can yield comprehensive datasets on voting behavior segmented by demographics and geographic regions, enriching research on electoral dynamics. Establishing standardized protocols for data collection and sharing will further enhance dataset quality and ensure balanced representation of diverse political perspectives.

\subsection{Addressing Bias and Fairness in Political Predictions}

The introduction of LLMs in CPS brings unique risks related to bias and fairness, as these models potentially reflect and amplify biases embedded in their training data. These biases may stem from over-representation of dominant political perspectives, exclusion of minority viewpoints, or inequalities in the underlying datasets~\cite{he2023inducing}. When deployed in sensitive political contexts such as voting behavior analysis or public sentiment modeling, biased predictions can skew insights, misrepresent trends, and even inadvertently influence public opinion or policymaking. This lack of fairness undermines the credibility of research findings and raises ethical concerns about the use of LLMs in these scenarios. Therefore, addressing bias and fairness is not only a technical challenge but also a critical requirement for ensuring the reliability and neutrality of political predictions.

\noindent \textbf{Knowledge Editing to Reduce Bias in Model Outputs.}
One effective approach to mitigating bias is knowledge editing, which adjusts specific model behaviors by re-training or fine-tuning on carefully curated datasets. This process involves systematically analyzing the model's outputs across diverse scenarios to identify recurring patterns of bias and then implementing targeted corrections. For example, a study on gender representation in political speech analysis can use a curated dataset that balances male and female political leaders' contributions, ensuring the model does not overemphasize one gender's discourse~\cite{wu2023large}. Knowledge editing ensures that model outputs are more representative, reducing the risk of perpetuating stereotypes or unbalanced narratives.

\noindent \textbf{Counterfactual Data Augmentation for Fairness.}
Counterfactual data augmentation is another promising solution that introduces synthetic examples to represent underrepresented perspectives or scenarios. In public opinion modeling, synthetic data can simulate viewpoints from minority or marginalized groups that are often absent from traditional datasets. By exposing LLMs to diverse perspectives during training, this method ensures that the models learn a more comprehensive representation of political discourse. In applications like policy framing analysis, counterfactual augmentation can enhance fairness by ensuring that outputs reflect a balance of ideological perspectives, helping policymakers avoid unintended biases in decision-making~\cite{alvarez2023generative}.

\noindent \textbf{Explainable AI for Transparency in Predictions.}
Transparency is also crucial aspect of addressing bias and ensuring fairness in political predictions. Explainable AI (XAI) methods can clarify how LLMs arrive at specific predictions, making their decision-making pathways interpretable to researchers and stakeholders~\cite{jiang2022communitylm}. For instance, when analyzing election forecasts, XAI tools can highlight the key feature attributes (e.g., polling trends or demographic statistics) that influenced the model’s predictions. This transparency enables researchers to detect and address potential biases in the underlying process, fostering trust in the model’s outputs. Additionally, XAI can facilitate collaboration between domain experts and data scientists, improving both the interpretability and accuracy of LLM-driven political analysis.

\subsection{Enhancing Explainability and Reducing Hallucination Risks} 

% Section~\ref{4-4-explainability} highlighted the promise of LLMs in advancing explainability and causal inference, yet significant hurdles remain. One pressing issue is hallucination, where LLMs generate plausible yet inaccurate outputs~\cite{argyle2023leveraging}. Future work should prioritize integrating validation checkpoints, wherein human experts cross-reference outputs against domain knowledge to detect inaccuracies.

% Improving LLM explainability through causal pathways is another vital area. Developing methods to identify the causal relationships that underpin model predictions will enhance their utility in policy and decision-making contexts. This could involve the incorporation of causal diagrams or explicit modeling of counterfactual scenarios to ensure alignment with empirical data.

The lack of explainability in LLMs poses a major challenge for their application in political science. These models often generate predictions or outputs without providing clear reasoning or insights into the underlying decision-making processes~\cite{argyle2023leveraging}. In tasks such as legislative sentiment analysis or election forecasting, the inability to trace how specific features (e.g., public opinion trends or demographic factors) contribute to the output limits the trust and utility of LLMs in high-stakes political research. This opacity hinders researchers and policymakers from validating or contextualizing the outputs, raising concerns about the reliability of these tools in guiding real-world political decisions.

Hallucination represents the output generated from LLMs that appears to be plausible but lacks grounding in factual data, representing a critical type of risk~\cite{argyle2023leveraging}. This issue is especially problematic in political science, where misinformation or inaccuracies can distort public understanding or influence policy decisions. To illustrate, hallucinations in the generation of political speeches or summaries of legislative texts could misrepresent key policy proposals, leading to flawed analyses or misinformed actions. The high stakes of political science applications demand robust strategies to mitigate these risks and ensure that LLMs produce outputs grounded in factual and reliable data.

\noindent \textbf{Improving Explainability through Feature Attribution.} One effective method to enhance explainability is feature attribution, which identifies the specific inputs that contribute most significantly to a model's predictions. Techniques such as Shapley Additive Explanations (SHAP)~\cite{antwarg2021explaining} or Integrated Gradients~\cite{lundstrom2022rigorous} can be used to highlight the influence of variables such as polling data, economic indicators, or demographic information in election forecasting~\cite{he2023inducing}. This can be seen in predicting voter turnout, where feature attribution reveals how regional trends impact turnout predictions, enabling researchers to better understand the model's reasoning. By making these connections explicit, feature attribution enhances transparency in LLM-assisted analysis.

\noindent \textbf{Leveraging Causal Modeling for Explainability.} Another approach to improving explainability is causal modeling, which explicitly maps the causal relationships underlying model predictions. For instance, causal diagrams can be used to represent how changes in one factor, such as campaign spending, influence another, like voter preferences~\cite{alvarez2023generative}. In policy impact analysis, integrating counterfactual scenarios—e.g., “What if a policy were implemented in a different region?”—can provide actionable insights while ensuring that the model's reasoning aligns with empirical data. By incorporating causal reasoning, LLMs can support more informed decision-making in complex political contexts.

\noindent \textbf{Enhancing Reliability through Uncertainty Quantification.} One effective method to enhance reliability is uncertainty quantification, which measures the uncertainty of the output from LLMs. Techniques such as Semantic Entropy~\cite{kuhn2023semantic} or consistency measurements~\cite{da2024llm,lin2023generating} can be used to profile the model’s behavior, including their intrinsic semantics features, linguistic ambiguity, and complex output structures~\cite{lin2023generating}. This can be seen in generating the results, where uncertainty in the generation reveals how consistent the model's predictions can be, enabling researchers to better trust the model's prediction. By making an explicit quantification of the generated contents, uncertainty quantification enhances reliability and builds trust in LLM-assisted generated contents.

\noindent \textbf{Incorporating Validation Checkpoints to Reduce Hallucinations.} To address hallucination risks, integrating validation checkpoints into LLM workflows is a promising strategy. These checkpoints involve periodic cross-referencing of LLM outputs against established datasets or domain expertise to ensure factual accuracy~\cite{argyle2023leveraging}. An example of this is in summarizing legislative bills, where validation checkpoints can require outputs to be verified against the original legislative text by human experts or through automated retrieval-augmented systems. The iterative process minimizes the likelihood of inaccuracies propagating through the analysis and ensures that outputs remain aligned with empirical evidence.

\subsection{Democratizing Access to Political Knowledge}

Democratizing access to political knowledge refers to the process of making political information, such as legislative texts, policy analyses, and electoral data, accessible and understandable to a broader audience. This goal seeks to address barriers such as the complexity of legal language, information asymmetry, and the marginalization of underrepresented groups in political discourse. Currently, these barriers prevent many citizens from engaging with and understanding political processes, limiting their participation in democratic systems. By leveraging LLMs, researchers aim to simplify complex political content, provide multilingual support, and adapt information to diverse cultural and ideological contexts, thereby enabling more inclusive and equitable access to political knowledge~\cite{stanczak2023quantifying}.

\noindent \textbf{Simplifying Political Language Through LLM-Driven Tools.} One effective strategy to democratize access is developing user-facing tools powered by LLMs to simplify complex political language. For example, interactive Q\&A systems can allow users to query legislative documents and receive concise, accurate, and accessible explanations. A practical application is the creation of voter guides that summarize candidates’ platforms or key legislative initiatives in plain language. Tools like these can help demystify complicated topics such as tax reforms or healthcare policies, making them more approachable for the general public~\cite{stanczak2023quantifying}. Ensuring that these simplifications maintain accuracy and neutrality is critical to preserving trust and effectiveness.

\noindent \textbf{Multilingual Support and Cultural Adaptation.} To overcome language and cultural barriers, LLMs should provide robust multilingual support and culturally adapted outputs. For instance, an LLM-powered platform could offer translations of legislative texts into minority languages while contextualizing the content to reflect local cultural norms. In a real-world application, such a system could support Indigenous or immigrant communities by delivering policy information in their native languages, enabling them to better understand and engage with democratic processes. Expanding multilingual and culturally sensitive capabilities ensures that political knowledge reaches diverse audiences, promoting fairness and inclusivity in political discourse~\cite{stanczak2023quantifying}.

\noindent \textbf{Incorporating Explainable and Ethical AI Principles.} Another critical measure is incorporating Explainable AI (XAI) and Ethical AI (EAI) principles into LLM applications for political science. These principles enhance the transparency and accountability of LLM outputs by clarifying how information is processed and ensuring that outputs align with ethical standards. For instance, XAI techniques can provide detailed explanations of how an LLM summarized a policy proposal, enabling users to trace the reasoning behind its conclusions. Similarly, EAI frameworks can ensure that outputs are unbiased and equitable, addressing ethical concerns and fostering trust among users~\cite{argyle2023leveraging}. By integrating these principles, political knowledge systems can balance accessibility with fairness, ensuring responsible use of AI in democratic contexts.

\subsection{Call for Novel Evaluation Criteria for Computational Political Science}

\noindent \textbf{The Importance of Comprehensive Evaluation Criteria.} Comprehensive and reasonable evaluation criteria are crucial for advancing computational political science (CPS). These criteria ensure that LLMs are not only judged by technical proficiency but also by their relevance and applicability in the political domain. While existing metrics such as accuracy, F1-score, and BLEU are widely used in NLP research, they often fail to address the nuanced requirements of CPS tasks~\cite{khurana2023natural}. In political science, models must go beyond technical performance to reflect deeper insights, such as their capacity to interpret policy implications, predict electoral dynamics, or analyze ideological framing. Without domain-specific evaluation metrics, the true potential and limitations of LLMs in political science may remain obscured.

Existing evaluation criteria in CPS often focus on generic NLP metrics, which usually includes:
\begin{itemize}[leftmargin=*]
\item {\textit{Accuracy and F1-Score}}: Widely used for classification tasks like sentiment analysis or ideological alignment detection, these metrics measure how well a model categorizes data into predefined labels.
\item {\textit{BLEU and ROUGE}}: Commonly applied to text generation tasks such as legislative summarization, these metrics evaluate the similarity between model-generated outputs and reference texts.
\item {\textit{Cross-Validation Performance}}: Utilized to assess the generalizability of models across different datasets, ensuring consistency in tasks like voter behavior prediction.
\end{itemize}

While these metrics provide valuable benchmarks, they fall short in capturing the real-world implications of LLM outputs in political contexts. For instance, BLEU may not adequately measure how well a generated legislative summary conveys critical policy details, and F1-score cannot evaluate whether sentiment analysis accurately reflects ideological subtleties.

\noindent \textbf{Proposed Novel Evaluation Framework for CPS.} To address the limitations of current metrics, future research should explore novel evaluation criteria tailored to the unique demands of CPS tasks. The evaluation framework should include metrics from these aspects:
\begin{enumerate}[label=\arabic*., leftmargin=*]
\item \textit{Policy Relevance:} This metric assesses how effectively a model contextualizes policy positions within legislative texts or debates. For example, when summarizing a healthcare policy, the model should highlight key trade-offs, stakeholder impacts, and alignment with political objectives. This is particularly valuable in tasks like legislative analysis, where the focus is on understanding the implications of proposed policies.

\item \textit{Electoral Impact:} Electoral impact measures a model's ability to accurately predict public opinion or campaign dynamics based on contextual variables such as demographic trends and polling data. For instance, in predicting swing state outcomes during U.S. elections, this metric can assess how well the model incorporates regional voting patterns and past election data~\cite{khurana2023natural}.

\item \textit{Legislative Influence:} Legislative influence evaluates the societal implications of proposed policies or legislative actions. For example, this metric can be used to analyze how well a model predicts the downstream effects of policy changes, such as shifts in public opinion, economic impact, or voting behavior. Researchers can leverage this metric to better understand the broader consequences of legislative decisions.

\item \textit{Fairness and Bias Metrics:} Given the importance of neutrality in political science, fairness and bias metrics should evaluate how balanced and inclusive LLM outputs are across diverse ideological perspectives. For example, when generating policy summaries, the model should avoid overemphasizing dominant viewpoints while marginalizing minority opinions, ensuring a fair representation of all stakeholders~\cite{khurana2023natural}.
\end{enumerate}

\section{Conclusion}
\label{sec:conclusion}

This survey provides the first comprehensive interdisciplinary exploration of the integration of LLMs into political science, bridging gaps between classical methodologies and modern computational approaches. We introduce a novel principled taxonomy that systematically categorizes political science tasks and LLM-driven methods, offering a structured framework to guide researchers in leveraging LLMs effectively. We begin by detailing LLM's capabilities across predictive modeling, generative tasks, simulation, causal inference, and societal impacts from political science perspective. Our analysis highlights both the transformative potential of LLMs and the pressing challenges they present, including data scarcity, biases, explainability limitations, and ethical concerns. Furthermore, we elaborate computational techniques from computer science perspective, including fine-tuning methods, LLM inference strategies, and the development of domain-specific benchmarks, tailored to political science applications. An empirical case study on ANES presidential election voting simulation exemplifies the practical implementation of these methods. Finally, we identify critical research directions, emphasizing the need for modular research pipelines, novel evaluation metrics, and robust approaches to bias mitigation and fairness enhancement. Our survey underscores the necessity of interdisciplinary collaboration and ethical considerations in advancing the use of LLMs in political science. By addressing the identified challenges, we aim to inspire further research that promotes the responsible and impactful application of LLMs, enabling more informed, transparent, and equitable political processes.

% \section{Acknowledgements}   \label{sec:acks}

\bibliographystyle{unsrt}
\bibliography{references}

\begin{thebibliography}{100}

\bibitem{wornow2023shaky}
Michael Wornow, Yizhe Xu, Rahul Thapa, Birju Patel, Ethan Steinberg, Scott Fleming, Michael~A Pfeffer, Jason Fries, and Nigam~H Shah.
\newblock The shaky foundations of large language models and foundation models for electronic health records.
\newblock {\em npj Digital Medicine}, 6(1):135, 2023.

\bibitem{wang2023large}
Yuqing Wang, Yun Zhao, and Linda Petzold.
\newblock Are large language models ready for healthcare? a comparative study on clinical language understanding.
\newblock In {\em Machine Learning for Healthcare Conference}, pages 804--823. PMLR, 2023.

\bibitem{xu2024retrieval}
Zerui Xu, Fang Wu, Tianfan Fu, and Yue Zhao.
\newblock Retrieval-reasoning large language model-based synthetic clinical trial generation.
\newblock {\em arXiv preprint arXiv:2410.12476}, 2024.

\bibitem{yue2024ct}
Ling Yue, Sixue Xing, Jintai Chen, and Tianfan Fu.
\newblock Clinicalagent: Clinical trial multi-agent with large language model-based reasoning.
\newblock {\em arXiv preprint arXiv:2404.14777}, 2024.

\bibitem{huang2023finbert}
Allen~H Huang, Hui Wang, and Yi~Yang.
\newblock Finbert: A large language model for extracting information from financial text.
\newblock {\em Contemporary Accounting Research}, 40(2):806--841, 2023.

\bibitem{wu2023bloomberggpt}
Shijie Wu, Ozan Irsoy, Steven Lu, Vadim Dabravolski, Mark Dredze, Sebastian Gehrmann, Prabhanjan Kambadur, David Rosenberg, and Gideon Mann.
\newblock Bloomberggpt: A large language model for finance.
\newblock {\em arXiv preprint arXiv:2303.17564}, 2023.

\bibitem{xie2024pixiu}
Qianqian Xie, Weiguang Han, Xiao Zhang, Yanzhao Lai, Min Peng, Alejandro Lopez-Lira, and Jimin Huang.
\newblock Pixiu: A comprehensive benchmark, instruction dataset and large language model for finance.
\newblock In {\em Advances in Neural Information Processing Systems}, 2024.

\bibitem{zhang2023artificial}
Xuan Zhang, Limei Wang, Jacob Helwig, Youzhi Luo, Cong Fu, Yaochen Xie, Meng Liu, Yuchao Lin, Zhao Xu, Keqiang Yan, Keir Adams, Maurice Weiler, Xiner Li, Tianfan Fu, Yucheng Wang, Haiyang Yu, YuQing Xie, Xiang Fu, Alex Strasser, Shenglong Xu, Yi~Liu, Yuanqi Du, Alexandra Saxton, Hongyi Ling, Hannah Lawrence, Hannes St{\"a}rk, Shurui Gui, Carl Edwards, Nicholas Gao, Adriana Ladera, Tailin Wu, Elyssa~F. Hofgard, Aria~Mansouri Tehrani, Rui Wang, Ameya Daigavane, Montgomery Bohde, Jerry Kurtin, Qian Huang, Tuong Phung, Minkai Xu, Chaitanya~K. Joshi, Simon~V. Mathis, Kamyar Azizzadenesheli, Ada Fang, Al{\'a}n Aspuru-Guzik, Erik Bekkers, Michael Bronstein, Marinka Zitnik, Anima Anandkumar, Stefano Ermon, Pietro Li{\`o}, Rose Yu, Stephan G{\"u}nnemann, Jure Leskovec, Heng Ji, Jimeng Sun, Regina Barzilay, Tommi Jaakkola, Connor~W. Coley, Xiaoning Qian, Xiaofeng Qian, Tess Smidt, and Shuiwang Ji.
\newblock Artificial intelligence for science in quantum, atomistic, and continuum systems.
\newblock {\em arXiv preprint arXiv:2307.08423}, 2023.

\bibitem{Zhang2024Survey}
Yu~Zhang, Xiusi Chen, Bowen Jin, Sheng Wang, Shuiwang Ji, Wei Wang, and Jiawei Han.
\newblock A comprehensive survey of scientific large language models and their applications in scientific discovery.
\newblock In {\em Conference on Empirical Methods in Natural Language Processing}, page 8783–8817, 2024.

\bibitem{liu2024drugagent}
Sizhe Liu, Yizhou Lu, Siyu Chen, Xiyang Hu, Jieyu Zhao, Tianfan Fu, and Yue Zhao.
\newblock Drugagent: Automating ai-aided drug discovery programming through llm multi-agent collaboration.
\newblock {\em arXiv}, 2024.

\bibitem{da2024open}
Longchao Da, Kuanru Liou, Tiejin Chen, Xuesong Zhou, Xiangyong Luo, Yezhou Yang, and Hua Wei.
\newblock Open-ti: Open traffic intelligence with augmented language model.
\newblock {\em International Journal of Machine Learning and Cybernetics}, pages 1--26, 2024.

\bibitem{da2024prompt}
Longchao Da, Minquan Gao, Hao Mei, and Hua Wei.
\newblock Prompt to transfer: Sim-to-real transfer for traffic signal control with prompt learning.
\newblock In {\em Proceedings of the AAAI Conference on Artificial Intelligence}, pages 82--90, 2024.

\bibitem{zhang2024urban}
Weijia Zhang, Jindong Han, Zhao Xu, Hang Ni, Hao Liu, and Hui Xiong.
\newblock Urban foundation models: A survey.
\newblock In {\em Proceedings of the 30th ACM SIGKDD Conference on Knowledge Discovery and Data Mining}, pages 6633--6643, 2024.

\bibitem{li2024opencity}
Zhonghang Li, Long Xia, Lei Shi, Yong Xu, Dawei Yin, and Chao Huang.
\newblock Opencity: Open spatio-temporal foundation models for traffic prediction.
\newblock {\em arXiv preprint arXiv:2408.10269}, 2024.

\bibitem{kasneci2023chatgpt}
Enkelejda Kasneci, Kathrin Se{\ss}ler, Stefan K{\"u}chemann, Maria Bannert, Daryna Dementieva, Frank Fischer, Urs Gasser, Georg Groh, Stephan G{\"u}nnemann, Eyke H{\"u}llermeier, et~al.
\newblock Chatgpt for good? on opportunities and challenges of large language models for education.
\newblock {\em Learning and individual differences}, 103:102274, 2023.

\bibitem{10.1145/3613372.3614197}
Gustavo Pinto, Isadora Cardoso-Pereira, Danilo Monteiro, Danilo Lucena, Alberto Souza, and Kiev Gama.
\newblock Large language models for education: Grading open-ended questions using chatgpt.
\newblock In {\em Proceedings of the XXXVII Brazilian Symposium on Software Engineering}, page 293–302, 2023.

\bibitem{henkel2024can}
Owen Henkel, Libby Hills, Adam Boxer, Bill Roberts, and Zach Levonian.
\newblock Can large language models make the grade? an empirical study evaluating llms ability to mark short answer questions in k-12 education.
\newblock In {\em Proceedings of the Eleventh ACM Conference on Learning@ Scale}, pages 300--304, 2024.

\bibitem{minaee2024large}
Shervin Minaee, Tomas Mikolov, Narjes Nikzad, Meysam Chenaghlu, Richard Socher, Xavier Amatriain, and Jianfeng Gao.
\newblock Large language models: A survey.
\newblock {\em arXiv preprint arXiv:2402.06196}, 2024.

\bibitem{zhao2023survey}
Wayne~Xin Zhao, Kun Zhou, Junyi Li, Tianyi Tang, Xiaolei Wang, Yupeng Hou, Yingqian Min, Beichen Zhang, Junjie Zhang, Zican Dong, et~al.
\newblock A survey of large language models.
\newblock {\em arXiv preprint arXiv:2303.18223}, 2023.

\bibitem{llmpolitical}
Mitchell Linegar, Rafal Kocielnik, and R.~Michael Alvarez.
\newblock Large language models and political science.
\newblock {\em Frontiers in Political Science}, 5, 2023.

\bibitem{argyle2023out}
Lisa~P Argyle, Ethan~C Busby, Nancy Fulda, Joshua~R Gubler, Christopher Rytting, and David Wingate.
\newblock Out of one, many: Using language models to simulate human samples.
\newblock {\em Political Analysis}, 31(3):337--351, 2023.

\bibitem{ziems2024can}
Caleb Ziems, William Held, Omar Shaikh, Jiaao Chen, Zhehao Zhang, and Diyi Yang.
\newblock Can large language models transform computational social science?
\newblock {\em Computational Linguistics}, 50(1):237--291, 2024.

\bibitem{demszky2023using}
Dorottya Demszky, Diyi Yang, David~S Yeager, Christopher~J Bryan, Margarett Clapper, Susannah Chandhok, Johannes~C Eichstaedt, Cameron Hecht, Jeremy Jamieson, Meghann Johnson, et~al.
\newblock Using large language models in psychology.
\newblock {\em Nature Reviews Psychology}, 2(11):688--701, 2023.

\bibitem{rotaru2024artificial}
George-Cristinel Rotaru, Sorin Anagnoste, and Vasile-Marian Oancea.
\newblock How artificial intelligence can influence elections: Analyzing the large language models (llms) political bias.
\newblock In {\em Proceedings of the International Conference on Business Excellence}, pages 1882--1891, 2024.

\bibitem{rodman2024political}
Emma Rodman.
\newblock On political theory and large language models.
\newblock {\em Political Theory}, 52(4):548--580, 2024.

\bibitem{gover2023political}
Lucas Gover.
\newblock Political bias in large language models.
\newblock {\em The Commons: Puget Sound Journal of Politics}, 4(1):2, 2023.

\bibitem{moe2005power}
Terry~M Moe.
\newblock Power and political institutions.
\newblock {\em Perspectives on politics}, 3(2):215--233, 2005.

\bibitem{gao2022post}
Chenxi Gao, Yini Li, et~al.
\newblock Post-war development analysis of political science: from behaviorism to new institutionalism: Political science development trend, challenges and suggestions.
\newblock {\em International Journal of Frontiers in Sociology}, 4(8), 2022.

\bibitem{wilkerson2017large}
John Wilkerson and Andreu Casas.
\newblock Large-scale computerized text analysis in political science: Opportunities and challenges.
\newblock {\em Annual Review of Political Science}, 20(1):529--544, 2017.

\bibitem{chen2022election2020}
Emily Chen, Ashok Deb, and Emilio Ferrara.
\newblock \# election2020: the first public twitter dataset on the 2020 us presidential election.
\newblock {\em Journal of Computational Social Science}, pages 1--18, 2022.

\bibitem{gerring2017qualitative}
John Gerring.
\newblock Qualitative methods.
\newblock {\em Annual Review of Political Science}, 20(1):15--36, 2017.

\bibitem{tornberg2023chatgpt}
Petter T{\"o}rnberg.
\newblock Chatgpt-4 outperforms experts and crowd workers in annotating political twitter messages with zero-shot learning.
\newblock {\em arXiv preprint arXiv:2304.06588}, 2023.

\bibitem{heseltine2024large}
Michael Heseltine and Bernhard Clemm~von Hohenberg.
\newblock Large language models as a substitute for human experts in annotating political text.
\newblock {\em Research \& Politics}, 11(1):20531680241236239, 2024.

\bibitem{liu2023summary}
Yiheng Liu, Tianle Han, Siyuan Ma, Jiayue Zhang, Yuanyuan Yang, Jiaming Tian, Hao He, Antong Li, Mengshen He, Zhengliang Liu, et~al.
\newblock Summary of chatgpt-related research and perspective towards the future of large language models.
\newblock {\em Meta-Radiology}, page 100017, 2023.

\bibitem{xu2024secap}
Yaoxun Xu, Hangting Chen, Jianwei Yu, Qiaochu Huang, Zhiyong Wu, Shi-Xiong Zhang, Guangzhi Li, Yi~Luo, and Rongzhi Gu.
\newblock Secap: Speech emotion captioning with large language model.
\newblock In {\em Proceedings of the AAAI Conference on Artificial Intelligence}, pages 19323--19331, 2024.

\bibitem{yue2023disc}
Shengbin Yue, Wei Chen, Siyuan Wang, Bingxuan Li, Chenchen Shen, Shujun Liu, Yuxuan Zhou, Yao Xiao, Song Yun, Xuanjing Huang, et~al.
\newblock Disc-lawllm: Fine-tuning large language models for intelligent legal services.
\newblock {\em arXiv preprint arXiv:2309.11325}, 2023.

\bibitem{gesnouin2024llamandement}
Joseph Gesnouin, Yannis Tannier, Christophe~Gomes Da~Silva, Hatim Tapory, Camille Brier, Hugo Simon, Raphael Rozenberg, Hermann Woehrel, Mehdi~El Yakaabi, Thomas Binder, et~al.
\newblock Llamandement: Large language models for summarization of french legislative proposals.
\newblock {\em arXiv preprint arXiv:2401.16182}, 2024.

\bibitem{tornberg2023simulating}
Petter T{\"o}rnberg, Diliara Valeeva, Justus Uitermark, and Christopher Bail.
\newblock Simulating social media using large language models to evaluate alternative news feed algorithms.
\newblock {\em arXiv preprint arXiv:2310.05984}, 2023.

\bibitem{najafi2024turkishbertweet}
Ali Najafi and Onur Varol.
\newblock Turkishbertweet: Fast and reliable large language model for social media analysis.
\newblock {\em Expert Systems with Applications}, 255:124737, 2024.

\bibitem{zhang2024benchmarking}
Tianyi Zhang, Faisal Ladhak, Esin Durmus, Percy Liang, Kathleen McKeown, and Tatsunori~B Hashimoto.
\newblock Benchmarking large language models for news summarization.
\newblock {\em Transactions of the Association for Computational Linguistics}, 12:39--57, 2024.

\bibitem{fang2024bias}
Xiao Fang, Shangkun Che, Minjia Mao, Hongzhe Zhang, Ming Zhao, and Xiaohang Zhao.
\newblock Bias of ai-generated content: an examination of news produced by large language models.
\newblock {\em Scientific Reports}, 14(1):5224, 2024.

\bibitem{rozado2024political}
David Rozado.
\newblock The political preferences of llms.
\newblock {\em arXiv preprint arXiv:2402.01789}, 2024.

\bibitem{breum2024persuasive}
Simon~Martin Breum, Daniel~V{\ae}dele Egdal, Victor~Gram Mortensen, Anders~Giovanni M{\o}ller, and Luca~Maria Aiello.
\newblock The persuasive power of large language models.
\newblock In {\em Proceedings of the International AAAI Conference on Web and Social Media}, volume~18, pages 152--163, 2024.

\bibitem{rivera2024escalation}
Juan-Pablo Rivera, Gabriel Mukobi, Anka Reuel, Max Lamparth, Chandler Smith, and Jacquelyn Schneider.
\newblock Escalation risks from language models in military and diplomatic decision-making.
\newblock In {\em The 2024 ACM Conference on Fairness, Accountability, and Transparency}, pages 836--898, 2024.

\bibitem{gujral2024can}
Pratik Gujral, Kshitij Awaldhi, Navya Jain, Bhavuk Bhandula, and Abhijnan Chakraborty.
\newblock Can llms help predict elections?(counter) evidence from the world's largest democracy.
\newblock {\em arXiv preprint arXiv:2405.07828}, 2024.

\bibitem{zhang2023sentiment}
Wenxuan Zhang, Yue Deng, Bing Liu, Sinno~Jialin Pan, and Lidong Bing.
\newblock Sentiment analysis in the era of large language models: A reality check.
\newblock {\em arXiv preprint arXiv:2305.15005}, 2023.

\bibitem{khan2023social}
Asif Khan, Nada Boudjellal, Huaping Zhang, Arshad Ahmad, and Maqbool Khan.
\newblock From social media to ballot box: Leveraging location-aware sentiment analysis for election predictions.
\newblock {\em Computers, Materials \& Continua}, 77(3), 2023.

\bibitem{zhang2024electionsim}
Xinnong Zhang, Jiayu Lin, Libo Sun, Weihong Qi, Yihang Yang, Yue Chen, Hanjia Lyu, Xinyi Mou, Siming Chen, Jiebo Luo, et~al.
\newblock Electionsim: Massive population election simulation powered by large language model driven agents.
\newblock {\em arXiv preprint arXiv:2410.20746}, 2024.

\bibitem{potter2024hidden}
Yujin Potter, Shiyang Lai, Junsol Kim, James Evans, and Dawn Song.
\newblock Hidden persuaders: Llms' political leaning and their influence on voters.
\newblock {\em arXiv preprint arXiv:2410.24190}, 2024.

\bibitem{santurkar2023whose}
Shibani Santurkar, Esin Durmus, Faisal Ladhak, Cinoo Lee, Percy Liang, and Tatsunori Hashimoto.
\newblock Whose opinions do language models reflect?
\newblock In {\em International Conference on Machine Learning}, pages 29971--30004, 2023.

\bibitem{bremer2023public}
Bj{\"o}rn Bremer and Reto B{\"u}rgisser.
\newblock Public opinion on welfare state recalibration in times of austerity: Evidence from survey experiments.
\newblock {\em Political Science Research and Methods}, 11(1):34--52, 2023.

\bibitem{wang2024intelligent}
Zhenyu Wang, Yi~Xu, Dequan Wang, Lingfeng Zhou, and Yiqi Zhou.
\newblock Intelligent computing social modeling and methodological innovations in political science in the era of large language models.
\newblock {\em arXiv preprint arXiv:2410.16301}, 2024.

\bibitem{halterman2024codebook}
Andrew Halterman and Katherine~A Keith.
\newblock Codebook llms: Adapting political science codebooks for llm use and adapting llms to follow codebooks.
\newblock {\em arXiv preprint arXiv:2407.10747}, 2024.

\bibitem{mou2024unifying}
Xinyi Mou, Zejun Li, Hanjia Lyu, Jiebo Luo, and Zhongyu Wei.
\newblock Unifying local and global knowledge: Empowering large language models as political experts with knowledge graphs.
\newblock In {\em Proceedings of the ACM on Web Conference 2024}, pages 2603--2614, 2024.

\bibitem{baker2024simulating}
Zachary~R Baker and Zarif~L Azher.
\newblock Simulating the us senate: An llm-driven agent approach to modeling legislative behavior and bipartisanship.
\newblock {\em arXiv preprint arXiv:2406.18702}, 2024.

\bibitem{chen2024susceptible}
Kai Chen, Zihao He, Jun Yan, Taiwei Shi, and Kristina Lerman.
\newblock How susceptible are large language models to ideological manipulation?
\newblock {\em arXiv preprint arXiv:2402.11725}, 2024.

\bibitem{he2023inducing}
Zihao He, Siyi Guo, Ashwin Rao, and Kristina Lerman.
\newblock Inducing political bias allows language models anticipate partisan reactions to controversies.
\newblock {\em arXiv preprint arXiv:2311.09687}, 2023.

\bibitem{yao2023llm}
Jia-Yu Yao, Kun-Peng Ning, Zhen-Hui Liu, Mu-Nan Ning, Yu-Yang Liu, and Li~Yuan.
\newblock Llm lies: Hallucinations are not bugs, but features as adversarial examples.
\newblock {\em arXiv preprint arXiv:2310.01469}, 2023.

\bibitem{yao2024survey}
Yifan Yao, Jinhao Duan, Kaidi Xu, Yuanfang Cai, Zhibo Sun, and Yue Zhang.
\newblock A survey on large language model (llm) security and privacy: The good, the bad, and the ugly.
\newblock {\em High-Confidence Computing}, page 100211, 2024.

\bibitem{marino2024integrating}
Giada Marino and Fabio Giglietto.
\newblock Integrating large language models in political discourse studies on social media: Challenges of validating an llms-in-the-loop pipeline.
\newblock {\em Sociologica}, 18(2):87--107, 2024.

\bibitem{de2021editing}
Nicola De~Cao, Wilker Aziz, and Ivan Titov.
\newblock Editing factual knowledge in language models.
\newblock {\em arXiv preprint arXiv:2104.08164}, 2021.

\bibitem{wang2023knowledge}
Song Wang, Yaochen Zhu, Haochen Liu, Zaiyi Zheng, Chen Chen, and Jundong Li.
\newblock Knowledge editing for large language models: A survey.
\newblock {\em ACM Computing Surveys}, 2023.

\bibitem{liu2024rethinking}
Sijia Liu, Yuanshun Yao, Jinghan Jia, Stephen Casper, Nathalie Baracaldo, Peter Hase, Yuguang Yao, Chris~Yuhao Liu, Xiaojun Xu, Hang Li, et~al.
\newblock Rethinking machine unlearning for large language models.
\newblock {\em arXiv preprint arXiv:2402.08787}, 2024.

\bibitem{liu2024towards}
Zheyuan Liu, Guangyao Dou, Zhaoxuan Tan, Yijun Tian, and Meng Jiang.
\newblock Towards safer large language models through machine unlearning.
\newblock {\em arXiv preprint arXiv:2402.10058}, 2024.

\bibitem{liu2024llm}
Zhengliang Liu, Yiwei Li, Oleksandra Zolotarevych, Rongwei Yang, and Tianming Liu.
\newblock Llm-potus score: A framework of analyzing presidential debates with large language models.
\newblock {\em arXiv preprint arXiv:2409.08147}, 2024.

\bibitem{motoki2024more}
Fabio Motoki, Valdemar Pinho~Neto, and Victor Rodrigues.
\newblock More human than human: Measuring chatgpt political bias.
\newblock {\em Public Choice}, 198(1):3--23, 2024.

\bibitem{linegar2023large}
Mitchell Linegar, Rafal Kocielnik, and R~Michael Alvarez.
\newblock Large language models and political science.
\newblock {\em Frontiers in Political Science}, 5:1257092, 2023.

\bibitem{liu2024poliprompt}
Menglin Liu and Ge~Shi.
\newblock Poliprompt: A high-performance cost-effective llm-based text classification framework for political science.
\newblock {\em arXiv preprint arXiv:2409.01466}, 2024.

\bibitem{kato2024u}
Ken Kato, Annabelle Purnomo, Christopher Cochrane, and Raeid Saqur.
\newblock L (u) pin: Llm-based political ideology nowcasting.
\newblock {\em arXiv preprint arXiv:2405.07320}, 2024.

\bibitem{yang2024llm}
Joshua~C Yang, Marcin Korecki, Damian Dailisan, Carina~I Hausladen, and Dirk Helbing.
\newblock Llm voting: Human choices and ai collective decision making.
\newblock {\em arXiv preprint arXiv:2402.01766}, 2024.

\bibitem{liu2024dellma}
Ollie Liu, Deqing Fu, Dani Yogatama, and Willie Neiswanger.
\newblock Dellma: Decision making under uncertainty with large language models.
\newblock {\em arXiv preprint arXiv:2402.02392}, 2024.

\bibitem{chatsiou2020deep}
Kakia Chatsiou and Slava~Jankin Mikhaylov.
\newblock Deep learning for political science.
\newblock {\em The SAGE handbook of research methods in political science and international relations}, pages 1053--1078, 2020.

\bibitem{wu2023large}
Patrick~Y Wu, Joshua~A Tucker, Jonathan Nagler, and Solomon Messing.
\newblock Large language models can be used to estimate the ideologies of politicians in a zero-shot learning setting.
\newblock {\em arXiv preprint arXiv:2303.12057}, 2023.

\bibitem{chalkidis2024investigating}
Ilias Chalkidis.
\newblock Investigating llms as voting assistants via contextual augmentation: A case study on the european parliament elections 2024.
\newblock {\em arXiv preprint arXiv:2407.08495}, 2024.

\bibitem{moghimifar2024modelling}
Farhad Moghimifar, Yuan-Fang Li, Robert Thomson, and Gholamreza Haffari.
\newblock Modelling political coalition negotiations using llm-based agents.
\newblock {\em arXiv preprint arXiv:2402.11712}, 2024.

\bibitem{sanders2023demonstrations}
Nathan~E Sanders, Alex Ulinich, and Bruce Schneier.
\newblock Demonstrations of the potential of ai-based political issue polling.
\newblock {\em arXiv preprint arXiv:2307.04781}, 2023.

\bibitem{hackenburg2023comparing}
Kobi Hackenburg, Lujain Ibrahim, Ben~M Tappin, and Manos Tsakiris.
\newblock Comparing the persuasiveness of role-playing large language models and human experts on polarized us political issues.
\newblock {\em OSF Preprints}, 10, 2023.

\bibitem{lazar2024can}
Seth Lazar and Lorenzo Manuali.
\newblock Can llms advance democratic values?
\newblock {\em arXiv preprint arXiv:2410.08418}, 2024.

\bibitem{gudino2024large}
Jairo~F Gudi{\~n}o, Umberto Grandi, and C{\'e}sar Hidalgo.
\newblock Large language models (llms) as agents for augmented democracy.
\newblock {\em Philosophical Transactions A}, 382(2285):20240100, 2024.

\bibitem{terechshenko2020comparison}
Zhanna Terechshenko, Fridolin Linder, Vishakh Padmakumar, Michael Liu, Jonathan Nagler, Joshua~A Tucker, and Richard Bonneau.
\newblock A comparison of methods in political science text classification: Transfer learning language models for politics.
\newblock {\em Available at SSRN 3724644}, 2020.

\bibitem{lee2024applications}
Kyuwon Lee, Simone Paci, Jeongmin Park, Hye~Young You, and Sylvan Zheng.
\newblock Applications of gpt in political science research, 2024.

\bibitem{rozado2023political}
David Rozado.
\newblock The political biases of chatgpt.
\newblock {\em Social Sciences}, 12(3):148, 2023.

\bibitem{ornstein2022train}
Joseph~T Ornstein, Elise~N Blasingame, and Jake~S Truscott.
\newblock How to train your stochastic parrot: Large language models for political texts.
\newblock Technical report, Working Paper, 2022.

\bibitem{weidinger2021ethical}
Laura Weidinger, John Mellor, Maribeth Rauh, Conor Griffin, Jonathan Uesato, Po-Sen Huang, Myra Cheng, Mia Glaese, Borja Balle, Atoosa Kasirzadeh, et~al.
\newblock Ethical and social risks of harm from language models.
\newblock {\em arXiv preprint arXiv:2112.04359}, 2021.

\bibitem{haq2020survey}
Ehsan~Ul Haq, Tristan Braud, Young~D Kwon, and Pan Hui.
\newblock A survey on computational politics.
\newblock {\em IEEE Access}, 8:197379--197406, 2020.

\bibitem{grimmer2021machine}
Justin Grimmer, Margaret~E Roberts, and Brandon~M Stewart.
\newblock Machine learning for social science: An agnostic approach.
\newblock {\em Annual Review of Political Science}, 24(1):395--419, 2021.

\bibitem{nicolau2007analysis}
Jairo Nicolau.
\newblock An analysis of the 2002 presidential elections using logistic regression.
\newblock {\em Brazilian political science review}, 1(1):125--135, 2007.

\bibitem{d2014separating}
Vito d'Orazio, Steven~T Landis, Glenn Palmer, and Philip Schrodt.
\newblock Separating the wheat from the chaff: Applications of automated document classification using support vector machines.
\newblock {\em Political analysis}, 22(2):224--242, 2014.

\bibitem{mikolov2013distributed}
Tomas Mikolov, Ilya Sutskever, Kai Chen, Greg~S Corrado, and Jeff Dean.
\newblock Distributed representations of words and phrases and their compositionality.
\newblock {\em Advances in neural information processing systems}, 26, 2013.

\bibitem{devlin2018bert}
Jacob Devlin.
\newblock Bert: Pre-training of deep bidirectional transformers for language understanding.
\newblock {\em arXiv preprint arXiv:1810.04805}, 2018.

\bibitem{vaswani2017attention}
Ashish Vaswani, Noam Shazeer, Niki Parmar, Jakob Uszkoreit, Llion Jones, Aidan~N. Gomez, Łukasz Kaiser, and Illia Polosukhin.
\newblock Attention is all you need.
\newblock In {\em The Thirty-first Annual Conference on Neural Information Processing Systems}, 2017.

\bibitem{brown2020languagemodelsfewshotlearners}
Tom~B. Brown, Benjamin Mann, Nick Ryder, and Others.
\newblock Language models are few-shot learners, 2020.

\bibitem{raffel2020exploring}
Colin Raffel, Noam Shazeer, Adam Roberts, Katherine Lee, Sharan Narang, Michael Matena, Yanqi Zhou, Wei Li, and Peter~J Liu.
\newblock Exploring the limits of transfer learning with a unified text-to-text transformer.
\newblock {\em Journal of machine learning research}, 21(140):1--67, 2020.

\bibitem{salemi2024evaluating}
Alireza Salemi and Hamed Zamani.
\newblock Evaluating retrieval quality in retrieval-augmented generation.
\newblock In {\em Proceedings of the 47th International ACM SIGIR Conference on Research and Development in Information Retrieval}, pages 2395--2400, 2024.

\bibitem{kirk2024improving}
James~R Kirk, Robert~E Wray, Peter Lindes, and John~E Laird.
\newblock Improving knowledge extraction from llms for task learning through agent analysis.
\newblock In {\em Proceedings of the AAAI Conference on Artificial Intelligence}, pages 18390--18398, 2024.

\bibitem{song2024low}
Lin Song, Yukang Chen, Shuai Yang, Xiaohan Ding, Yixiao Ge, Ying-Cong Chen, and Ying Shan.
\newblock Low-rank approximation for sparse attention in multi-modal llms.
\newblock In {\em Proceedings of the IEEE/CVF Conference on Computer Vision and Pattern Recognition}, pages 13763--13773, 2024.

\bibitem{zhu2024sampleattention}
Qianchao Zhu, Jiangfei Duan, Chang Chen, Siran Liu, Xiuhong Li, Guanyu Feng, Xin Lv, Huanqi Cao, Xiao Chuanfu, Xingcheng Zhang, et~al.
\newblock Sampleattention: Near-lossless acceleration of long context llm inference with adaptive structured sparse attention.
\newblock {\em arXiv preprint arXiv:2406.15486}, 2024.

\bibitem{touvron2023llamaopenefficientfoundation}
Hugo Touvron, Thibaut Lavril, Gautier Izacard, Xavier Martinet, Marie-Anne Lachaux, Timothée Lacroix, Baptiste Rozière, Naman Goyal, Eric Hambro, Faisal Azhar, Aurelien Rodriguez, Armand Joulin, Edouard Grave, and Guillaume Lample.
\newblock Llama: Open and efficient foundation language models, 2023.

\bibitem{achiam2023gpt}
Josh Achiam, Steven Adler, Sandhini Agarwal, Lama Ahmad, Ilge Akkaya, Florencia~Leoni Aleman, Diogo Almeida, Janko Altenschmidt, Sam Altman, Shyamal Anadkat, et~al.
\newblock Gpt-4 technical report.
\newblock {\em arXiv preprint arXiv:2303.08774}, 2023.

\bibitem{schuurmans2024autoregressive}
Dale Schuurmans, Hanjun Dai, and Francesco Zanini.
\newblock Autoregressive large language models are computationally universal.
\newblock {\em arXiv preprint arXiv:2410.03170}, 2024.

\bibitem{nozza2020mask}
Debora Nozza, Federico Bianchi, and Dirk Hovy.
\newblock What the [mask]? making sense of language-specific bert models.
\newblock {\em arXiv preprint arXiv:2003.02912}, 2020.

\bibitem{li2023label}
Zongxi Li, Xianming Li, Yuzhang Liu, Haoran Xie, Jing Li, Fu-lee Wang, Qing Li, and Xiaoqin Zhong.
\newblock Label supervised llama finetuning.
\newblock {\em arXiv preprint arXiv:2310.01208}, 2023.

\bibitem{zhang2023instruction}
Shengyu Zhang, Linfeng Dong, Xiaoya Li, Sen Zhang, Xiaofei Sun, Shuhe Wang, Jiwei Li, Runyi Hu, Tianwei Zhang, Fei Wu, et~al.
\newblock Instruction tuning for large language models: A survey.
\newblock {\em arXiv preprint arXiv:2308.10792}, 2023.

\bibitem{leerlaif}
Harrison Lee, Samrat Phatale, Hassan Mansoor, Thomas Mesnard, Johan Ferret, Kellie~Ren Lu, Colton Bishop, Ethan Hall, Victor Carbune, Abhinav Rastogi, et~al.
\newblock Rlaif vs. rlhf: Scaling reinforcement learning from human feedback with ai feedback.
\newblock In {\em Forty-first International Conference on Machine Learning}, 2024.

\bibitem{kojima2022large}
Takeshi Kojima, Shixiang~Shane Gu, Machel Reid, Yutaka Matsuo, and Yusuke Iwasawa.
\newblock Large language models are zero-shot reasoners.
\newblock {\em Advances in neural information processing systems}, 35:22199--22213, 2022.

\bibitem{perez2021true}
Ethan Perez, Douwe Kiela, and Kyunghyun Cho.
\newblock True few-shot learning with language models.
\newblock {\em Advances in neural information processing systems}, 34:11054--11070, 2021.

\bibitem{ram2023context}
Ori Ram, Yoav Levine, Itay Dalmedigos, Dor Muhlgay, Amnon Shashua, Kevin Leyton-Brown, and Yoav Shoham.
\newblock In-context retrieval-augmented language models.
\newblock {\em Transactions of the Association for Computational Linguistics}, 11:1316--1331, 2023.

\bibitem{prabhu2024pedal}
Sumanth Prabhu.
\newblock Pedal: Enhancing greedy decoding with large language models using diverse exemplars.
\newblock {\em arXiv preprint arXiv:2408.08869}, 2024.

\bibitem{xie2024self}
Yuxi Xie, Kenji Kawaguchi, Yiran Zhao, James~Xu Zhao, Min-Yen Kan, Junxian He, and Michael Xie.
\newblock Self-evaluation guided beam search for reasoning.
\newblock {\em Advances in Neural Information Processing Systems}, 36, 2024.

\bibitem{grubisic2024priority}
Dejan Grubisic, Volker Seeker, Gabriel Synnaeve, Hugh Leather, John Mellor-Crummey, and Chris Cummins.
\newblock Priority sampling of large language models for compilers.
\newblock In {\em Proceedings of the 4th Workshop on Machine Learning and Systems}, pages 91--97, 2024.

\bibitem{white2023prompt}
Jules White, Quchen Fu, Sam Hays, Michael Sandborn, Carlos Olea, Henry Gilbert, Ashraf Elnashar, Jesse Spencer-Smith, and Douglas~C Schmidt.
\newblock A prompt pattern catalog to enhance prompt engineering with chatgpt.
\newblock {\em arXiv preprint arXiv:2302.11382}, 2023.

\bibitem{yao2024tree}
Shunyu Yao, Dian Yu, Jeffrey Zhao, Izhak Shafran, Tom Griffiths, Yuan Cao, and Karthik Narasimhan.
\newblock Tree of thoughts: Deliberate problem solving with large language models.
\newblock {\em Advances in Neural Information Processing Systems}, 36, 2024.

\bibitem{martino2023knowledge}
Ariana Martino, Michael Iannelli, and Coleen Truong.
\newblock Knowledge injection to counter large language model (llm) hallucination.
\newblock In {\em European Semantic Web Conference}, pages 182--185. Springer, 2023.

\bibitem{narayanan2021efficient}
Deepak Narayanan, Mohammad Shoeybi, Jared Casper, et~al.
\newblock Efficient large-scale language model training on gpu clusters using megatron-lm.
\newblock In {\em Proceedings of the International Conference for High Performance Computing, Networking, Storage and Analysis}, pages 1--15, 2021.

\bibitem{ding2023parameter}
Ning Ding, Yujia Qin, Guang Yang, Fuchao Wei, Zonghan Yang, Yusheng Su, Shengding Hu, Yulin Chen, Chi-Min Chan, Weize Chen, et~al.
\newblock Parameter-efficient fine-tuning of large-scale pre-trained language models.
\newblock {\em Nature Machine Intelligence}, 5(3):220--235, 2023.

\bibitem{wu2023fast}
Bingyang Wu, Yinmin Zhong, Zili Zhang, Shengyu Liu, Fangyue Liu, Yuanhang Sun, Gang Huang, Xuanzhe Liu, and Xin Jin.
\newblock Fast distributed inference serving for large language models.
\newblock {\em arXiv preprint arXiv:2305.05920}, 2023.

\bibitem{song2024powerinfer}
Yixin Song, Zeyu Mi, Haotong Xie, and Haibo Chen.
\newblock Powerinfer: Fast large language model serving with a consumer-grade gpu.
\newblock In {\em Proceedings of the ACM SIGOPS 30th Symposium on Operating Systems Principles}, pages 590--606, 2024.

\bibitem{ren2024melora}
Pengjie Ren, Chengshun Shi, Shiguang Wu, Mengqi Zhang, Zhaochun Ren, Maarten Rijke, Zhumin Chen, and Jiahuan Pei.
\newblock Melora: Mini-ensemble low-rank adapters for parameter-efficient fine-tuning.
\newblock In {\em Proceedings of the 62nd Annual Meeting of the Association for Computational Linguistics (Volume 1: Long Papers)}, pages 3052--3064, 2024.

\bibitem{fu2023effectiveness}
Zihao Fu, Haoran Yang, Anthony Man-Cho So, Wai Lam, Lidong Bing, and Nigel Collier.
\newblock On the effectiveness of parameter-efficient fine-tuning.
\newblock In {\em Proceedings of the AAAI conference on artificial intelligence}, pages 12799--12807, 2023.

\bibitem{kwon2023efficient}
Woosuk Kwon, Zhuohan Li, Siyuan Zhuang, Ying Sheng, Lianmin Zheng, Cody~Hao Yu, Joseph Gonzalez, Hao Zhang, and Ion Stoica.
\newblock Efficient memory management for large language model serving with pagedattention.
\newblock In {\em Proceedings of the 29th Symposium on Operating Systems Principles}, pages 611--626, 2023.

\bibitem{tensorrt-llm}
NVIDIA.
\newblock {TensorRT-LLM}.
\newblock \url{https://github.com/NVIDIA/TensorRT-LLM}.

\bibitem{zhang2023survey}
Hanqing Zhang, Haolin Song, Shaoyu Li, Ming Zhou, and Dawei Song.
\newblock A survey of controllable text generation using transformer-based pre-trained language models.
\newblock {\em ACM Computing Surveys}, 56(3):1--37, 2023.

\bibitem{potter2024hiddenpersuadersllmspolitical}
Yujin Potter, Shiyang Lai, Junsol Kim, James Evans, and Dawn Song.
\newblock Hidden persuaders: Llms' political leaning and their influence on voters, 2024.

\bibitem{cheong2024not}
Inyoung Cheong, King Xia, KJ~Kevin Feng, Quan~Ze Chen, and Amy~X Zhang.
\newblock (a) i am not a lawyer, but...: Engaging legal experts towards responsible llm policies for legal advice.
\newblock In {\em The 2024 ACM Conference on Fairness, Accountability, and Transparency}, pages 2454--2469, 2024.

\bibitem{wu2024fake}
Jiaying Wu, Jiafeng Guo, and Bryan Hooi.
\newblock Fake news in sheep's clothing: Robust fake news detection against llm-empowered style attacks.
\newblock In {\em Proceedings of the 30th ACM SIGKDD Conference on Knowledge Discovery and Data Mining}, pages 3367--3378, 2024.

\bibitem{treisman2023great}
Daniel Treisman.
\newblock How great is the current danger to democracy? assessing the risk with historical data.
\newblock {\em Comparative Political Studies}, 56(12):1924--1952, 2023.

\bibitem{meta2022human}
Meta Fundamental AI Research Diplomacy~Team (FAIR)†, Anton Bakhtin, Noam Brown, Emily Dinan, Gabriele Farina, Colin Flaherty, Daniel Fried, Andrew Goff, Jonathan Gray, Hengyuan Hu, et~al.
\newblock Human-level play in the game of diplomacy by combining language models with strategic reasoning.
\newblock {\em Science}, 378(6624):1067--1074, 2022.

\bibitem{colombo2024leveraging}
Andrea Colombo.
\newblock Leveraging knowledge graphs and llms to support and monitor legislative systems.
\newblock In {\em Proceedings of the 33rd ACM International Conference on Information and Knowledge Management}, pages 5443--5446, 2024.

\bibitem{egami2024using}
Naoki Egami, Musashi Hinck, Brandon Stewart, and Hanying Wei.
\newblock Using imperfect surrogates for downstream inference: Design-based supervised learning for social science applications of large language models.
\newblock {\em Advances in Neural Information Processing Systems}, 36, 2024.

\bibitem{liu-etal-2022-politics}
Yujian Liu, Xinliang~Frederick Zhang, David Wegsman, Nicholas Beauchamp, and Lu~Wang.
\newblock {POLITICS}: Pretraining with same-story article comparison for ideology prediction and stance detection.
\newblock In {\em Findings of the Association for Computational Linguistics: NAACL 2022}, pages 1354--1374, Seattle, United States, 2022. Association for Computational Linguistics.

\bibitem{chalkidis_llama_2024}
Ilias Chalkidis and Stephanie Brandl.
\newblock Llama meets {EU}: {Investigating} the {European} political spectrum through the lens of {LLMs}.
\newblock In {\em Proceedings of the 2024 {Conference} of the {North} {American} {Chapter} of the {Association} for {Computational} {Linguistics}: {Human} {Language} {Technologies} ({Volume} 2: {Short} {Papers})}, pages 481--498, 2024.

\bibitem{cao2024can}
Yupeng Cao, Aishwarya~Muralidharan Nair, Elyon Eyimife, Nastaran~Jamalipour Soofi, KP~Subbalakshmi, John~R Wullert~II, Chumki Basu, and David Shallcross.
\newblock Can large language models detect misinformation in scientific news reporting?
\newblock {\em arXiv preprint arXiv:2402.14268}, 2024.

\bibitem{gambini2024evaluating}
Margherita Gambini, Caterina Senette, Tiziano Fagni, and Maurizio Tesconi.
\newblock Evaluating large language models for user stance detection on x (twitter).
\newblock {\em Machine Learning}, pages 1--24, 2024.

\bibitem{wang_explainable_2024}
Bo~Wang, Jing Ma, Hongzhan Lin, Zhiwei Yang, Ruichao Yang, Yuan Tian, and Yi~Chang.
\newblock Explainable {Fake} {News} {Detection} with {Large} {Language} {Model} via {Defense} {Among} {Competing} {Wisdom}.
\newblock In {\em Proceedings of the {ACM} {Web} {Conference} 2024}, {WWW} '24, pages 2452--2463, New York, NY, USA, May 2024. Association for Computing Machinery.

\bibitem{wu_fake_2024}
Jiaying Wu, Jiafeng Guo, and Bryan Hooi.
\newblock Fake {News} in {Sheep}'s {Clothing}: {Robust} {Fake} {News} {Detection} {Against} {LLM}-{Empowered} {Style} {Attacks}.
\newblock In {\em Proceedings of the 30th {ACM} {SIGKDD} {Conference} on {Knowledge} {Discovery} and {Data} {Mining}}, pages 3367--3378, 2024.
\newblock arXiv:2310.10830 [cs].

\bibitem{hu2024bad}
Beizhe Hu, Qiang Sheng, Juan Cao, Yuhui Shi, Yang Li, Danding Wang, and Peng Qi.
\newblock Bad actor, good advisor: Exploring the role of large language models in fake news detection.
\newblock In {\em Proceedings of the AAAI Conference on Artificial Intelligence}, volume~38, pages 22105--22113, 2024.

\bibitem{whitehouse2022evaluation}
Chenxi Whitehouse, Tillman Weyde, Pranava Madhyastha, and Nikos Komninos.
\newblock Evaluation of fake news detection with knowledge-enhanced language models.
\newblock In {\em Proceedings of the international AAAI conference on web and social media}, volume~16, pages 1425--1429, 2022.

\bibitem{kocielnik2023can}
Rafal Kocielnik, Sara Kangaslahti, Shrimai Prabhumoye, Meena Hari, Michael Alvarez, and Anima Anandkumar.
\newblock Can you label less by using out-of-domain data? active \& transfer learning with few-shot instructions.
\newblock In {\em Transfer Learning for Natural Language Processing Workshop}, pages 22--32. PMLR, 2023.

\bibitem{lashitew2024corporate}
Addisu Lashitew and Youqing Mu.
\newblock Corporate opposition to climate change disclosure regulation in the united states.
\newblock {\em Climate Policy}, pages 1--16, 2024.

\bibitem{fu2024deciphering}
Xinyu Fu, Thomas~W Sanchez, Chaosu Li, and Juliana Reu~Junqueira.
\newblock Deciphering public voices in the digital era: Benchmarking chatgpt for analyzing citizen feedback in hamilton, new zealand.
\newblock {\em Journal of the American Planning Association}, pages 1--14, 2024.

\bibitem{napolio2024executive}
Nicholas~G Napolio.
\newblock Measuring executive agency ideology using large language models.
\newblock {\em Working Paper}, 2024.

\bibitem{bisbee2024synthetic}
James Bisbee, Joshua~D Clinton, Cassy Dorff, Brenton Kenkel, and Jennifer~M Larson.
\newblock Synthetic replacements for human survey data? the perils of large language models.
\newblock {\em Political Analysis}, pages 1--16, 2024.

\bibitem{alvarez2023generative}
R~Michael Alvarez, Frederick Eberhardt, and Mitchell Linegar.
\newblock Generative ai and the future of elections, 2023.

\bibitem{palmer2023large}
Alexis Palmer and Arthur Spirling.
\newblock Large language models can argue in convincing and novel ways about politics: Evidence from experiments and human judgement.
\newblock {\em Github Prepr}, 2023.

\bibitem{alvarez2024evaluating}
R~Michael Alvarez and Jacob Morrier.
\newblock Evaluating the quality of answers in political q\&a sessions with large language models.
\newblock {\em arXiv preprint arXiv:2404.08816}, 2024.

\bibitem{mellon2024ais}
Jonathan Mellon, Jack Bailey, Ralph Scott, James Breckwoldt, Marta Miori, and Phillip Schmedeman.
\newblock Do ais know what the most important issue is? using language models to code open-text social survey responses at scale.
\newblock {\em Research \& Politics}, 11(1):20531680241231468, 2024.

\bibitem{park2023generative}
Joon~Sung Park, Joseph O'Brien, Carrie~Jun Cai, Meredith~Ringel Morris, Percy Liang, and Michael~S Bernstein.
\newblock Generative agents: Interactive simulacra of human behavior.
\newblock In {\em Proceedings of the 36th annual acm symposium on user interface software and technology}, pages 1--22, 2023.

\bibitem{gao2023large}
Chen Gao, Xiaochong Lan, Nian Li, Yuan Yuan, Jingtao Ding, Zhilun Zhou, Fengli Xu, and Yong Li.
\newblock Large language models empowered agent-based modeling and simulation: A survey and perspectives.
\newblock {\em arXiv preprint arXiv:2312.11970}, 2023.

\bibitem{wang2024survey}
Lei Wang, Chen Ma, Xueyang Feng, Zeyu Zhang, Hao Yang, Jingsen Zhang, Zhiyuan Chen, Jiakai Tang, Xu~Chen, Yankai Lin, et~al.
\newblock A survey on large language model based autonomous agents.
\newblock {\em Frontiers of Computer Science}, 18(6):186345, 2024.

\bibitem{dai2024artificial}
Gordon Dai, Weijia Zhang, Jinhan Li, Siqi Yang, Srihas Rao, Arthur Caetano, Misha Sra, et~al.
\newblock Artificial leviathan: Exploring social evolution of llm agents through the lens of hobbesian social contract theory.
\newblock {\em arXiv preprint arXiv:2406.14373}, 2024.

\bibitem{hua2023war}
Wenyue Hua, Lizhou Fan, Lingyao Li, Kai Mei, Jianchao Ji, Yingqiang Ge, Libby Hemphill, and Yongfeng Zhang.
\newblock War and peace (waragent): Large language model-based multi-agent simulation of world wars.
\newblock {\em arXiv preprint arXiv:2311.17227}, 2023.

\bibitem{jin2024if}
Mingyu Jin, Beichen Wang, Zhaoqian Xue, Suiyuan Zhu, Wenyue Hua, Hua Tang, Kai Mei, Mengnan Du, and Yongfeng Zhang.
\newblock What if llms have different world views: Simulating alien civilizations with llm-based agents.
\newblock {\em arXiv preprint arXiv:2402.13184}, 2024.

\bibitem{chuang2023simulating}
Yun-Shiuan Chuang, Agam Goyal, Nikunj Harlalka, Siddharth Suresh, Robert Hawkins, Sijia Yang, Dhavan Shah, Junjie Hu, and Timothy~T Rogers.
\newblock Simulating opinion dynamics with networks of llm-based agents.
\newblock {\em arXiv preprint arXiv:2311.09618}, 2023.

\bibitem{guan2024richelieu}
Zhenyu Guan, Xiangyu Kong, Fangwei Zhong, and Yizhou Wang.
\newblock Richelieu: Self-evolving llm-based agents for ai diplomacy.
\newblock {\em arXiv preprint arXiv:2407.06813}, 2024.

\bibitem{moghimifar2024coalition}
Farhad Moghimifar, Yuan-Fang Li, Robert Thomson, and Gholamreza Haffari.
\newblock Modelling political coalition negotiations using llm-based agents.
\newblock {\em arXiv preprint arXiv:2402.11712}, 2024.

\bibitem{feder2022causal}
Amir Feder, Katherine~A Keith, Emaad Manzoor, Reid Pryzant, Dhanya Sridhar, Zach Wood-Doughty, Jacob Eisenstein, Justin Grimmer, Roi Reichart, Margaret~E Roberts, et~al.
\newblock Causal inference in natural language processing: Estimation, prediction, interpretation and beyond.
\newblock {\em Transactions of the Association for Computational Linguistics}, 10:1138--1158, 2022.

\bibitem{ashwani_cause_2024}
Swagata Ashwani, Kshiteesh Hegde, Nishith~Reddy Mannuru, Mayank Jindal, Dushyant~Singh Sengar, Krishna Chaitanya~Rao Kathala, Dishant Banga, Vinija Jain, and Aman Chadha.
\newblock Cause and {Effect}: {Can} {Large} {Language} {Models} {Truly} {Understand} {Causality}?, 2024.
\newblock arXiv:2402.18139 [cs].

\bibitem{zevcevic2023causal}
Matej Ze{\v{c}}evi{\'c}, Moritz Willig, Devendra~Singh Dhami, and Kristian Kersting.
\newblock Causal parrots: Large language models may talk causality but are not causal.
\newblock {\em arXiv preprint arXiv:2308.13067}, 2023.

\bibitem{kiciman_causal_2024}
Emre Kıcıman, Robert Ness, Amit Sharma, and Chenhao Tan.
\newblock Causal {Reasoning} and {Large} {Language} {Models}: {Opening} a {New} {Frontier} for {Causality}, 2024.
\newblock arXiv:2305.00050 [cs].

\bibitem{li_prompting_2024}
Yongqi Li, Mayi Xu, Xin Miao, Shen Zhou, and Tieyun Qian.
\newblock Prompting {Large} {Language} {Models} for {Counterfactual} {Generation}: {An} {Empirical} {Study}, 2024.
\newblock arXiv:2305.14791 [cs].

\bibitem{bhattacharjee_zero-shot_2024}
Amrita Bhattacharjee, Raha Moraffah, Joshua Garland, and Huan Liu.
\newblock Zero-shot {LLM}-guided {Counterfactual} {Generation} for {Text}, 2024.

\bibitem{gui_challenge_2023}
George Gui and Olivier Toubia.
\newblock The {Challenge} of {Using} {LLMs} to {Simulate} {Human} {Behavior}: {A} {Causal} {Inference} {Perspective}.
\newblock {\em SSRN Electronic Journal}, 2023.
\newblock arXiv:2312.15524 [cs].

\bibitem{wood-doughty_generating_2021}
Zach Wood-Doughty, Ilya Shpitser, and Mark Dredze.
\newblock Generating {Synthetic} {Text} {Data} to {Evaluate} {Causal} {Inference} {Methods}, February 2021.
\newblock arXiv:2102.05638.

\bibitem{zhao2024explainability}
Haiyan Zhao, Hanjie Chen, Fan Yang, Ninghao Liu, Huiqi Deng, Hengyi Cai, Shuaiqiang Wang, Dawei Yin, and Mengnan Du.
\newblock Explainability for large language models: A survey.
\newblock {\em ACM Transactions on Intelligent Systems and Technology}, 15(2):1--38, 2024.

\bibitem{luo2024understanding}
Haoyan Luo and Lucia Specia.
\newblock From understanding to utilization: A survey on explainability for large language models.
\newblock {\em arXiv preprint arXiv:2401.12874}, 2024.

\bibitem{de2024use}
Jef de~Slegte, Filip Van~Droogenbroeck, Bram Spruyt, Sam Verboven, and Vincent Ginis.
\newblock The use of machine learning methods in political science: An in-depth literature review.
\newblock {\em Political Studies Review}, page 14789299241265084, 2024.

\bibitem{dhawanend}
Nikita Dhawan, Leonardo Cotta, Karen Ullrich, Rahul Krishnan, and Chris~J Maddison.
\newblock End-to-end causal effect estimation from unstructured natural language data.
\newblock In {\em The Thirty-eighth Annual Conference on Neural Information Processing Systems}, 2024.

\bibitem{johnson2022ai}
Steven Johnson and Nikita Iziev.
\newblock Ai is mastering language. should we trust what it says?
\newblock {\em The New York Times}, 4:15, 2022.

\bibitem{kim2023rise}
Yunju Kim and Heejun Lee.
\newblock The rise of chatbots in political campaigns: The effects of conversational agents on voting intention.
\newblock {\em International Journal of Human--Computer Interaction}, 39(20):3984--3995, 2023.

\bibitem{lee2024large}
Messi~HJ Lee, Jacob~M Montgomery, and Calvin~K Lai.
\newblock Large language models portray socially subordinate groups as more homogeneous, consistent with a bias observed in humans.
\newblock In {\em The 2024 ACM Conference on Fairness, Accountability, and Transparency}, pages 1321--1340, 2024.

\bibitem{stanczak2023quantifying}
Karolina Sta{\'n}czak, Sagnik Ray~Choudhury, Tiago Pimentel, Ryan Cotterell, and Isabelle Augenstein.
\newblock Quantifying gender bias towards politicians in cross-lingual language models.
\newblock {\em Plos one}, 18(11):e0277640, 2023.

\bibitem{jiang2022communitylm}
Hang Jiang, Doug Beeferman, Brandon Roy, and Deb Roy.
\newblock Communitylm: Probing partisan worldviews from language models.
\newblock {\em arXiv preprint arXiv:2209.07065}, 2022.

\bibitem{simmons2023moral}
Gabriel Simmons.
\newblock Moral mimicry: Large language models produce moral rationalizations tailored to political identity.
\newblock In {\em Proceedings of the 61st Annual Meeting of the Association for Computational Linguistics (Volume 4: Student Research Workshop)}, pages 282--297, 2023.

\bibitem{hackenburg2024microtarget}
Kobi Hackenburg and Helen Margetts.
\newblock Evaluating the persuasive influence of political microtargeting with large language models.
\newblock {\em Proceedings of the National Academy of Sciences}, 121(24):e2403116121, 2024.

\bibitem{bonikowski2022politics}
Bart Bonikowski, Yuchen Luo, and Oscar Stuhler.
\newblock Politics as usual? measuring populism, nationalism, and authoritarianism in us presidential campaigns (1952--2020) with neural language models.
\newblock {\em Sociological Methods \& Research}, 51(4):1721--1787, 2022.

\bibitem{foos2024use}
Florian Foos.
\newblock The use of ai by election campaigns.
\newblock {\em OSF}, 2024.

\bibitem{yu2024will}
Chenxiao Yu, Zhaotian Weng, Zheng Li, Xiyang Hu, and Yue Zhao.
\newblock Will trump win in 2024? predicting the us presidential election via multi-step reasoning with large language models.
\newblock {\em arXiv preprint arXiv:2411.03321}, 2024.

\bibitem{argyle2023leveraging}
Lisa~P Argyle, Christopher~A Bail, Ethan~C Busby, Joshua~R Gubler, Thomas Howe, Christopher Rytting, Taylor Sorensen, and David Wingate.
\newblock Leveraging ai for democratic discourse: Chat interventions can improve online political conversations at scale.
\newblock {\em Proceedings of the National Academy of Sciences}, 120(41):e2311627120, 2023.

\bibitem{ma2024chatgpt}
ZIlin Ma, Yiyang Mei, Claude Bruderlein, Krzysztof~Z Gajos, and Weiwei Pan.
\newblock " chatgpt, don't tell me what to do": Designing ai for context analysis in humanitarian frontline negotiations.
\newblock {\em arXiv preprint arXiv:2410.09139}, 2024.

\bibitem{bastan-etal-2020-authors}
Mohaddeseh Bastan, Mahnaz Koupaee, Youngseo Son, Richard Sicoli, and Niranjan Balasubramanian.
\newblock Author{'}s sentiment prediction.
\newblock In {\em Proceedings of the 28th International Conference on Computational Linguistics}, pages 604--615, 2020.

\bibitem{chebolu2022survey}
Siva Uday~Sampreeth Chebolu, Franck Dernoncourt, Nedim Lipka, and Thamar Solorio.
\newblock Survey of aspect-based sentiment analysis datasets.
\newblock {\em arXiv preprint arXiv:2204.05232}, 2022.

\bibitem{sharma2022fake}
Srishti Sharma, Mala Saraswat, and Anil~Kumar Dubey.
\newblock Fake news detection on twitter.
\newblock {\em International Journal of Web Information Systems}, 18(5/6):388--412, 2022.

\bibitem{saha2022sentiment}
Uchchhwas Saha, Md~Shihab Mahmud, Aisharjo Chakrobortty, Mst~Tuhin Akter, MD~Rakib Islam, and Ahmed Al~Marouf.
\newblock Sentiment classification in bengali news comments using a hybrid approach with glove.
\newblock In {\em 2022 6th International Conference on Trends in Electronics and Informatics (ICOEI)}, pages 01--08, 2022.

\bibitem{waspodo2022indonesia}
Bayu Waspodo, Amalia Khaerunnisa~Nursya Bany, Rinda~Hesti Kusumaningtyas, Eri Rustamaji, et~al.
\newblock Indonesia covid-19 online media news sentiment analysis with lexicon-based approach and emotion detection.
\newblock In {\em 2022 10th International Conference on Cyber and IT Service Management (CITSM)}, pages 1--6, 2022.

\bibitem{DVN/ER9XTV_2022}
MIT~Election Data and Science Lab.
\newblock {U.S. Senate Precinct-Level Returns 2020}.
\newblock {\em Harvard Dataverse}, 2022.

\bibitem{DVN/IVIXLK_2022}
MIT~Election Data and Science Lab.
\newblock {U.S. House of Representatives Precinct-Level Returns 2018}.
\newblock {\em Harvard Dataverse}, 2022.

\bibitem{DVN/ZFXEJU_2022}
MIT~Election Data and Science Lab.
\newblock {State Precinct-Level Returns 2018}.
\newblock {\em Harvard Dataverse}, 2022.

\bibitem{payne2010implicit}
B~Keith Payne, Jon~A Krosnick, Josh Pasek, Yphtach Lelkes, Omair Akhtar, and Trevor Tompson.
\newblock Implicit and explicit prejudice in the 2008 american presidential election.
\newblock {\em Journal of Experimental Social Psychology}, 46(2):367--374, 2010.

\bibitem{yu2024trumpwin2024predicting}
Chenxiao Yu, Zhaotian Weng, Zheng Li, Xiyang Hu, and Yue Zhao.
\newblock Will trump win in 2024? predicting the us presidential election via multi-step reasoning with large language models, 2024.

\bibitem{DVN/42MVDX_2017}
MIT~Election Data and Science Lab.
\newblock {U.S. President 1976–2020}.
\newblock {\em Harvard Dataverse}, 2017.

\bibitem{kornilova2019billsum}
Anastassia Kornilova and Vlad Eidelman.
\newblock Billsum: A corpus for automatic summarization of us legislation.
\newblock {\em arXiv preprint arXiv:1910.00523}, 2019.

\bibitem{shu2024lawllm}
Dong Shu, Haoran Zhao, Xukun Liu, David Demeter, Mengnan Du, and Yongfeng Zhang.
\newblock Lawllm: Law large language model for the us legal system.
\newblock {\em arXiv preprint arXiv:2407.21065}, 2024.

\bibitem{arregui2022new}
Javier Arregui and Clement Perarnaud.
\newblock A new dataset on legislative decision-making in the european union: the deu iii dataset.
\newblock {\em Journal of European Public Policy}, 29(1):12--22, 2022.

\bibitem{shu2020fakenewsnet}
Kai Shu, Deepak Mahudeswaran, Suhang Wang, Dongwon Lee, and Huan Liu.
\newblock Fakenewsnet: A data repository with news content, social context, and spatiotemporal information for studying fake news on social media.
\newblock {\em Big data}, 8(3):171--188, 2020.

\bibitem{grover2022public}
Karish Grover, SM~Angara, Md~Shad Akhtar, and Tanmoy Chakraborty.
\newblock Public wisdom matters! discourse-aware hyperbolic fourier co-attention for social text classification.
\newblock {\em Advances in Neural Information Processing Systems}, 35:9417--9431, 2022.

\bibitem{jin2017multimodal}
Zhiwei Jin, Juan Cao, Han Guo, Yongdong Zhang, and Jiebo Luo.
\newblock Multimodal fusion with recurrent neural networks for rumor detection on microblogs.
\newblock In {\em Proceedings of the 25th ACM international conference on Multimedia}, pages 795--816, 2017.

\bibitem{cunningham2013non}
David~E Cunningham, Kristian~Skrede Gleditsch, and Idean Salehyan.
\newblock Non-state actors in civil wars: A new dataset.
\newblock {\em Conflict management and peace science}, 30(5):516--531, 2013.

\bibitem{ari2023peace}
Bar{\i}{\c{s}} Ar{\i}.
\newblock Peace negotiations in civil conflicts: A new dataset.
\newblock {\em Journal of Conflict Resolution}, 67(1):150--177, 2023.

\bibitem{mochtak2023parlasent}
Michal Mochtak, Peter Rupnik, and Nikola Ljube{\v{s}}i{\'c}.
\newblock The parlasent multilingual training dataset for sentiment identification in parliamentary proceedings.
\newblock {\em arXiv preprint arXiv:2309.09783}, 2023.

\bibitem{balloccu2024leak}
Simone Balloccu, Patr{\'\i}cia Schmidtov{\'a}, Mateusz Lango, and Ond{\v{r}}ej Du{\v{s}}ek.
\newblock Leak, cheat, repeat: Data contamination and evaluation malpractices in closed-source llms.
\newblock {\em arXiv preprint arXiv:2402.03927}, 2024.

\bibitem{rauniyar2023multi}
Kritesh Rauniyar, Sweta Poudel, Shuvam Shiwakoti, Surendrabikram Thapa, Junaid Rashid, Jungeun Kim, Muhammad Imran, and Usman Naseem.
\newblock Multi-aspect annotation and analysis of nepali tweets on anti-establishment election discourse.
\newblock {\em IEEE Access}, 2023.

\bibitem{tan2024large}
Zhen Tan, Dawei Li, Song Wang, Alimohammad Beigi, Bohan Jiang, Amrita Bhattacharjee, Mansooreh Karami, Jundong Li, Lu~Cheng, and Huan Liu.
\newblock Large language models for data annotation and synthesis: A survey.
\newblock In {\em Proceedings of the 2024 Conference on Empirical Methods in Natural Language Processing}, pages 930--957, 2024.

\bibitem{ming2024autolabel}
Xuran Ming, Shoubin Li, Mingyang Li, Lvlong He, and Qing Wang.
\newblock Autolabel: Automated textual data annotation method based on active learning and large language model.
\newblock In {\em International Conference on Knowledge Science, Engineering and Management}, pages 400--411, 2024.

\bibitem{qu2024performance}
Yao Qu and Jue Wang.
\newblock Performance and biases of large language models in public opinion simulation.
\newblock {\em Humanities and Social Sciences Communications}, 11(1):1--13, 2024.

\bibitem{shahbazi2023representation}
Nima Shahbazi, Yin Lin, Abolfazl Asudeh, and HV~Jagadish.
\newblock Representation bias in data: A survey on identification and resolution techniques.
\newblock {\em ACM Computing Surveys}, 55(13s):1--39, 2023.

\bibitem{nakada2024synthetic}
Ryumei Nakada, Yichen Xu, Lexin Li, and Linjun Zhang.
\newblock Synthetic oversampling: Theory and a practical approach using llms to address data imbalance.
\newblock {\em arXiv preprint arXiv:2406.03628}, 2024.

\bibitem{cloutier2023fine}
Nicolas~Antonio Cloutier and Nathalie Japkowicz.
\newblock Fine-tuned generative llm oversampling can improve performance over traditional techniques on multiclass imbalanced text classification.
\newblock In {\em 2023 IEEE International Conference on Big Data (BigData)}, pages 5181--5186, 2023.

\bibitem{sahu2023promptmix}
Gaurav Sahu, Olga Vechtomova, Dzmitry Bahdanau, and Issam~H Laradji.
\newblock Promptmix: A class boundary augmentation method for large language model distillation.
\newblock {\em arXiv preprint arXiv:2310.14192}, 2023.

\bibitem{abaskohi2023lm}
Amirhossein Abaskohi, Sascha Rothe, and Yadollah Yaghoobzadeh.
\newblock Lm-cppf: Paraphrasing-guided data augmentation for contrastive prompt-based few-shot fine-tuning.
\newblock In {\em Proceedings of the 61st Annual Meeting of the Association for Computational Linguistics (Volume 2: Short Papers)}, pages 670--681, 2023.

\bibitem{dos2024identifying}
Vitor~Gaboardi dos Santos, Guto~Leoni Santos, Theo Lynn, and Boualem Benatallah.
\newblock Identifying citizen-related issues from social media using llm-based data augmentation.
\newblock In {\em International Conference on Advanced Information Systems Engineering}, pages 531--546. Springer, 2024.

\bibitem{ding2024data}
Bosheng Ding, Chengwei Qin, Ruochen Zhao, Tianze Luo, Xinze Li, Guizhen Chen, Wenhan Xia, Junjie Hu, Anh~Tuan Luu, and Shafiq Joty.
\newblock Data augmentation using llms: Data perspectives, learning paradigms and challenges.
\newblock {\em arXiv preprint arXiv:2403.02990}, 2024.

\bibitem{zhang2024scaling}
Biao Zhang, Zhongtao Liu, Colin Cherry, and Orhan Firat.
\newblock When scaling meets llm finetuning: The effect of data, model and finetuning method.
\newblock {\em arXiv preprint arXiv:2402.17193}, 2024.

\bibitem{vm2024fine}
Kushala VM, Harikrishna Warrier, Yogesh Gupta, et~al.
\newblock Fine tuning llm for enterprise: Practical guidelines and recommendations.
\newblock {\em arXiv preprint arXiv:2404.10779}, 2024.

\bibitem{jaradat2024multitask}
Shadi Jaradat, Richi Nayak, Alexander Paz, Huthaifa~I Ashqar, and Mohammad Elhenawy.
\newblock Multitask learning for crash analysis: A fine-tuned llm framework using twitter data.
\newblock {\em Smart Cities}, 7(5):2422--2465, 2024.

\bibitem{wu2024dlora}
Bingyang Wu, Ruidong Zhu, Zili Zhang, Peng Sun, Xuanzhe Liu, and Xin Jin.
\newblock $\{$dLoRA$\}$: Dynamically orchestrating requests and adapters for $\{$LoRA$\}$$\{$LLM$\}$ serving.
\newblock In {\em 18th USENIX Symposium on Operating Systems Design and Implementation (OSDI 24)}, pages 911--927, 2024.

\bibitem{meloux2024novel}
Maxime M{\'e}loux and Christophe Cerisara.
\newblock Novel-wd: Exploring acquisition of novel world knowledge in llms using prefix-tuning.
\newblock {\em arXiv preprint arXiv:2408.17070}, 2024.

\bibitem{nabli2024acco}
Adel Nabli, Louis Fournier, Pierre Erbacher, Louis Serrano, Eugene Belilovsky, and Edouard Oyallon.
\newblock Acco: Accumulate while you communicate, hiding communications in distributed llm training.
\newblock {\em arXiv preprint arXiv:2406.02613}, 2024.

\bibitem{guan2024aptq}
Ziyi Guan, Hantao Huang, Yupeng Su, Hong Huang, Ngai Wong, and Hao Yu.
\newblock Aptq: Attention-aware post-training mixed-precision quantization for large language models.
\newblock In {\em Proceedings of the 61st ACM/IEEE Design Automation Conference}, pages 1--6, 2024.

\bibitem{banerjee2023benchmarking}
Debarag Banerjee, Pooja Singh, Arjun Avadhanam, and Saksham Srivastava.
\newblock Benchmarking llm powered chatbots: methods and metrics.
\newblock {\em arXiv preprint arXiv:2308.04624}, 2023.

\bibitem{petrova2020extracting}
Alina Petrova, John Armour, and Thomas Lukasiewicz.
\newblock Extracting outcomes from appellate decisions in us state courts.
\newblock In {\em Legal Knowledge and Information Systems}, pages 133--142. IOS Press, 2020.

\bibitem{santosh2024lexsumm}
TYSS Santosh, Cornelius Weiss, and Matthias Grabmair.
\newblock Lexsumm and lext5: Benchmarking and modeling legal summarization tasks in english.
\newblock {\em arXiv preprint arXiv:2410.09527}, 2024.

\bibitem{guha2024legalbench}
Neel Guha, Julian Nyarko, Daniel Ho, Christopher R{\'e}, Adam Chilton, Alex Chohlas-Wood, Austin Peters, Brandon Waldon, Daniel Rockmore, Diego Zambrano, et~al.
\newblock Legalbench: A collaboratively built benchmark for measuring legal reasoning in large language models.
\newblock {\em Advances in Neural Information Processing Systems}, 36, 2024.

\bibitem{wei2021finetuned}
Jason Wei, Maarten Bosma, Vincent~Y Zhao, Kelvin Guu, Adams~Wei Yu, Brian Lester, Nan Du, Andrew~M Dai, and Quoc~V Le.
\newblock Finetuned language models are zero-shot learners.
\newblock {\em arXiv preprint arXiv:2109.01652}, 2021.

\bibitem{kumar2023zero}
Puneet Kumar, Kshitij Pathania, and Balasubramanian Raman.
\newblock Zero-shot learning based cross-lingual sentiment analysis for sanskrit text with insufficient labeled data.
\newblock {\em Applied Intelligence}, 53(9):10096--10113, 2023.

\bibitem{di2024mapping}
Riccardo Di~Leo, Chen Zeng, Elias Dinas, and Reda Tamtam.
\newblock Mapping (a) ideology: A taxonomy of european parties using generative llms as zero-shot learners.
\newblock {\em Available at SSRN 4907347}, 2024.

\bibitem{allaway2023zero}
Emily Allaway and Kathleen McKeown.
\newblock Zero-shot stance detection: Paradigms and challenges.
\newblock {\em Frontiers in Artificial Intelligence}, 5:1070429, 2023.

\bibitem{malladynamic}
Ranadheer Malla, Travis~G Coan, Vivek Srinivasan, and Constantine Boussalis.
\newblock Dynamic few-shot learning for computational social science.
\newblock {\em OSF}, 2024.

\bibitem{kuila2024deciphering}
Alapan Kuila and Sudeshna Sarkar.
\newblock Deciphering political entity sentiment in news with large language models: Zero-shot and few-shot strategies.
\newblock {\em arXiv preprint arXiv:2404.04361}, 2024.

\bibitem{hu2023synthesizing}
Yibo Hu, Erick~Skorupa Parolin, Latifur Khan, Patrick~T Brandt, Javier Osorio, and Vito~J D'Orazio.
\newblock Synthesizing political zero-shot relation classification via codebook knowledge, nli, and chatgpt.
\newblock {\em arXiv preprint arXiv:2308.07876}, 2023.

\bibitem{burnham2024political}
Michael Burnham, Kayla Kahn, Ryan~Yank Wang, and Rachel~X Peng.
\newblock Political debate: Efficient zero-shot and few-shot classifiers for political text.
\newblock {\em arXiv preprint arXiv:2409.02078}, 2024.

\bibitem{wahidur2024enhancing}
Rahman~SM Wahidur, Ishmam Tashdeed, Manjit Kaur, and Heung-No Lee.
\newblock Enhancing zero-shot crypto sentiment with fine-tuned language model and prompt engineering.
\newblock {\em IEEE Access}, 2024.

\bibitem{yao2024more}
Bingsheng Yao, Guiming Chen, Ruishi Zou, Yuxuan Lu, Jiachen Li, Shao Zhang, Yisi Sang, Sijia Liu, James Hendler, and Dakuo Wang.
\newblock More samples or more prompts? exploring effective few-shot in-context learning for llms with in-context sampling.
\newblock In {\em Findings of the Association for Computational Linguistics: NAACL 2024}, pages 1772--1790, 2024.

\bibitem{ibrahim2024analyzing}
Hazem Ibrahim, Farhan Khan, Hend Alabdouli, Maryam Almatrooshi, Tran Nguyen, Talal Rahwan, and Yasir Zaki.
\newblock Analyzing political stances on twitter in the lead-up to the 2024 us election.
\newblock {\em arXiv preprint arXiv:2412.02712}, 2024.

\bibitem{kuntur2024under}
Soveatin Kuntur, Anna Wr{\'o}blewska, Marcin Paprzycki, and Maria Ganzha.
\newblock Under the influence: A survey of large language models in fake news detection.
\newblock {\em IEEE Transactions on Artificial Intelligence}, 2024.

\bibitem{pavlyshenko2024using}
Bohdan~M Pavlyshenko.
\newblock Using gpt models for qualitative and quantitative news analytics in the 2024 us presidental election process.
\newblock {\em arXiv preprint arXiv:2410.15884}, 2024.

\bibitem{hu2024multi}
Weiqi Hu, Ye~Wang, Yan Jia, Qing Liao, and Bin Zhou.
\newblock A multi-modal prompt learning framework for early detection of fake news.
\newblock In {\em Proceedings of the International AAAI Conference on Web and Social Media}, volume~18, pages 651--662, 2024.

\bibitem{chen2024fine}
Xiaojun Chen, Ting Liu, Philippe Fournier-Viger, Bowen Zhang, Guodong Long, and Qin Zhang.
\newblock A fine-grained self-adapting prompt learning approach for few-shot learning with pre-trained language models.
\newblock {\em Knowledge-Based Systems}, page 111968, 2024.

\bibitem{wang2024evaluating}
Yang Wang, Alberto~Garcia Hernandez, Roman Kyslyi, and Nicholas Kersting.
\newblock Evaluating quality of answers for retrieval-augmented generation: A strong llm is all you need.
\newblock {\em arXiv preprint arXiv:2406.18064}, 2024.

\bibitem{dong2024understand}
Guanting Dong, Yutao Zhu, Chenghao Zhang, Zechen Wang, Zhicheng Dou, and Ji-Rong Wen.
\newblock Understand what llm needs: Dual preference alignment for retrieval-augmented generation.
\newblock {\em arXiv preprint arXiv:2406.18676}, 2024.

\bibitem{arslan2024political}
Muhammad Arslan, Saba Munawar, and Christophe Cruz.
\newblock Political-rag: using generative ai to extract political information from media content.
\newblock {\em Journal of Information Technology \& Politics}, pages 1--16, 2024.

\bibitem{zhanggenerating}
Zhen-Yu Zhang, Siwei Han, Huaxiu Yao, Gang Niu, and Masashi Sugiyama.
\newblock Generating chain-of-thoughts with a pairwise-comparison approach to searching for the most promising intermediate thought.
\newblock In {\em Forty-first International Conference on Machine Learning}, 2024.

\bibitem{kareem2023fighting}
Waleed Kareem and Noorhan Abbas.
\newblock Fighting lies with intelligence: Using large language models and chain of thoughts technique to combat fake news.
\newblock In {\em International Conference on Innovative Techniques and Applications of Artificial Intelligence}, pages 253--258, 2023.

\bibitem{dobrinoiu2024leveraging}
Adina Dobrinoiu.
\newblock {\em Leveraging Large Language Models for Classifying Subjective Arguments in Public Discourse}.
\newblock PhD thesis, Delft University of Technology, 2024.

\bibitem{tutunov2023can}
Rasul Tutunov, Antoine Grosnit, Juliusz Ziomek, Jun Wang, and Haitham Bou-Ammar.
\newblock Why can large language models generate correct chain-of-thoughts?
\newblock {\em arXiv preprint arXiv:2310.13571}, 2023.

\bibitem{zhang2024comprehensive}
Ningyu Zhang, Yunzhi Yao, Bozhong Tian, Peng Wang, Shumin Deng, Mengru Wang, Zekun Xi, Shengyu Mao, Jintian Zhang, Yuansheng Ni, et~al.
\newblock A comprehensive study of knowledge editing for large language models.
\newblock {\em arXiv preprint arXiv:2401.01286}, 2024.

\bibitem{gupta2024stackfeed}
Naman Gupta, Shashank Kirtania, Priyanshu Gupta, and Others.
\newblock Stackfeed: Structured textual actor-critic knowledge base editing with feedback.
\newblock {\em arXiv preprint arXiv:2410.10584}, 2024.

\bibitem{zhang2024oneedit}
Ningyu Zhang, Zekun Xi, Yujie Luo, Peng Wang, Bozhong Tian, Yunzhi Yao, Jintian Zhang, Shumin Deng, Mengshu Sun, Lei Liang, et~al.
\newblock Oneedit: A neural-symbolic collaboratively knowledge editing system.
\newblock {\em arXiv preprint arXiv:2409.07497}, 2024.

\bibitem{peng2024event}
Hao Peng, Xiaozhi Wang, Chunyang Li, Kaisheng Zeng, Jiangshan Duo, Yixin Cao, Lei Hou, and Juanzi Li.
\newblock Event-level knowledge editing.
\newblock {\em arXiv preprint arXiv:2402.13093}, 2024.

\bibitem{ahmed2023better}
Toufique Ahmed and Premkumar Devanbu.
\newblock Better patching using llm prompting, via self-consistency.
\newblock In {\em 2023 38th IEEE/ACM International Conference on Automated Software Engineering (ASE)}, pages 1742--1746, 2023.

\bibitem{huang2023enhancing}
Baizhou Huang, Shuai Lu, Weizhu Chen, Xiaojun Wan, and Nan Duan.
\newblock Enhancing large language models in coding through multi-perspective self-consistency.
\newblock {\em arXiv preprint arXiv:2309.17272}, 2023.

\bibitem{cheng2024integrative}
Yi~Cheng, Xiao Liang, Yeyun Gong, Wen Xiao, Song Wang, Yuji Zhang, Wenjun Hou, Kaishuai Xu, Wenge Liu, Wenjie Li, et~al.
\newblock Integrative decoding: Improve factuality via implicit self-consistency.
\newblock {\em arXiv preprint arXiv:2410.01556}, 2024.

\bibitem{chen2023two}
Angelica Chen, Jason Phang, Alicia Parrish, Vishakh Padmakumar, Chen Zhao, Samuel~R Bowman, and Kyunghyun Cho.
\newblock Two failures of self-consistency in the multi-step reasoning of llms.
\newblock {\em arXiv preprint arXiv:2305.14279}, 2023.

\bibitem{islam2024gpt}
Raisa Islam and Owana~Marzia Moushi.
\newblock Gpt-4o: The cutting-edge advancement in multimodal llm.
\newblock {\em Authorea Preprints}, 2024.

\bibitem{rasheed2024taskcomplexity}
Areeg~Fahad Rasheed, M~Zarkoosh, Safa~F Abbas, and Sana~Sabah Al-Azzawi.
\newblock Taskcomplexity: A dataset for task complexity classification with in-context learning, flan-t5 and gpt-4o benchmarks.
\newblock {\em arXiv preprint arXiv:2409.20189}, 2024.

\bibitem{chen2024magicdec}
Jian Chen, Vashisth Tiwari, Ranajoy Sadhukhan, Zhuoming Chen, Jinyuan Shi, Ian En-Hsu Yen, and Beidi Chen.
\newblock Magicdec: Breaking the latency-throughput tradeoff for long context generation with speculative decoding.
\newblock {\em arXiv preprint arXiv:2408.11049}, 2024.

\bibitem{singh2024scidqa}
Shruti Singh, Nandan Sarkar, and Arman Cohan.
\newblock Scidqa: A deep reading comprehension dataset over scientific papers.
\newblock {\em arXiv preprint arXiv:2411.05338}, 2024.

\bibitem{ANES2016}
American National~Election Studies.
\newblock Anes 2016 time series study full release.
\newblock \url{https://www.electionstudies.org}, 2019.
\newblock [Dataset and documentation]. September 4, 2019 version.

\bibitem{kennedy2018evaluation}
Courtney Kennedy, Mark Blumenthal, Scott Clement, et~al.
\newblock An evaluation of the 2016 election polls in the united states.
\newblock {\em Public Opinion Quarterly}, 82(1):1--33, 2018.

\bibitem{Argyle_Busby_Fulda_Gubler_Rytting_Wingate_2023}
Lisa~P. Argyle, Ethan~C. Busby, Nancy Fulda, Joshua~R. Gubler, Christopher Rytting, and David Wingate.
\newblock Out of one, many: Using language models to simulate human samples.
\newblock {\em Political Analysis}, 31(3):337–351, 2023.

\bibitem{qi2024representation}
Weihong Qi, Hanjia Lyu, and Jiebo Luo.
\newblock Representation bias in political sample simulations with large language models.
\newblock {\em arXiv preprint arXiv:2407.11409}, 2024.

\bibitem{majumdar2024generative}
Srijoni Majumdar, Edith Elkind, and Evangelos Pournaras.
\newblock Generative ai voting: Fair collective choice is resilient to llm biases and inconsistencies.
\newblock {\em arXiv preprint arXiv:2406.11871}, 2024.

\bibitem{wagner2024power}
Stefan~Sylvius Wagner, Maike Behrendt, Marc Ziegele, and Stefan Harmeling.
\newblock The power of llm-generated synthetic data for stance detection in online political discussions.
\newblock {\em arXiv preprint arXiv:2406.12480}, 2024.

\bibitem{von2023assessing}
Leah von~der Heyde, Anna-Carolina Haensch, and Alexander Wenz.
\newblock Assessing bias in llm-generated synthetic datasets: The case of german voter behavior.
\newblock Technical report, Center for Open Science, 2023.

\bibitem{huang2024social}
Yue Huang, Zhengqing Yuan, Yujun Zhou, Kehan Guo, Xiangqi Wang, Haomin Zhuang, Weixiang Sun, Lichao Sun, Jindong Wang, Yanfang Ye, et~al.
\newblock Social science meets llms: How reliable are large language models in social simulations?
\newblock {\em arXiv preprint arXiv:2410.23426}, 2024.

\bibitem{ling2024deductive}
Zhan Ling, Yunhao Fang, Xuanlin Li, Zhiao Huang, Mingu Lee, Roland Memisevic, and Hao Su.
\newblock Deductive verification of chain-of-thought reasoning.
\newblock {\em Advances in Neural Information Processing Systems}, 36, 2024.

\bibitem{chen2024can}
Canyu Chen, Baixiang Huang, Zekun Li, Zhaorun Chen, Shiyang Lai, Xiongxiao Xu, Jia-Chen Gu, Jindong Gu, Huaxiu Yao, Chaowei Xiao, et~al.
\newblock Can editing llms inject harm?
\newblock {\em arXiv preprint arXiv:2407.20224}, 2024.

\bibitem{khurana2023natural}
Diksha Khurana, Aditya Koli, Kiran Khatter, and Sukhdev Singh.
\newblock Natural language processing: state of the art, current trends and challenges.
\newblock {\em Multimedia tools and applications}, 82(3):3713--3744, 2023.

\bibitem{bang2024measuring}
Yejin Bang, Delong Chen, Nayeon Lee, and Pascale Fung.
\newblock Measuring political bias in large language models: What is said and how it is said.
\newblock {\em arXiv preprint arXiv:2403.18932}, 2024.

\bibitem{wang2024twin}
Yue Wang, Yinlong Xu, Zihan Ma, Hongxia Xu, Bang Du, Honghao Gao, Jian Wu, and Jintai Chen.
\newblock Twin-gpt: Digital twins for clinical trials via large language model.
\newblock {\em ACM Transactions on Multimedia Computing, Communications and Applications}, 2024.

\bibitem{de2014agent}
Scott De~Marchi and Scott~E Page.
\newblock Agent-based models.
\newblock {\em Annual Review of political science}, 17(1):1--20, 2014.

\bibitem{de2005computational}
Scott De~Marchi.
\newblock {\em Computational and mathematical modeling in the social sciences}.
\newblock Cambridge University Press, 2005.

\bibitem{yao2024comal}
Huaiyuan Yao, Longchao Da, Vishnu Nandam, Justin Turnau, Zhiwei Liu, Linsey Pang, and Hua Wei.
\newblock Comal: Collaborative multi-agent large language models for mixed-autonomy traffic.
\newblock {\em arXiv preprint arXiv:2410.14368}, 2024.

\bibitem{bagheri_c2p_2024}
Abdolmahdi Bagheri, Matin Alinejad, Kevin Bello, and Alireza Akhondi-Asl.
\newblock {C2P}: {Featuring} {Large} {Language} {Models} with {Causal} {Reasoning}, 2024.
\newblock arXiv:2407.18069 [cs].

\bibitem{jiralerspong_efficient_2024}
Thomas Jiralerspong, Xiaoyin Chen, Yash More, Vedant Shah, and Yoshua Bengio.
\newblock Efficient {Causal} {Graph} {Discovery} {Using} {Large} {Language} {Models}, 2024.
\newblock arXiv:2402.01207 [cs].

\bibitem{batzdorfer2024conspiracy}
Veronika Batzdorfer.
\newblock Conspiracy narratives on voat: A longitudinal analysis of cognitive activation and evolutionary psychology features.
\newblock In {\em Proceedings of the 16th ACM Web Science Conference}, pages 42--47, 2024.

\bibitem{vashishtha2023causal}
Aniket Vashishtha, Abbavaram~Gowtham Reddy, Abhinav Kumar, Saketh Bachu, Vineeth~N Balasubramanian, and Amit Sharma.
\newblock Causal inference using llm-guided discovery.
\newblock {\em arXiv preprint arXiv:2310.15117}, 2023.

\bibitem{ashwani2024cause}
Swagata Ashwani, Kshiteesh Hegde, Nishith~Reddy Mannuru, Dushyant~Singh Sengar, Mayank Jindal, Krishna Chaitanya~Rao Kathala, Dishant Banga, Vinija Jain, and Aman Chadha.
\newblock Cause and effect: Can large language models truly understand causality?
\newblock In {\em Proceedings of the AAAI Symposium Series}, volume~4, pages 2--9, 2024.

\bibitem{bai2023artificial}
Hui Bai, Jan Voelkel, Johannes Eichstaedt, and Robb Willer.
\newblock Artificial intelligence can persuade humans on political issues.
\newblock {\em Osf}, 2023.

\bibitem{DVN/PEJ5QU_2017}
MIT~Election Data and Science Lab.
\newblock {U.S. Senate statewide 1976–2020}.
\newblock {\em Harvard Dataverse}, 2017.

\bibitem{DVN/IG0UN2_2017}
MIT~Election Data and Science Lab.
\newblock {U.S. House 1976–2022}.
\newblock {\em Harvard Dataverse}, 2017.

\bibitem{DVN/ZH7J2G_2018}
Emily Moore.
\newblock {Federal Register Final Rule Data 2000-2014}.
\newblock {\em Harvard Dataverse}, 2018.

\bibitem{yu2024makes}
Xiao Yu, Zexian Zhang, Feifei Niu, Xing Hu, Xin Xia, and John Grundy.
\newblock What makes a high-quality training dataset for large language models: A practitioners' perspective.
\newblock In {\em Proceedings of the 39th IEEE/ACM International Conference on Automated Software Engineering}, pages 656--668, 2024.

\bibitem{lin2024designing}
Haocheng Lin.
\newblock Designing domain-specific large language models: The critical role of fine-tuning in public opinion simulation.
\newblock {\em arXiv preprint arXiv:2409.19308}, 2024.

\bibitem{huang2024selective}
Chen Huang, Yang Deng, Wenqiang Lei, Jiancheng Lv, and Ido Dagan.
\newblock Selective annotation via data allocation: These data should be triaged to experts for annotation rather than the model.
\newblock {\em arXiv preprint arXiv:2405.12081}, 2024.

\bibitem{chai2023comparison}
Christine~P Chai.
\newblock Comparison of text preprocessing methods.
\newblock {\em Natural Language Engineering}, 29(3):509--553, 2023.

\bibitem{ehrmann2023named}
Maud Ehrmann, Ahmed Hamdi, Elvys~Linhares Pontes, Matteo Romanello, and Antoine Doucet.
\newblock Named entity recognition and classification in historical documents: A survey.
\newblock {\em ACM Computing Surveys}, 56(2):1--47, 2023.

\bibitem{wicke2024red}
Philipp Wicke and Marianna~M Bolognesi.
\newblock Red and blue language: Word choices in the trump \& harris 2024 presidential debate.
\newblock {\em arXiv preprint arXiv:2410.13654}, 2024.

\bibitem{molina2021fake}
Maria~D Molina, S~Shyam Sundar, Thai Le, and Dongwon Lee.
\newblock “fake news” is not simply false information: A concept explication and taxonomy of online content.
\newblock {\em American behavioral scientist}, 65(2):180--212, 2021.

\bibitem{khaliq2024ragar}
M~Abdul Khaliq, P~Chang, M~Ma, Bernhard Pflugfelder, and F~Mileti{\'c}.
\newblock Ragar, your falsehood radar: Rag-augmented reasoning for political fact-checking using multimodal large language models.
\newblock {\em arXiv preprint arXiv:2404.12065}, 2024.

\bibitem{antwarg2021explaining}
Liat Antwarg, Ronnie~Mindlin Miller, Bracha Shapira, and Lior Rokach.
\newblock Explaining anomalies detected by autoencoders using shapley additive explanations.
\newblock {\em Expert systems with applications}, 186:115736, 2021.

\bibitem{lundstrom2022rigorous}
Daniel~D Lundstrom, Tianjian Huang, and Meisam Razaviyayn.
\newblock A rigorous study of integrated gradients method and extensions to internal neuron attributions.
\newblock In {\em International Conference on Machine Learning}, pages 14485--14508. PMLR, 2022.

\bibitem{kuhn2023semantic}
Lorenz Kuhn, Yarin Gal, and Sebastian Farquhar.
\newblock Semantic uncertainty: Linguistic invariances for uncertainty estimation in natural language generation.
\newblock {\em arXiv preprint arXiv:2302.09664}, 2023.

\bibitem{da2024llm}
Longchao Da, Tiejin Chen, Lu~Cheng, and Hua Wei.
\newblock Llm uncertainty quantification through directional entailment graph and claim level response augmentation.
\newblock {\em arXiv preprint arXiv:2407.00994}, 2024.

\bibitem{lin2023generating}
Zhen Lin, Shubhendu Trivedi, and Jimeng Sun.
\newblock Generating with confidence: Uncertainty quantification for black-box large language models.
\newblock {\em arXiv preprint arXiv:2305.19187}, 2023.

\end{thebibliography}

\clearpage
\appendix

%\section{Appendix Section}




\end{document}